\ifdefined\XeTeXversion\else\ifdefined\pdfoutput\pdfoutput=1\fi\fi
\documentclass{article}

\usepackage[T1]{fontenc}
\usepackage{microtype}
\usepackage{graphicx}
\usepackage{booktabs}
\usepackage{amsmath}
\usepackage{amssymb}
\usepackage{multirow}
\usepackage{array}
\usepackage{xcolor}
\usepackage{url}

\usepackage{hyperref}

\usepackage[accepted]{mlsys2024}
\makeatletter
\renewcommand{\Notice@String}{\unskip}
\makeatother

\newcommand{\method}{Scale-QLoRA}              
\newcommand{\scaleqast}{Scale-QLoRA}            
\newcommand{\scalelora}{Scale-QLoRA}           
\newcommand{\qat}{QAT-LoRA}                   
\newcommand{\ptq}{RTN-merge}                  
\newcommand{\pp}{\,pp}
\newcommand{\code}{\ensuremath{c}}

\newcommand{\best}[1]{\textbf{#1}}             
\newcommand{\bad}[1]{\textbf{#1}}              

\makeatletter
\g@addto@macro\UrlBreaks{%
  \do\a\do\b\do\c\do\d\do\e\do\f\do\g\do\h\do\i\do\j\do\k\do\l\do\m%
  \do\n\do\o\do\p\do\q\do\r\do\s\do\t\do\u\do\v\do\w\do\x\do\y\do\z%
  \do\A\do\B\do\C\do\D\do\E\do\F\do\G\do\H\do\I\do\J\do\K\do\L\do\M%
  \do\N\do\O\do\P\do\Q\do\R\do\S\do\T\do\U\do\V\do\W\do\X\do\Y\do\Z%
  \do\0\do\1\do\2\do\3\do\4\do\5\do\6\do\7\do\8\do\9}
\makeatother

\mlsystitlerunning{\smash{Scale-QLoRA: Code-Invariant Adapter Merging}}

\begin{document}

\twocolumn[
\mlsystitle{Scale-QLoRA: Code-Invariant Adapter Merging\\
for Native 4-bit Microscaling LLMs}

\begin{mlsysauthorlist}
\mlsysauthor{Tung-Ling Li\textsuperscript{$\dagger$}}{}
\mlsysauthor{Lee-Chi Wang}{}
\mlsysauthor{Jiale Huang}{}
\mlsysauthor{Janaki Ram Gotei}{}
\end{mlsysauthorlist}
\vspace{2pt}
{\centering\normalsize Crusoe.ai\par}
\vspace{4pt}
\mlsyscorrespondingauthor{Tung-Ling Li}{tli@crusoe.ai}
\mlsyskeywords{Scale-QLoRA, quantization, LoRA, MXFP4, NVFP4, adapter merging, efficient fine-tuning}

\vskip 0.3in

\begin{abstract}
Merging a LoRA adapter into its base model is standard deployment practice: it removes the runtime
adapter's per-forward overhead and leaves a single standalone checkpoint any serving stack can load.
On a \emph{native} 4-bit microscaling checkpoint (NVFP4, MXFP4) that step stops being free. The merged
weights must pass back through a quantizer, which re-derives the checkpoint's discrete E2M1 code
plane (roughly $90\%$ of the artifact's bytes). The deployed artifact is then coupled to one
quantization convention, and every later code-touching event in its lifecycle can move it. Done naively
the step is worse than fragile: it \emph{deletes} the adaptation, by up to $39\pp$, because against an
already-on-grid base the reconstruction optimum \emph{is} that base.

\textbf{\method{}} instead adapts only the native per-block scale field, trains those scales on the
deployment grid, and freezes every E2M1 code. Within a fixed native format, scale grid, block layout and
code plane, merging is then a \textbf{bit-exact identity} and the merged artifact is
\textbf{code-invariant}.
Across four models and four tasks, \method{} and merge-aware QAT-LoRA are both accuracy-lossless, so we
claim no accuracy ordering between them. They differ structurally: QAT-LoRA re-derives the code
plane through a quantizer, while \method{} preserves it exactly. That difference is what the lifecycle
prices. Nearest-rounding implementations disagree by about a point on the measured task, and more extreme
rule mismatches can drive the weight-space artifact to ${\sim}0\%$; we report that as a sensitivity bound
rather than a deployment frequency. Preserving the code plane also drops the weight-space straight-through
estimator from training ($3.9\times$ per step on the dense 8B model) and enables exact rollback, code-plane
deduplication, and a ${\sim}125\times$ faster scale-only task swap.
\end{abstract}
]

\printAffiliationsAndNotice{\textsuperscript{$\dagger$}Major contributor}

\label{introstart}
\section{Introduction}
\label{sec:intro}

\begin{figure}[t]
  \centering
  \includegraphics[width=\columnwidth]{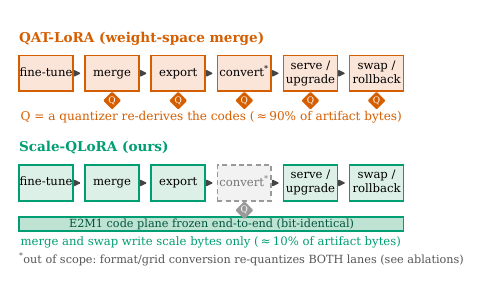}
  \caption{\textbf{A merged checkpoint is an artifact with a lifecycle.} A weight-space (\qat{})
  merge re-derives the E2M1 codes (roughly $90\%$ of the artifact's bytes) through a quantizer; any
  later code-plane re-derivation reintroduces that quantizer dependence.
  \method{} freezes the code plane end-to-end; merge and swap write scale bytes only. Format/grid
  conversion (marked $^{*}$) re-quantizes both lanes and is out of scope.}
  \label{fig:lifecycle}
\end{figure}

Post-training quantization to 4 bits is standard practice for serving large language models, increasingly
in \emph{microscaling} formats such as MXFP4 and NVFP4~\citep{ocp2023mxformats,nvfp4_2024}. There each
weight is a shared low-precision block scale times a 4-bit E2M1 code. Fine-tuning with
LoRA~\citep{hu2022lora} is equally standard, and one way to serve a fine-tuned quantized model is to merge
the adapter into the quantized weights. \textbf{Merging removes runtime-adapter overhead and produces a
standalone native checkpoint; we study the lifecycle of such merged artifacts.} \textbf{Whether deployments
predominantly merge is a claim about practice we did not survey, and we do not assert it}. Several
toolchains we exercise, vLLM included, \emph{also} serve a base plus runtime adapters. We price that
alternative in Appendix~\ref{sec:deployopts}; it beats merged scale planes on storage and swap latency. What
merging buys is measurable: serving the adapter at runtime instead costs $5.6\%$ prefill and $24\%$ decode
\emph{latency}, one run at one batch regime, which we do not generalise. A merged checkpoint is then an
\emph{artifact with a lifecycle}: re-exported by other tools, converted between formats, re-quantized at
engine upgrades, deduplicated, hot-swapped, rolled back. The E2M1 codes are roughly $90\%$ of its bytes
(Figure~\ref{fig:lifecycle}). This paper asks which merge strategy leaves them intact across those
events, within the artifact's native format.

\paragraph{The naive merge deletes the adapter.} The obvious recipe (dequantize, add the LoRA delta,
re-quantize) perturbs every code and scale. Under our full-data protocol it costs $37.4\pp$ on NVFP4
Llama-8B Banking77 ($93.0\to55.6$), with losses up to $39\pp$ across four post-hoc quantizers and tasks
(Table~\ref{tab:ptqbaselines}). The magnitude in raw points scales with how much gain the cell had to lose.
\emph{Which} rule deletes and which survives follows from the rule's reconstruction quality on the grid it
meets, and calibrated PTQ only partially rescues it (Section~\ref{sec:results-matrix}).

\paragraph{Merge-aware retraining ties on merge loss, and the tie hides a structural difference.} \qat{},
the merge-aware weight-space baseline, is LoRA trained with a native-grid straight-through estimator (STE)
that fake-quantizes the full weight every forward pass. The model trains on-grid, and its re-quantized
merge reproduces training. Its merge loss is at most $0.2\pp$ against full-data CIs of $\pm0.5$--$1\pp$.
Both merge-aware methods are therefore \textbf{accuracy-lossless}, and \textbf{merge loss is a tie}:
it is shared \emph{motivation}, not a contribution. But the two merged artifacts differ in kind. A \qat{} merge
re-derives every E2M1 code through a quantizer that must bit-match the training STE, and each later
code-touching event can repeat that re-derivation under a different rule. A scale merge writes only scale
bytes and provably never touches a code. We call this \textbf{code-invariance}
(Section~\ref{sec:codeinv-short}). Pre-quantizing once with the exact training rule is a real but partial
answer, discharging the export event but not the requirement to re-run an exactly-matching quantizer at
\emph{every} later one. What that convention costs comes in two tiers. \emph{Measured}: no two NVFP4
quantization outputs we audit agree bit-exactly, and divergence within the nearest-rounding family costs
about a point on the measured task. As \emph{a sensitivity bound, not an observed frequency}: a
rule from outside that family drives \qat{} to $\sim0\%$ on every dataset while \method{} reads flat.
How often the latter happens in deployment we do not measure.

\paragraph{Our approach: adapt the scales, freeze the codes, train on the native grid.} \method{} learns a
low-rank correction to the per-block scales only, leaves all codes frozen, and applies an STE that rounds
the effective scale onto the native grid during fine-tuning. Because training already runs on the on-grid
scale, merge writes exactly that scale byte and copies every code verbatim. Losslessness is
\emph{structural}: training establishes it, rather than the serving stack re-deriving it. The guarantee
is precise. Within a fixed native format, scale grid, block layout and code plane, \emph{no quantizer runs
at merge} and the merge is bit-exact: QAST has already rounded the scale onto that grid during training, so
merge writes the byte the model trained on and copies every code. Format conversion re-quantizes
\emph{both} methods' artifacts and sits outside both guarantees. Freezing the codes also removes the weight-space STE from training: \method{} trains
$3.9\times$ cheaper per step on the dense 8B model measured ($2.2\times$ on the 30B MoE).

\paragraph{Contributions.} Low-rank scale adaptation has close neighbors~\citep{tang2026lords}
(Section~\ref{sec:related}); our contribution is target-grid-aware low-rank adaptation of the native
microscale bytes that preserves the E2M1 code plane, with a lifecycle analysis and measurement of the
resulting artifact-level benefits.
\begin{enumerate}\itemsep1pt
  \item \textbf{Code-invariance and its five payoffs} (invariance of the served code plane, multi-adapter
  serving on a shared code plane, exact rollback, storage deduplication, and auditability), identified as
  the property separating the two accuracy-lossless merge strategies and grounded in a threat model of four
  re-quantization event classes (Section~\ref{sec:codeinv-short}).
  \item \textbf{Why the naive merge collapses.} On an on-grid base the reconstruction minimizing weight
  error \emph{is} the un-adapted base, so lower reconstruction error is associated with stronger deletion
  across the reconstruction-oriented quantizers we test, with explicitly reported exceptions. On the
  NVFP4 rounding cells the weight probe and accuracy disagree, which we record as open
  (Section~\ref{sec:deletion}).
  \item \textbf{Measured differentiators.} A deploy-quantizer sweep on Llama-8B (real vLLM), the Qwen-30B
  MoE and a 120B pair; multi-adapter serving with bitwise-identical code planes. Sharing one code plane cuts
  storage $3.0\times$ at $M{=}4$, and scale-only swapping is $\sim125\times$ faster than a weight-space swap
  that re-quantizes the checkpoint. Even so, \textbf{we claim no latency advantage over an all-resident
  pointer swap or a runtime-adapter system} (Appendices~\ref{app:codeinv},~\ref{app:cost}).
  \item \textbf{A representational bridge} from PEQA to weight-space
  adaptation~\citep{kim2023peqa,dettmers2023qlora}: the \emph{unrestricted} families nest as
  $\text{per-row}\subsetneq\text{per-block frozen-code}\subsetneq\text{weight-space}$, with \method{} a
  low-rank parameterization of the missing middle. That is a statement about representation, not accuracy.
  We \emph{do not detect} a deficit for the middle term, which is strictly weaker than equivalence and not
  to be read as it (Appendix~\ref{sec:bridge}).
\end{enumerate}

\label{pwstart}
\section{Background and the Merge Vulnerability}
\label{sec:background}
\label{sec:related}

\paragraph{Native FP4 microscaling storage.} A native FP4 linear stores each weight as
$W[o,i] = \code[o,i]\cdot \operatorname{eff\_scale}[o,g(i)]$, where $\code$ is a 4-bit \textbf{E2M1}
value (one of $\pm\{0,0.5,1,1.5,2,3,4,6\}$) and $\operatorname{eff\_scale}$ is shared across a
contiguous block of input elements, $g(i) = i \,\mathbin{/\!/}\, \text{block}$. An
$[\text{out},\text{in}]$ weight has $G = \text{in}/\text{block}$ blocks per row; the codes are the
bulk of the storage, the scales only $\sim 1/\text{block}$ of the parameters. The two deployed formats
differ only in how the block scale is stored. \textbf{NVFP4} (block 16) factors it as
$\operatorname{weight\_scale}\cdot\texttt{ws2}$, an \textbf{E4M3} byte (FP8, max 448) times one fp32
per-tensor scalar. \textbf{MXFP4} (block 32) stores one \textbf{E8M0} byte, a power-of-two exponent
with \emph{no mantissa}~\citep{ocp2023mxformats,rouhani2023mx,nvfp4_2024}. Both are symmetric with no
zero-point, so QA-LoRA's INT4 zero-point absorption trick does not apply.

\paragraph{The naive re-quantized merge is the failure mode.} Weight-space LoRA~\citep{hu2022lora} learns
$W \mathrel{+}= (BA)(\alpha/r)$ and must be merged as
$W' = \operatorname{requant}(\operatorname{dequant}(W) + (\alpha/r)\,BA)$, producing new codes
\emph{and} new scales, a step the trained model never saw, whose error is the collapse of
Section~\ref{sec:intro}. To make merge exact, whatever field the merge writes must already be what the
model trained on.

\paragraph{Prior families, and what each pays at merge time.} Appendix~\ref{app:related} states each in
full. \emph{Scale-only.} PEQA~\citep{kim2023peqa} fine-tunes one quantization scale per output channel with
codes frozen, and AlphaTuning~\citep{kwon2022alphatuning} is the earliest frozen-code precedent. Merge is
free and exact because only the scale moves. PEQA is a strong floor and even matches weight-space QLoRA on
some cells, and where a gap exists \textbf{that deficit is a direction supported on one cell, not a measured
range}. We do \emph{not} explain it by projection geometry.
\emph{Weight-space.} QLoRA~\citep{dettmers2023qlora} trains a low-rank \emph{weight} delta at high capacity,
but off the native grid, so the re-quantized merge is lossy. GPTQ~\citep{frantar2023gptq} and
AWQ~\citep{lin2024awq} are the calibrated re-quantizers we benchmark. LoftQ~\citep{li2024loftq} and the
quantized-PEFT wave~\citep{qin2024irqlora,guo2024lqlora,liao2024apiq,chen2024efficientqat} improve the
quantized base or its initialization, and all still merge through weight-space re-quantization.
\emph{Merge-aware weight-space.} Training LoRA with a native-grid straight-through estimator (STE),
$W \leftarrow W + (\operatorname{quantize}(W) - W)\texttt{.detach()}$, trains on-grid so the re-quantized
merge reproduces training. We call this baseline \qat{}; L4Q~\citep{jeon2024l4q} is the published prior in
this family and our baseline is a faithful instantiation of it on the native microscaling grid. It removes
the collapse, but its guarantee is \emph{conditional}: it re-derives every E2M1 code at merge, so it holds
only when whatever tool writes those codes uses the exact rule of the training STE. Re-quantizing
every weight at each forward step also makes it costly to train. LoTA-QAF~\citep{chen2025lotaqaf} also
merges losslessly, via grid-aligned \emph{ternary} adjustments to the quantized weights, sharpening a
distinction this paper relies on. Lossless merge and code-invariance are different properties, and
rewriting the codes forfeits the latter's payoffs. LR-QAT~\citep{bondarenko2024lrqat} absorbs low-rank auxiliaries into the
quantized tensor at the end, so the artifact is again a function of the quantizer.

\paragraph{Nearest neighbors.} Format-aware PTQ (MR-GPTQ~\citep{mrgptq2026},
ARCQuant~\citep{arcquant2026}) quantizes \emph{specifically} for the microscaling grids and is a stronger
post-hoc baseline than the four quantizers we sweep. Our collapse result should therefore be read as a
statement about what a \emph{merge} does to an adapter, not as a claim that FP4 PTQ is weak in general,
and neither work targets merge-time code preservation. QA-LoRA~\citep{xu2023qalora} is the nearest INT4
neighbor, absorbing the delta into INT4 \emph{zero-points}. Ours is the zero-point-free microscaling
analogue: the absorbing parameter is the per-block scale, and the coarse grid \emph{forces} the QAST STE
of Section~\ref{sec:method}. LSQ/LSQ+~\citep{esser2020lsq,bhalgat2020lsqplus}, PACT~\citep{choi2018pact} and
PoT/APoT~\citep{li2020apot} are the learned-step-size and power-of-two STE lineage QAST is adjacent to.
The STE technique is not new; its application to lossless adapter merge on microscaling hardware is.
LoRDS~\citep{tang2026lords} is the nearest algorithmic neighbor, also learning a low-rank correction in
scale space, but its scaling is continuous and served through custom kernels, so it produces no
standard-format checkpoint. \method{} instead adapts the \emph{existing} per-block scale bytes and
QAST-quantizes them onto the hardware grid during training. \textbf{What we ran is our own grid-STE-off
ablation, not LoRDS's implementation.}

\label{methodstart}
\section{Method: \method{}}
\label{sec:method}

\begin{figure*}[t]
  \centering
  \includegraphics[width=0.92\textwidth]{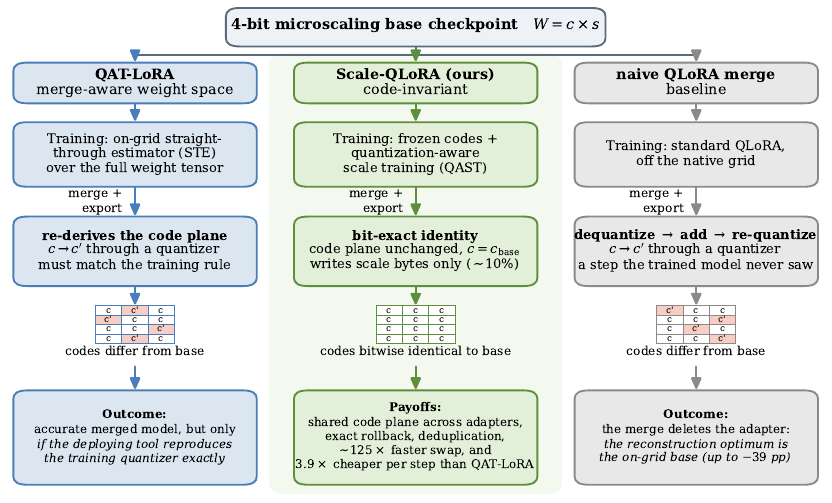}
  \caption{\textbf{Three merge paths from one native 4-bit checkpoint} ($W = c \times s$).
  \emph{Left:} \qat{} trains with an on-grid STE and merges accurately, but re-derives the code plane
  ($c \to c'$), so the served artifact holds different codes from the base. Its guarantee is
  conditional on the deploying tool reproducing the training quantizer. \emph{Centre:} \method{} freezes
  the codes and trains the block scales with QAST, so merge is a bit-exact identity
  ($c = c_{\text{base}}$) that writes scale bytes only. The code plane stays bitwise identical to the
  base, which is what the payoffs of Section~\ref{sec:codeinv-short} rest on. \emph{Right:} the naive
  path dequantizes, adds the delta and re-quantizes; that step alone deletes the adaptation. On
  an already-on-grid base the reconstruction optimum \emph{is} that base
  (Section~\ref{sec:deletion}), so no mismatch between quantizers is required. Training cost is
  in Appendix~\ref{app:cost}: \method{} is $3.9\times$ cheaper per step than \qat{} on the dense model,
  and we claim no accuracy ordering between the two merge-aware paths.}
  \label{fig:schematic}
\end{figure*}

\paragraph{Terminology.} \method{} denotes low-rank adaptation of a native checkpoint's per-block
\emph{scale} field, trained with \textbf{Quantization-Aware Scale Training (QAST)} so that the effective
scale sits on the deployment grid throughout training. We write \method{} throughout, in prose and in
table and figure labels alike. \emph{\method{} without QAST} names the ablation that adapts the same
scales off the native grid and rounds only at merge. We keep four properties distinct throughout:
a merge is \textbf{bit-exact} when the merged tensors equal the trained ones exactly
($\max|\Delta W|=0$); \textbf{accuracy-lossless} when merged and unmerged accuracy agree within the
stated interval; \textbf{code-invariant} when the merge provably writes no E2M1 code byte; and two
numbers are \textbf{statistically indistinguishable} when their difference lies inside the relevant
noise floor. For \method{} the construction simultaneously guarantees bit-exactness and code-invariance, and bit-exactness implies zero deterministic merge error in the represented weights. Accuracy-losslessness is an empirical downstream property. None of accuracy-losslessness, code-invariance, or statistical indistinguishability alone implies bit-exactness.

\paragraph{\scalelora{} forward.} Rather than correcting weights, we learn a low-rank correction to the
block scales only:
\begin{align*}
  \Delta S &= (B A)(\alpha/r), \qquad B:[\text{out},r],\;\; A:[r,G],\\
  \operatorname{eff\_scale}' &= \operatorname{eff\_scale} + \Delta S \quad \text{(per block)},\\
  W'[o,i] &= \code[o,i]\cdot \operatorname{eff\_scale}'[o,g(i)].
\end{align*}
The codes are unchanged, and the delta $\Delta S$ has the $[\text{out},G]$ shape of the stored
scales. $A$ is \textbf{block-level} ($[r,G]$, one column per scale block) rather than full-width
($[r,\text{in}]$), because there is one scale per block to correct. At equal rank \method{} therefore
carries fewer trainable parameters than weight-space QLoRA. The forward
$y = x\,W'^{\top}$ runs in fp32 for the scale arithmetic, because a sum of 32 bf16 products can overflow
fp16's 65504 ceiling.

\textbf{The precisions do not match, and that is the whole difficulty.} The factors $B$ and $A$ are
ordinary trainable tensors in bf16, so $\Delta S$ and hence $\operatorname{eff\_scale}'$ are continuous
values at training precision. The field they must be written into is not: a native checkpoint stores one
\emph{8-bit} scale per block, an E4M3 byte times a per-tensor fp32 on NVFP4 and an E8M0 exponent byte on
MXFP4. A continuous $\operatorname{eff\_scale}'$ is therefore not representable in general, and some
quantization of the scale is unavoidable. The only question is \emph{when} it happens. Applying it at
merge time is the naive choice and it costs $-27.5\pp$ (Table~\ref{tab:money}); QAST applies it in every
forward pass instead, which is what the next paragraph makes precise.

\paragraph{QAST makes merge exact by construction.} On real hardware the scale sits on a coarse grid
(E4M3 or E8M0). A model trained with a continuous fp32 $\operatorname{eff\_scale}'$ must round onto that
grid at merge, a rounding it never saw. That rounding produces the $-27.5\pp$ ``\scalelora{} without
QAST'' loss in Table~\ref{tab:money}. QAST closes this by rounding
the effective scale to the native grid \emph{in the forward pass}, with an STE so
gradients still reach the low-rank factors:
\begin{align*}
  \operatorname{eff\_q}   &= \operatorname{round\_grid}(\operatorname{eff\_scale}'),\\
  \operatorname{eff\_ste} &= \operatorname{eff\_scale}'
    + \big(\operatorname{eff\_q} - \operatorname{eff\_scale}'\big)\texttt{.detach()},
\end{align*}
so the forward uses the on-grid value while the gradient flows through
$\operatorname{eff\_scale}'$ to the low-rank factors. Here $\operatorname{round\_grid}(s)$ is
$\operatorname{e4m3}(s/\texttt{ws2})\cdot\texttt{ws2}$ for
NVFP4 and $2^{\operatorname{round}(\log_2 s)}$ for MXFP4 (snap to the nearest power of two). Because
training already runs on $\operatorname{round\_grid}(\operatorname{eff\_scale}')$, merge writes exactly
that on-grid scale byte and leaves the codes untouched. The merged model is therefore \textbf{bit-identical to
the trained model and merge loss is $0.0$ by construction}. QAST is a two-line change inside the
existing forward, run in the same single fine-tuning pass with no extra phase.

\paragraph{Losslessness holds on both native grids, for the grid QAST targets.} Since
$\operatorname{round\_grid}$ lands on a value the target byte represents exactly, the guarantee holds on
the mantissa-bearing E4M3 grid of NVFP4 and the coarser, mantissa-free E8M0 grid of MXFP4 alike. We
verify it directly by the weight-space $\max|\Delta W| = 0$ at merge, up to 100B+-parameter E8M0 MoEs,
the grids the real serving stacks use. The zero is exact, not a rounding coincidence: we verify it per
cell by $\max|\Delta W|=0$, regardless of seed count. Any residual $\le 0.01\pp$ in
aggregate accuracy is cross-seed eval noise. The guarantee carries one clause: merge reproduces training
only when the merge grid equals the training grid. A QAST-e4m3 adapter merged onto a mismatched E8M0 grid
reintroduces exactly the rounding QAST eliminates ($-13.0\pp$ Banking77, $-8.2\pp$ AGNews; Appendix~\ref{sec:ablations}). We train QAST for the target grid, and losslessness is then structural.
The precise statement, used throughout: \textbf{within a fixed native format, scale grid, block layout,
and code plane, no quantizer runs at merge and the merge is bit-exact.} The distinction matters, because
the scale \emph{is} quantized: QAST rounds it onto the target grid in every forward pass, so the rounding
has already happened by the time the merge runs. What the merge avoids is a \emph{second}, unseen
rounding of its own. We therefore say ``no quantizer at merge'' rather than ``quantizer-free''.

\paragraph{MoE: a batched grouped-expert kernel.} For MoE models the quantized linears are the experts,
and a per-expert Python loop over thousands of adapters builds an autograd graph that hangs
or runs out of memory. Our MoE implementation computes all experts' effective weights with batched
tensor operations, one \texttt{bmm} for the \method{} delta across every expert, then feeds the model's
native fused expert loop. In the DeepSpeed expert-parallel path we reuse the per-expert structure.

\paragraph{A single-setting stability recipe.} Scale-space training is more delicate than weight-space
because a scale error affects every weight in its block, and the correction
grows $\sim\sqrt{r}$ with rank. Three ingredients make one setting per model work, with no per-task
tuning, and each is a real cost the weight-space baseline does not pay.
\textbf{(i) Validation early stopping gives a robust single learning-rate (LR) strategy across the
regimes we test}: no fixed LR wins both. $2\times10^{-4}$ wins a weak-base task (Banking77 $92.4$) but
collapses a competent-base one (CLINC150 $18.6$), while $5\times10^{-5}$ underfits Banking77 ($82.0$).
Keeping the best-validation checkpoint, with the untrained base at step 0 as a candidate, reaches
$89.6$/$76.0$: near-best on both, with no collapse (Appendix~\ref{sec:ablations}).
\textbf{(ii) The E4M3 $\min(\operatorname{eff},448\cdot\texttt{ws2})$ clamp keeps MoE training
numerically stable}: large MoE bases have many scales pinned at the E4M3 ceiling from step 0, and the
plain STE leaks gradient above the ceiling, triggering unbounded growth in $B$ and overflow. On Qwen-30B,
clamp-off hits its first non-finite value at \textbf{step 18} and merges to $0.0$, while clamp-on trains
clean ($62.6$, merge loss $0.0$). We pair it with rollback-on-NaN.
\textbf{(iii) Merge must reuse the identical grid-and-clamp path of the QAST forward}: a merge path that
omits the forward's ceiling clamp casts ceiling-hitting merged scales to NaN, scoring $0\%$ and
introducing an artificial $\sim-92\pp$ merge loss. ``Lossless by construction'' holds only if merge and
forward share the same grid-and-clamp path.

\paragraph{Param-matching protocol.} \method{}'s block-level $A$ has $\sim1.8\times$ fewer
parameters per rank than weight-space QLoRA's $A:[r,\text{in}]$, so at equal rank \method{} is
under-budgeted. For a matched-budget comparison we raise the scale rank until the parameter counts agree
(scale \textbf{r58 $\approx$ qlora r32} on Llama-8B, \textbf{r54 $\approx$ r32} on Qwen-30B: $1293$M vs
$1285$M trainable). Between the two lossless methods the accuracy
ordering is task-dependent even within Llama, and the equal-rank ablation
(Appendix~\ref{sec:ablations}) shows scale is not disadvantaged at matched rank.

\section{Experimental setup}
\label{sec:setup}

\paragraph{Models, grids and datasets.} We evaluate four native-microscaling checkpoints spanning a
dense model and three MoEs, on both grids: Llama-3.1-8B dense, hereafter Llama-8B
(\texttt{nvidia/Llama-3.1-8B-Instruct-FP4}, NVFP4/E4M3, block 16)~\citep{dubey2024llama3};
Qwen3-30B-A3B MoE, hereafter Qwen-30B (\texttt{nvidia/Qwen3-30B-A3B-NVFP4}, NVFP4/E4M3)~\citep{qwen3_2025};
gpt-oss-120B MoE (128 experts, native \textbf{MXFP4/E8M0}, expert-parallel)~\citep{openai2025gptoss};
and DeepSeek-V4-Flash MoE (256 experts, native \textbf{MXFP4/E8M0}, 8-GPU expert-parallel, the largest
model in the study)~\citep{deepseekv4_2025}. The two 100B+ MoEs use the coarse power-of-2
\textbf{E8M0} block-scale grid of the OCP-MXFP4 standard that real serving stacks ship. QAST therefore
runs on the grid it must ultimately target. \textbf{Four fine-tuning tasks} form the main matrix, the
model $\times$ method grid we run in full, with held-out test sets spanning classification and
text-to-SQL: Banking77 intent classification
($n_{\text{test}}=3080$)~\citep{casanueva2020banking77}, AGNews topic classification
($7600$)~\citep{zhang2015agnews}, CLINC150 intent classification ($4500$)~\citep{larson2019clinc150},
and Spider text-to-SQL ($1034$, execution match)~\citep{yu2018spider}. MBPP code generation ($500$ test,
\textbf{pass@1} against each problem's asserts in a sandbox)~\citep{austin2021mbpp} is an
\textbf{additional generative evaluation}, not a main-matrix cell. Its training set and fine-tuning
headroom are both small, as Section~\ref{sec:limitations} quantifies.

\paragraph{Three methods (plus references).} \scaleqast{} (ours) trains a low-rank correction to the
per-block scales with an on-grid STE and freezes all E2M1 codes. \qat{}, the merge-aware weight-space
baseline, uses the STE of Section~\ref{sec:related} to fake-quantize the full weight onto the native
grid. \ptq{} is the naive re-quantized merge (round-to-nearest, RTN, unless a quantizer is named),
the motivating failure mode. We instantiate the \ptq{} row with four post-hoc quantizers (RTN, MSE,
AWQ, GPTQ), all applied to the \emph{same} single naive adapter with no per-quantizer retraining.
Calibrated PTQ (GPTQ) is therefore a strong baseline rather than a naive one. Two \emph{unquantized} references
bound the range, an un-fine-tuned bf16 model and a standard fp16 LoRA at matched rank; \textbf{these are
not run per cell} and carry no merge-loss concept. We also name \texttt{fp16\_lora+PTQ} (fine-tune in
fp16, merge, then PTQ) as the standard deploy workflow that \scaleqast{}/\qat{} avoid, but \textbf{that
is a named workflow, not a row we measure}. Its merge step performs the identical
dequantize--add--requantize operation as our \ptq{} rows and inherits their loss by construction
(Appendix~\ref{app:protocol}).

\paragraph{Recipe, and how we establish losslessness.} We use the \textbf{full training set} for every
task and train up to \textbf{3 epochs} with \textbf{validation early-stopping}, so weak bases train fully
and competent bases stop before over-perturbation. A \textbf{single uniform LR
$5\times10^{-5}$} holds across every model $\times$ dataset $\times$ method, with no per-task tuning.
Loss is \textbf{completion-masked} and scale ranks are \textbf{param-matched} to \texttt{qlora} r32. All
four \ptq{} rows derive from the same naive (no-STE) r32 adapter, and QAST uses the E4M3 ceiling clamp
with the matching clamp at merge. Merge loss $=$ merged (on the native grid) $-$ unmerged, and
\textbf{deployable accuracy $=$ accuracy of the merged, on-grid model}. We verify exactness itself
in weight space via $\max|\Delta W|$ ($=0$ exactly for \scaleqast{}, $\sim 3\times10^{-3}$ for \qat{}), so
``merged $=$ unmerged'' holds independently of the eval rather than as a coincidence of two
noisy accuracies. A few sub-studies retain numbers from a smaller \textbf{pilot protocol} used during
development. The full-data protocol supersedes it, and we mark every retained pilot number ``(pilot
protocol)''. Appendix~\ref{app:protocol} gives the per-model ranks, dataset sizes and clamp details.

\paragraph{Statistical rigor.} We evaluate the main matrix on the \textbf{entire} test set, giving
Wilson $95\%$ binomial confidence intervals (CIs) of roughly $\pm 0.5$--$1\pp$, an order of magnitude
below the collapse magnitudes of interest. \textbf{3 seeds} cover the Llama Banking77/CLINC150 and
Qwen-30B classification cells; every other row is single-seed. Two sub-studies use smaller evals and a
wider floor ($n{\approx}500$--$1000$, half-width $\pm 2.6\pp$), though there merge exactness rests on
$\max|\Delta|$ and is independent of $n$. The generative Spider metric carries \textbf{run-to-run} noise
well above its sampling floor: two runs of the same configuration \emph{and seed} differ by $1.84\pp$,
because execution-match scoring plus validation-based checkpoint selection amplifies nondeterminism.
The classification cells carry a smaller but non-zero floor of the same kind. A $120$B bridge eval
re-run with \emph{identical} code and adapter reproduces its unmerged accuracy only to within
${\sim}0.3\pp$, which matters because we report E8M0 merge losses of that order, so we pair every merged
number with an unmerged reference from the same run (Appendix~\ref{app:protocol}). \textbf{Any accuracy
gap smaller than the stated CI, or than $\sim2\pp$ on Spider, is indistinguishable from noise}.
Comparing two \emph{quantizers} against each other, rather than a merged number against its own
reference, widens that floor by roughly $\sqrt2$ (to $\pm4.1\pp$ at $n{=}500$ and ${\sim}88\%$ accuracy), so we claim no orderings inside it.

\paragraph{Real deployment stack.} Deployment experiments use vLLM 0.23.0 with modelopt-NVFP4 (Marlin)
on H200; merged checkpoints load as stock native checkpoints with \emph{no} custom kernel. To keep a
cross-backend floor from faking merge loss, we enforce \textbf{single-backend merge-loss discipline}:
a cell's merged and unmerged numbers always come from the \emph{same} backend. The 120B and DeepSeek
models train via DeepSpeed expert parallelism (EP)~\citep{rajbhandari2022deepspeedmoe}. Because EP
\texttt{generate} is broken (KV-cached decode through EP-MoE degrades to chance), we evaluate those
models via a single-process bridge or an in-EP logprob-over-candidates path, never with in-EP
generation.

\section{Main results: merge exactness, code-invariance, and its payoffs}
\label{sec:results}

Unless noted, every number in this section comes from the authoritative \textbf{full-data} protocol,
which supersedes the pilot protocol of Section~\ref{sec:setup}: full train set, up to 3 epochs with
validation early-stop, the \emph{entire} test set, uniform LR $5\times10^{-5}$.

\subsection{Both merge-aware methods are accuracy-lossless; only one is bit-exact}
\label{sec:results-matrix}

\begin{table*}[t]
  \centering
  \footnotesize
  \caption{\textbf{Full-data 3-way matrix: both merge-aware methods are \emph{accuracy}-lossless on every
  cell ($|\Delta|\le0.2\pp$), but only \scaleqast{} is bit-exact in weight space
  ($\max|\Delta W|$: scale $=0$ exactly, qat $\sim3\times10^{-3}$), while the \ptq{} path is lossy on
  every cell but one, catastrophically on the NVFP4 models and mildly on the two MXFP4 MoEs.} Merged
  accuracy \% / merge loss (pp). \emph{Base} is the un-adapted $4$-bit checkpoint on the identical
  harness as the row it sits in, and it is what makes the \ptq{} column legible: on several NVFP4 cells
  the naively merged model lands \emph{at} the base it started from, and elsewhere keeps only a small
  fraction of the gain, whose headroom-normalised form is Table~\ref{tab:retention}.
  Param-matched (scale rank $\approx$ qlora r32); NVFP4 for Llama/Qwen, native MXFP4/E8M0 for the two 100B+ MoEs; protocol in Section~\ref{sec:setup}.
  Row provenance and seed counts, noise floors, the Spider repetitions and the per-example SQL analysis are in Sections~\ref{sec:setup}, \ref{sec:results-matrix} and \ref{sec:limitations}; \textbf{we draw no accuracy ordering between the two merge-aware methods from this table}. Bold marks the exact $0.00$ merge-loss cells; the non-zero entries in the \scaleqast{} column are cross-seed evaluation noise over a merge verified exact per seed in weight space.}
  \label{tab:main}
  \begin{tabular}{@{}ll r ccc@{}}
    \toprule
    Model / grid & Dataset & Base & \scaleqast{} & \qat{} (merge-aware) & \ptq{} (naive) \\
    \midrule
    \multirow{4}{*}{\shortstack[l]{\textbf{Llama-3.1-8B}\\ NVFP4}}
      & banking77 & 49.2 & 93.7 / \best{0.00} & 93.0 / $-0.01$ & 55.6 / \bad{$-37.35$} \\
      & agnews    & 76.3 & 94.3 / \best{0.00} & 94.1 / \best{0.00} & 80.4 / \bad{$-13.30$} \\
      & clinc150  & 60.1 & 97.6 / $+0.01$     & 97.7 / $-0.01$ & 68.4 / \bad{$-29.30$} \\
      & spider    & 62.4 & 69.2 / \best{0.00} & 71.3 / $+0.10$ & 60.7 / \bad{$-9.67$} \\
    \midrule
    \multirow{4}{*}{\shortstack[l]{\textbf{Qwen3-30B-A3B}\\ NVFP4 (MoE)}}
      & banking77 & 69.6 & 93.9 / \best{0.00} & 93.2 / \best{0.00} & 69.5 / \bad{$-23.28$} \\
      & agnews    & 80.8 & 93.4 / \best{0.00} & 93.1 / \best{0.00} & 82.8 / \bad{$-11.11$} \\
      & clinc150  & 81.0 & 96.3 / \best{0.00} & 97.2 / \best{0.00} & 83.5 / \bad{$-13.96$} \\
      & spider    & 71.3 & 74.7 / \best{0.00} & 74.7 / \best{0.00} & 70.6 / \bad{$-3.29$} \\
    \midrule
    \multirow{4}{*}{\shortstack[l]{\textbf{gpt-oss-120B}\\ MXFP4/E8M0 (MoE)}}
      & banking77 & 73.9 & 91.3 / \best{0.00} & 89.6 / \best{0.00} & 90.9 / $-0.80$ \\
      & agnews    & 74.0 & 92.3 / \best{0.00} & 92.3 / \best{0.00} & 89.9 / $-2.60$ \\
      & clinc150  & 67.5 & 96.8 / \best{0.00} & 96.0 / \best{0.00} & 94.3 / $-2.60$ \\
      & spider    & 71.3 & 70.7 / \best{0.00} & 73.5 / \best{0.00} & 71.0 / $-1.45$ \\
    \midrule
    \multirow{4}{*}{\shortstack[l]{\textbf{DeepSeek-V4-Flash}\\ MXFP4/E8M0 (MoE)}}
      & banking77 & 60.6 & 96.8 / \best{0.00} & 95.0 / \best{0.00} & 94.0 / $-2.00$ \\
      & agnews    & 33.6 & 92.8 / \best{0.00} & 91.2 / \best{0.00} & 86.8 / \bad{$-3.60$} \\
      & clinc150  & 66.0 & 97.8 / \best{0.00} & 97.4 / \best{0.00} & 98.2 / $-0.80$ \\
      & spider    & 40.4 & 70.2 / \best{0.00} & 76.0 / \best{0.00} & 72.2 / $+0.6$ \\
    \bottomrule
  \end{tabular}

  \vspace{3pt}
  {\footnotesize Eval protocol, subsampling and noise floors: Section~\ref{sec:setup}.}
\end{table*}

\paragraph{Both merge-aware methods are accuracy-lossless, so merge loss is a tie.}
On every cell $|\text{merge loss}|\le0.2\pp$. The two methods earn that differently. For
\scaleqast{} losslessness is a numerical identity ($\max|\Delta W|=0$ exactly, codes copied verbatim).
\qat{}'s merge is not bit-exact ($\max|\Delta W|\sim3\times10^{-3}$), so its $\le0.2\pp$ is an
empirical result inside the $\pm0.5$--$1\pp$ CI rather than an identity. Merge loss alone is therefore a
\textbf{tie}, and it is our \emph{motivation}, not the headline. The tie holds on both native grids: the
two 100B+ MoEs are native \textbf{MXFP4/E8M0}, trained under expert parallelism and evaluated through a
single-process logprob bridge, and \textbf{both merge-aware methods have negligible accuracy merge loss
on those largest-MoE cells}. The two remain different in kind there: \scaleqast{} is exact in weight
space ($\max|\Delta W|=0$, codes copied verbatim), whereas \qat{} is empirically lossless in downstream
accuracy but re-derives the discrete code plane and is not bit-exact. The exact-merge result therefore
reaches the grid real serving stacks ship.

\paragraph{No accuracy ordering survives, and the reason differs by cell.} The gaps between the
two lossless methods are small and model/task-dependent: \scaleqast{} reads marginally higher on
several coarse-grained classification cells, \qat{} on Llama/Qwen CLINC150 and the single-run Spider
pairs. \textbf{None of those orderings is one we can support.} A 3-seed study on the Qwen-30B
classification cells supports no ordering at that seed count. Two of the three \scaleqast{}$-$\qat{} gaps
sit at zero, $-0.04$ (AGNews) and $-0.14\pp$ (CLINC150). The third, $+0.82$ (Banking77), does
\emph{not} shrink under repetition, but a per-method cross-seed spread of its own order ($\le1.1\pp$)
leaves it unresolved. Three seeds no more establish equivalence than one establishes an ordering, and
every \scaleqast{} seed does still merge bit-exactly ($\max|\Delta W|=0$).
\textbf{The Spider ordering is not a method difference either, and we have the repetitions to say so.}
Running the identical Llama configuration \textbf{seven} times per method gives \scaleqast{}
$69.88\pm0.83$ against \qat{} $69.38\pm1.15$: $+0.50\pp$ with $95\%$ CI
$[-0.55,+1.55]$, indistinguishable from zero and the \emph{opposite sign} to the single-run pair in
Table~\ref{tab:main} ($-2.03\pp$). \qat{}'s own seven runs of one configuration span $3.77\pp$, so a
single Spider run cannot resolve a $2\pp$ method difference. On gpt-oss the Spider gap ($-2.8\pp$)
exceeds that cell's entire fine-tuning headroom. DeepSeek Spider, whose gap is \emph{above} its cell's
noise floor, is addressed instead by Section~\ref{sec:limitations}, to which every later mention points.
That section states this paper's Spider position in full: a swing of nearly twenty points between two
clean \scaleqast{} runs, and a per-example scoring that locates the difference as \emph{syntactic}
rather than semantic, an instability in decoding rather than a capacity ceiling. That instability is
what we flag as the open problem.
On the remaining single-seed \emph{classification} cells, each gap is smaller than a single cell's own
confidence interval, a weaker check than the wider two-proportion interval a between-method
comparison requires (Section~\ref{sec:setup}).
\textbf{We claim no general accuracy ordering between \method{} and \qat{}}: the
code-invariance payoffs (Section~\ref{sec:codeinv-short}) are the central contributions and the
lossless-vs-naive-collapse contrast is the shared motivation. At matched parameters on Llama-8B the two
lossless methods are likewise tied (mean $88.7$ vs $89.0$; Appendix~\ref{app:cost}, which also isolates
QAST against a scale adaptation trained off the native grid).

\paragraph{The \ptq{} path collapses on the NVFP4 models and is mild on the MXFP4 MoEs.} Under the
full-data protocol, naive re-quantized merge is lossy on every cell but one, and its magnitude in raw
points scales with how much gain the cell had to lose. At comparable movement, what separates a deleting
rule from a surviving one is not the movement but the rounding rule's
reconstruction quality on the grid it meets (Section~\ref{sec:deletion}). On
many-class classification on the NVFP4 models it is catastrophic (Llama Banking77 $-37.4\pp$, CLINC150
$-29.3\pp$; Qwen Banking77 $-23.3\pp$; AGNews $-11$ to $-13\pp$ on both). \textbf{The E8M0 MoEs behave
differently, and we state it plainly}: there the loss is mild ($-0.8$ to $-3.6\pp$) even where the
fine-tuning headroom is large. The \emph{Base} column, measured on each cell's own harness, makes
those figures legible. On seed 0, Llama Banking77 falls from $93.96$ unmerged to $55.06$ merged under
both RTN and AWQ ($54.87$ under MSE), just $5.8\pp$ above
the $49.22$ base. Averaged over that cell's three seeds, RTN retains $15\%$ of the fine-tuning gain
(Table~\ref{tab:retention}) and
destroys about $85\%$ of it, while GPTQ recovers $91.46$ and keeps $94\%$. \emph{The triple is one seed
and the retention its cell's seed mean; the two come from separate runs and are not to be combined.}

\paragraph{The grid axis is confounded, and we name it before drawing anything from it.} The two
grids differ in scale precision, block size, model family and scale \emph{and} in which models sit on
them. The NVFP4-versus-E8M0 contrast in Table~\ref{tab:retention} is therefore not a clean grid
experiment. We ran the ablation that separates block size from scale precision, and it comes back
\emph{null} against the grid explanation: rounding rule, not grid, predicts retention. Appendix~\ref{app:deletion} gives the
ablation and the full list of what is confounded with what.

\paragraph{Reaching for a better rounding-only quantizer does not rescue the merge; calibration does.}
Swapping RTN for a better \emph{rounding} rule (MSE, AWQ) leaves the collapse essentially intact, because
the failure is in the reconstruction objective rather than in the search over roundings. Error-feedback
calibration (GPTQ) does escape it on both grids, which is why we carry it as a strong baseline rather
than a strawman. It escapes at a price: a calibration set, a deploy-coupled artifact, and a per-expert
Hessian that a sparse MoE under a calibration budget cannot fully supply. Table~\ref{tab:ptqbase} gives the per-quantizer
rows and Appendix~\ref{app:gptq} the conditioning statistics.

\subsection{Code-invariance and its five payoffs}
\label{sec:codeinv-short}

Both merge-aware methods are accuracy-lossless, so the contribution is not merge loss but the
\emph{artifact} each merge leaves behind. A \method{} merge writes scale bytes only and copies the E2M1
code plane verbatim, so the merged artifact is \textbf{code-invariant}; a \qat{} merge re-derives that
plane through a quantizer. \textbf{The payoffs below are five consequences of that one
property, not five independent ones}: four carry their own measurement, and auditability is a corollary
of the code-byte identity measured in the second. Appendix~\ref{app:codeinv} develops them with the
lifecycle threat model behind them. That model names four event classes that can re-derive codes:
export-tool mismatch, format conversion (which re-quantizes \emph{both} methods' artifacts and so sits
outside both guarantees), engine upgrade, and a continuous merged tensor handed to a serving-side quantizer.

\begin{figure*}[t]
  \centering
  \includegraphics[width=0.64\textwidth]{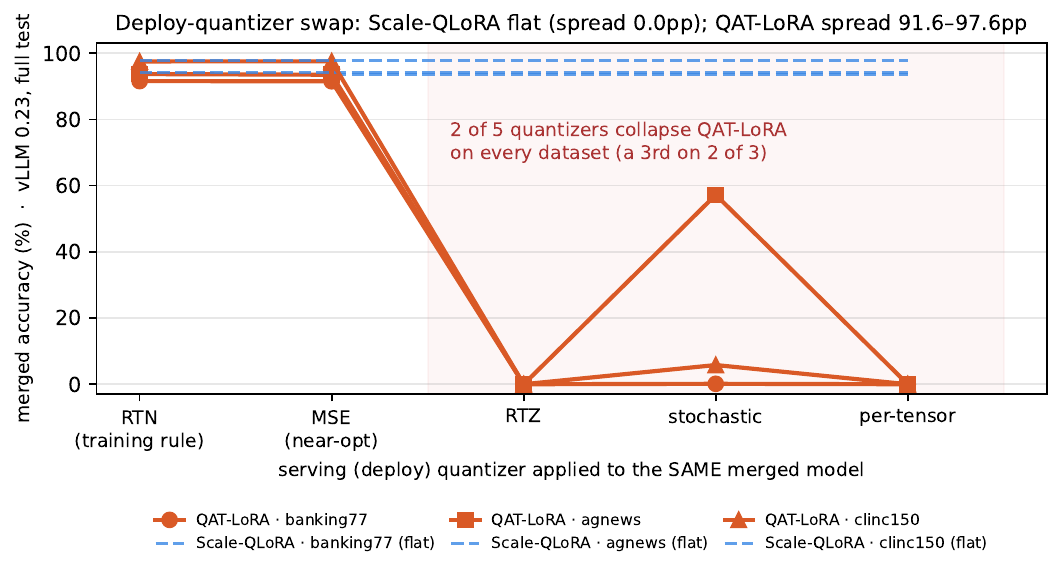}
  \caption{\textbf{Deployment-quantizer sweep on a real vLLM 0.23.0 NVFP4 stack (Llama-3.1-8B),
  full-data adapters.} Merged accuracy under 5 deploy rounding rules $\times$ 3 datasets.
  \qat{} collapses to $\sim0\%$ under RTZ and under the \emph{per-tensor} rule (whole-tensor
  scaling, which replaces the per-block scales with one scale for the whole tensor, a granularity
  change rather than a rounding change, and the reason that column collapses the un-adapted base too
  and is flagged confounded), and under stochastic on 2 of 3 datasets: a per-dataset spread of
  $91.6$--$97.6\pp$ on the full test set.
  \textbf{The two conditions are not the same measurement.} The \qat{} rows are five re-quantizations of one
  \emph{continuous merged weight tensor} (merged in full precision, handed to a serving-side quantizer
  with no intervening native export), leftmost column its own training rule; the \scaleqast{} rows are
  \emph{one} evaluation of \emph{one} artifact repeated across the five columns, because no rule is
  applied to it: its code plane is served as stored. That $0.0\pp$ spread is the property under
  test, not a five-way agreement; the positive control, in which the rules \emph{are} run over a
  scale-merged artifact and return a bit-level identity, is the gpt-oss pair in
  Appendix~\ref{sec:deploysweep}, \textbf{on a different grid, model and harness, so the scale arm
  has no on-grid positive control here}, and the one finer-grid control it does have, the zero-delta
  base control, is an item we mark open (that chain is assembled in Appendix~\ref{sec:deploysweep}).
  Only round-to-nearest of the five is substantiated as shipping behaviour; the others span
  deterministic conventions the format permits and we did not observe them in a deployed stack. Which
  columns move is tracked by the rule family, not by the input: RTN and MSE lie inside the
  nearest-rounding family on this grid, RTZ, stochastic and per-tensor outside it
  (Section~\ref{sec:differentiators} partitions all five).
  What a rule instead costs a \emph{correctly} pre-materialized \qat{} artifact, and what a load-time
  re-quantizer would cost a scale-merged artifact (the zero-delta base control), are inferences scoped
  in Appendix~\ref{sec:deploysweep}.}
  \label{fig:deploysweep}
\end{figure*}

\paragraph{(1) The served code plane is invariant under the deploy-rule family.} Swept across five
deployment rounding rules on a real vLLM 0.23.0 NVFP4 stack (Llama-8B, three datasets, full test sets;
Figure~\ref{fig:deploysweep}),
\qat{} spans nearly the entire accuracy range: round-toward-zero and whole-tensor scaling collapse it
to $\sim0\%$, for per-dataset best-to-worst spreads of $91.6$--$97.6\pp$. \scaleqast{} is flat,
$0.0\pp$ spread on every dataset. The pattern replicates on the Qwen-30B NVFP4 MoE ($93.2$--$97.2\pp$),
though on our own single-backend re-quantization harness rather than on vLLM. At 120B one rule pair
(RTN vs.\ RTZ) on one dataset confirms it: \qat{} drops $95.6\to9.1$.
\textbf{The two conditions are not the same measurement, and the asymmetry is the point}: five actual
re-quantizations of a \qat{} adapter merged in full precision, a configuration we construct rather
than one we observed a stack ship. Against that stands \emph{one} \scaleqast{} artifact read across five
columns with no rule applied to it. Its flatness is therefore a property of that artifact, licensed by the
power-of-two grid, where the rules \emph{were} run over a scale-merged artifact and returned an identity
($\max|\Delta W|=0$). \textbf{That is a different grid, model and harness, so on the swept grid the scale arm
has no on-grid positive control.} The one finer-grid control it has, the zero-delta base control,
reads non-zero and is marked open. \textbf{What this establishes is a property contrast on the rule
axis, not a lifecycle price}, and of the five rules only round-to-nearest is substantiated as shipping
behaviour. The lifecycle price is instead the cross-tool audit's in-family divergence, about a point,
plus the requirement to re-run an exactly-matching quantizer at every later code-touching event.
Appendix~\ref{sec:deploysweep} states both tiers canonically and in full, with the controls,
the whole-tensor confound, and the two halves of the argument that are inferences rather than measured
rows.

\paragraph{(2)--(5) Sharing, rollback, deduplication, audit.} Across the four saved Llama-8B task
adapters, all \textbf{224/224} target-linear code planes are \textbf{bitwise identical} to the base and
to each other ($0$ differing code bytes), and only the scale bytes differ. \qat{}'s re-quantized merges
change $7.4$--$8.0\%$ of \emph{code} bytes per task, so its merged models share no code plane.
Relative to storing $M$ independently merged weight-space checkpoints, that one identity yields
$3.0\times$ less storage at $M{=}4$ tasks. In our one-resident-checkpoint experiment, scale-only
swapping is $\sim125\times$ faster than a weight-space swap that re-quantizes the checkpoint, and
\textbf{we claim no latency advantage over an all-resident pointer swap or a runtime-adapter system}.
The Qwen-30B MoE reproduces the storage ratio, and the swap direction and byte accounting at a smaller
factor (Appendix~\ref{sec:multiadapter}). A merged artifact is restored by subtracting the on-grid
scale delta actually applied at merge, verified byte-for-byte, so no second full base
checkpoint need be retained. Distribution patches are $21.1\times$ smaller
compressed ($9.0\times$ by chunk-level dedup), and a byte-level audit follows as a corollary of the same
identity. \textbf{We do not win on every axis, and three of these
are payoffs against the merged-artifact design specifically.} A quantized base plus $M$ runtime
adapters stores as little or less and can switch by pointer, so against it, and against a server holding
all merged variants resident, we claim only zero per-forward overhead. The weight-space swap we time
re-quantizes, where an in-place code-plane overwrite from an already-materialized sibling would not, a
configuration we leave untimed. Retaining the base checkpoint plus the small adapter gives \emph{either}
method exact rollback, and a base-plus-runtime-adapter server produces no per-task merged artifact to
deduplicate, patch or audit at all. On those last three axes the code-plane identity adds that
they hold with no retained base and no runtime-adapter path (Appendix~\ref{app:codeinv} prices all of
it).

\subsection{Why the naive merge collapses: the reconstruction objective returns the un-adapted base}
\label{sec:deletion}

The collapse in Table~\ref{tab:ptqbaselines} reads like quantization noise swamping a small update. It
is not. The cause is a near-immediate structural fact about the objective a naive merge poses, and we
state it before testing it:

\begin{quote}\small
\textbf{Claim (two cases, one reason).} Let (i) the base $W$ be exactly on the representable grid and
(ii) the update be sub-step: $|\Delta_i|$ below the distance from $W_i$ to its rounding midpoint, for all
but the small fraction of elements measured below. \emph{Case~1, a fixed nearest-rounding rule,} needs a
third premise, (iii): that the rule re-derives the \emph{same} block scale for a sub-step update; given
it, an on-grid value is a fixed point of nearest-rounding, so re-quantizing $W+\Delta$ returns $W$.
\emph{Case~2, a reconstruction minimizer that searches over scales,} needs no premise about scale
re-selection, but it does need a \emph{block-level} version of (ii). Because $W$ is representable, the
error $\lVert\Delta\rVert$ is always attainable; whether it is optimal is the Voronoi question. Writing
$D = Q'-W$ for any competing representable block $Q'$, the base is the minimizer exactly when
$\langle\Delta, D\rangle \le \tfrac12\lVert D\rVert^2$ for all $Q' \neq W$. A clean sufficient
condition for that is $\lVert\Delta\rVert \le \tfrac12\min_{Q'\neq W}\lVert D\rVert$.
\textbf{Element-wise sub-step does not by itself imply this}, because a quantizer free to move the scale
and the codes together can reach a block that no single-coordinate midpoint test sees
(Appendix~\ref{app:deletion} gives an explicit two-element counterexample). We therefore state Case~2 at
block level and rest it on the measured margin (below, and $45$--$200\times$) rather than on the
element-wise test alone. The block-level criterion is exactly computable, because at a fixed scale the
reconstruction-minimising codes are determined by rounding and the search collapses to a scan of the
finite scale grid. Measured that way on $140{,}000$ blocks, the un-adapted block is the global
reconstruction optimum on $99.4\%$ of them, and the fixed-rule premise (iii) holds on $99.9\%$
(Appendix~\ref{app:deletion}). \emph{Consequence:} either way the
merge returns the un-adapted model up to that sub-step fraction (it \emph{deletes} the adapter rather
than degrading it), and in Case~2 a \emph{better} reconstructor is a \emph{worse} merge.
\emph{Premise (iii) is what divides the grids, and it is where the exception lives}: the power-of-two
grid's $\lceil\cdot\rceil$ rule is a \emph{scale-selection} rule and $\Delta$ makes it re-select the
exponent on a measured minority of blocks, so (iii) demonstrably fails there and that grid escapes. On
the finer grid the scale byte is instead stable, which the percent-level displacement below evidences.
That is joint evidence for (ii) and (iii) together, the NVFP4 scale-byte stability fraction not being measured
separately (Appendix~\ref{app:deletion} adds three refinements of this scope).
\end{quote}

\noindent Three measurements test the claim (an intervention that reverses the effect, and two
predictions whose falsifiers were written down first), and Appendix~\ref{app:deletion} gives all three
in full.

\paragraph{The objective's optimum \emph{is} the un-adapted model.} On the natively-quantized E8M0 MoEs a
weight-MSE scale search is catastrophic where plain RTN is nearly harmless. On gpt-oss-120B,
MSE costs $-16.8\pp$ on AGNews and $-17.3\pp$ on Banking77 against RTN's $-2.6$ and $-0.8$; each merge
loss here is taken against its own run's unmerged reference as above, not against the column
reference of Table~\ref{tab:ptqbaselines}. The merged accuracy lands at the
un-adapted base ($75.6$ vs base $74.0$; $74.4$ vs base $73.9$, both inside the base's Wilson interval), which is
what the Claim predicts for \emph{any} merge minimizing weight reconstruction error. Both premises hold
and the predicted solution is the one selected: the base is on the grid (on-grid residual
$0.000\mathrm{e}{+}00$ across $40$ expert matrices), the median $|\Delta_i|$ is $45$--$200\times$ smaller
than the distance to its own rounding midpoint (every element sub-threshold in $>99.5\%$ of blocks), and
MSE's scales land $7.5\mathrm{e}{-}5$ from the base, recovering $1.7$--$8\%$ of $\Delta$ at an attained
reconstruction error of $\|\Delta\|$ itself. RTN is not solving that minimization at all: its fixed
$\lceil\cdot\rceil$ rule bumps the block exponent \emph{because of} $\Delta$ on $30.2\%$ of blocks and
retains essentially all of the gain. That census covers \emph{every} expert matrix in the model
($9{,}216$ matrices, $3.58\times10^{9}$ blocks). A $\Delta{=}0$ control re-quantizing the un-adapted
checkpoint through the identical path moves exactly zero blocks and zero codes, so the entire $30.2\%$ is
attributable to $\Delta$ rather than to re-quantization being non-idempotent.

\textbf{We measured the attribution that this premise was owed, and it holds from both sides.} Splitting
the merge by whether a block's exponent moved, we evaluated each half on the same bridge harness and
device split ($n{=}1000$, unmerged reference $92.6$ in every row). Applying the merge on the bumped
$30.2\%$ alone gives $89.6$, reproducing the full merge's $89.4$; the complementary
$69.8\%$ alone gives $74.7$, against a base of $74.6$ measured in the same run. \textbf{The bumped
minority reproduces the entire merge; the non-bumped majority recovers none of the fine-tuning gain}
($83.3\%$ versus $0.6\%$ of the gain retained). This is a two-sided attribution rather than an inference
from complementarity, and the prediction was registered before the cells were run. The weight-space census agrees
independently: the bumped blocks hold $>99.99\%$ of the merge's displacement energy, and $99.99\%$ of all
code changes fall inside them, so on non-bumped blocks $\Delta$ is sub-step and rounds away exactly.

\textbf{The alignment we had inferred is false, and we report the measurement instead.} The displacement
does \emph{not} lie along $\Delta$: $\cos(Q(W{+}\Delta)-W,\,\Delta)$ is $0.017$ aggregated and $0.141$ at
the per-matrix median, never above $0.23$ in any grouping we computed. What is true is weaker and
sufficient for the mechanism: the displacement's \emph{component} along $\Delta$ exceeds $\|\Delta\|$
itself (a regression coefficient of $1.5$ on \texttt{gate\_up}, $16.9$ aggregated). RTN therefore
overshoots $\Delta$ while burying it inside a much larger, essentially orthogonal re-rounding
perturbation that the model tolerates. The $12\times\|\Delta\|$ displacement quoted above is a
\texttt{gate\_up} figure ($11.92$ on this census). It does not generalise across projections:
\texttt{down}'s ratio grows with depth and the norm-weighted value over all matrices is far larger.
Accuracy space is then a consistency check rather than a second measurement. MSE retains $3$--$9\%$ of the
headroom-normalized gain on the gpt-oss E8M0 cells \emph{with substantial headroom} (the range
excludes the sub-headroom gpt-oss Spider row Table~\ref{tab:retention} italicises and excludes), and is
negative on all four DeepSeek cells, while RTN
retains $86$--$102\%$. \textbf{We quote that as a same-order correspondence on this one case, not as a
proportionality between the two instruments in general.}

\paragraph{Three tests, and what each one establishes.} \emph{An intervention that reverses the effect}:
GPTQ (error feedback on a fixed grid, whose optimum is \emph{not} the base), run on top of the very MSE
scales that had deleted the adapter, recovers gpt-oss AGNews from $75.6$ to $92.1$. That localizes the
failure in scale selection alone rather than in the FP4 grid or in calibrated quantization generally.
\emph{A pre-registered prediction with a stated falsifier on a
second model}: on DeepSeek-V4-Flash Spider, RTN merge keeps essentially all of the headroom
while the weight-MSE search lands inside the base's own Wilson interval, a $33.8\pp$ separation on the
identical adapter and eval path. \emph{A pre-registered offline weight statistic}: the displacement a
merge produces from the base, in units of the adapter, separates the two NVFP4 rounding rules from E8M0
RTN by a factor of ${\sim}1000$ and tracks retention across
both grids. \textbf{Minimizing weight reconstruction error is thus the wrong objective for a merge}:
\textbf{across the reconstruction-oriented quantizers we test, lower reconstruction error is associated
with stronger deletion of the adaptation}, a relation an independent DeepSeek-V4 replication traces
across an order of magnitude of reconstruction
quality. \textbf{We report one exception explicitly, AWQ on DeepSeek AGNews, which breaks
the ordering across every cell we measured}: reconstruction quality does not explain it, and we treat it
as one anomalous cell rather than a property of AWQ on microscaled grids (Appendix~\ref{app:awq}). The relation
is directional only, and reconstruction error is not a sufficient statistic for merge loss in raw points.
Normalising by headroom (Table~\ref{tab:retention}) yields the sharper invariant, that \textbf{a
weight-MSE scale search destroys $105$--$110\%$ of the adaptation gain on every DeepSeek dataset}. We
read that headroom as an \emph{upper bound}, those cells being evaluated with expert-parallel capacity
dropping on, which inflates the measured gain (Table~\ref{tab:retention} quantifies the shift).
\textbf{We state what the pre-registration does and does not buy}: the finer grid's value is
\emph{entailed} by premises (i) and (ii) as the same probe measures them, and the coarse grid's had been
measured before the prediction was written. The test therefore buys the \emph{magnitude} of the
separation, not its sign. \textbf{Its two comparison groups are also unequal in extent and the surviving one is far
the smaller} (sixteen sampled linears of one model against a single layer's experts), so the
separation must not be read with the breadth of the sixteen-cell retention table whose pattern it tracks.
Appendix~\ref{app:deletion} gives all three tests, their falsifiers and the probe's provenance in full.

\paragraph{Where the claim is literal, and where the two instruments disagree.} The ``returns the
un-adapted model'' statement is literal on the coarse grid and approximate on the fine one, and our two
instruments, a weight-space displacement probe and end-task accuracy, do not agree on the NVFP4
rounding cells. We record that disagreement as open rather than resolving it in favour of whichever
instrument suits the claim; Appendix~\ref{app:deletion} states both readings in full.

\subsection{Why adapting only the scales suffices}

\begin{figure}[t]
  \centering
  \includegraphics[width=\columnwidth]{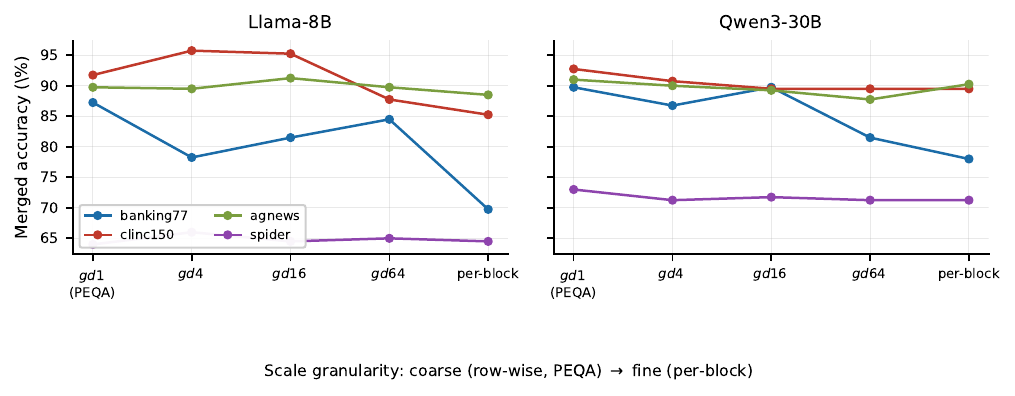}
  \caption{\textbf{Granularity is a smooth dial, not a cliff.} Merged accuracy of \emph{direct}
  (non-low-rank) scale fine-tuning as the scale granularity is swept from row-wise (PEQA) to per-block.
  No setting collapses, and the coarsest one is a strong floor rather than a deficit; the capacity
  that matters is therefore not bought by granularity alone. Every point merges exactly. Same values as
  Table~\ref{tab:peqaladder}; pilot bridge protocol.}
  \label{fig:ladder}
\end{figure}

\label{sec:whyscales}

\textbf{Separate expressivity} names two findings of different evidential standing, and we keep them
apart. \emph{First, measured under the full-data protocol on two cells}: the weight update's task-useful
direction demonstrably does not live in the scale subspace. Projecting a freely-trained weight update
onto the per-block-scale subspace recovers only a minority of the task gain, which rules out the
``captures the rescaling component'' story. Its captured energy sits exactly at the chance value for a
subspace of that size ($1/16$), a sanity check rather than a second measurement. \emph{The matched-norm
random-direction control that would calibrate how low a ``minority'' is was not run} (we ran it only
for the geometric statistics we set aside). This finding therefore rests where the appendix rests it, on the
independent-training result below. (A weight update trained \emph{on} the native grid instead projects almost
perfectly, but its own training constrained it toward the representable set, so that projection reports
the constraint rather than the alignment of the beneficial direction; Appendix~\ref{sec:bridge}.)
\emph{Second, a null rather than a match}: an \emph{independently}-trained scale correction is not
distinguishable from weight-space LoRA at the protocols we ran. The gap it has to close is smaller than
that protocol's noise floor, so the result is a null: we cannot detect a difference, and we cannot rule
one out. A
full-precision control isolates the mechanism: for the same per-block multiplicative adaptation on an
\emph{unquantized} bf16 model we detect no deficit against
weight-space LoRA, so the capacity is multiplicative block structure
and not the 4-bit grid. That control is single-seed on one dense model and two classification tasks, against a
cross-seed floor imported from the quantized cells, which is the scope of that claim. Taken together they
license freezing the codes at no accuracy cost this evidence can detect. \textbf{That is a null, not an
equivalence.} The containment $\text{per-row}\subsetneq\text{per-block frozen-code}\subsetneq\text{weight-space}$, which holds between the
\emph{unrestricted} families rather than at fixed rank, states what each parameterization can
\emph{represent}. ``We do not detect an accuracy deficit'' is strictly weaker, and the two must not be conflated.
Appendix~\ref{sec:expressivity} develops that containment, the projection and full-precision
probes, the granularity ladder and the low-rank cap that regularizes it, rank curves, the QAST grid
ablation, a Lean-4 study that detects no fitting-capacity deficit (a fitting-capacity result, explicitly
not a task-success one), and the geometric statistics we report only to set aside.

\section{Scope and Discussion}
\label{sec:limitations}

\paragraph{The claims of this paper, ordered by evidential standing.} Our results are of four kinds,
labelled once here so each later statement can be read at its own strength.
\emph{Structural (mathematical).} Within a fixed native format, scale grid, block layout and code
plane, a \method{} merge is a bit-exact identity ($\max|\Delta W|=0$, codes copied verbatim), and the
served code plane is invariant by construction. \qat{} re-derives that plane and is not bit-exact.
\emph{Strong empirical.} Both merge-aware methods are accuracy-lossless
($\le0.2\pp$ against full-data CIs of $\pm0.5$--$1\pp$), so merge loss is motivation, not
contribution. The naive re-quantized merge is lossy on every cell but one, and freezing the codes makes
training cheaper per step.
\emph{Limited empirical.} The one measured lifecycle event separating the two merge-aware methods is
in-family export-tool divergence, about a point on the measured task. Those code-plane payoffs are
measured on one dense model with a partial MoE replication, and only against
merged-artifact designs. To a quantized base plus $M$ runtime adapters we concede storage, swap and
the existence of any per-task artifact to deduplicate, patch or audit at all; there we claim only zero
per-forward overhead. \emph{Inference or open.} The out-of-family drive of the weight-space artifact to
${\sim}0\%$ shows how sensitive a re-derived code plane can be, not how often that happens. Also
unresolved are the attribution of retained adaptation to exponent-changing blocks, the
weight-probe-versus-accuracy disagreement on the NVFP4 rounding cells, and the zero-delta base
control's non-zero reading on the finer grid. \emph{Accuracy}: we claim no ordering, anywhere.

\paragraph{Scope, in brief; Appendix~\ref{app:scope} states each of these in full.} The contribution is
code-invariance and its payoffs, plus the expressivity and bridge characterisations. The naive-PTQ collapse
is motivation, and we assert \textbf{no accuracy ordering anywhere}: the apparent orderings between the
two merge-aware methods do not survive repetition where repetition exists, and are unsupported where it does
not. QAST targets a \emph{specific} hardware scale grid, so train-time grid selection must match the
deployment target. The guarantee holds within a fixed native format, scale grid, block layout and code
plane: there no quantizer runs at merge and the merge is bit-exact. A QAST-e4m3 adapter merged to E8M0 instead loses
$13.0/8.2\pp$. The study spans a dense 8B model and three MoEs across four main-matrix tasks on full test
sets. We ran three seeds on the Llama Banking77/CLINC150 and Qwen-30B classification cells; every other
row is single-seed, and two sub-studies use a wider $\pm2.6\pp$ floor. \textbf{Every measured baseline here
is our own instantiation}: we do not run the original PEQA, QA-LoRA, L4Q, LoTA-QAF or LoRA-Inlaid
implementations, so comparisons to those works are analytical. The deploy sweep is narrower than it looks
and three of its limits are load-bearing: it is confounded in one column, asymmetric by design, and it
prices no lifecycle event. Spider is our noisiest metric ($1.84\pp$ run-to-run at fixed seed), and the one
place the two lossless methods appear to separate. We repeated both arms on the only cell with both ample
headroom and a gap above the floor, and \scaleqast{} did not reproduce ($70.2$ then $51.8$). The difference
concentrated in \emph{non-executable} rather than semantically wrong output. That is decoding instability, not a
demonstrated capacity ceiling. Finally, GPTQ's error feedback needs a per-expert Hessian that a sparse MoE
under a calibration budget cannot supply. We report that as a limit on calibrated PTQ's \emph{guarantees},
not as a doubt about these rows.

\section{Conclusion}
\label{sec:conclusion}

Merging a LoRA adapter into a native 4-bit microscaling model does not end the checkpoint's life:
the merged artifact is exported, converted, re-quantized, swapped, and rolled back. The naive
re-quantized merge silently loses accuracy (up to $39\pp$ on the dense NVFP4 model and more on the
MXFP4 MoEs, with unpredictable sign). A merge-aware weight-space STE removes the collapse, so merge
loss is a tie: both merge-aware methods are accuracy-lossless, and \textbf{we claim no accuracy advantage
from merging itself, and no general accuracy ordering between the two}. \method{} instead freezes the
codes and trains a low-rank correction to the block scales on the native grid, making the merged
artifact \emph{code-invariant}. Within a fixed native format, scale grid, block layout, and code plane,
the merge is a bit-exact identity by construction. The weight-space method re-derives the
code plane, and is accuracy-lossless only when its training quantization convention is
reproduced. Codes are roughly $90\%$ of the artifact's bytes, and the merge never hands one to a
quantizer. A scale-merged artifact therefore carries no more exposure than the un-adapted base, verified as
an identity on the power-of-two grid. On the finer grid it is bounded rather than nil by the
zero-delta base control, whose non-zero reading there we record as open. The
weight-space artifact is driven to ${\sim}0\%$ only by a rule from outside the nearest-rounding family
the export tools all implement, a demonstration of how sensitive a re-derived code plane can be, not
a measured frequency. An in-family tool costs about a point on the measured task
(Appendix~\ref{sec:deploysweep} states both tiers in full). Pre-materializing once
with the exact training rule discharges that accuracy exposure, at the cost the same rule imposes on
the base. It discharges neither the requirement to repeat it at every later code-touching event, nor format
conversion (which re-quantizes both methods' artifacts), nor the code-plane payoffs.
Those payoffs begin with invariance of the served code plane under the deploy-rule family: flat $0.0\pp$
against \qat{} spreads of $91.6$--$97.6\pp$ on a real vLLM stack, replicated on a 30B MoE (on
our own re-quantization harness, not on vLLM), and confirmed
by a 120B pair. That is a property contrast on the rule axis, not a lifecycle price.
Multi-adapter serving runs on one shared code plane: bitwise-identical codes across four task
adapters. \emph{Relative to storing $M$ independently merged weight-space checkpoints}, that gives
$3.0\times$ storage sharing at $M{=}4$; and, in our one-resident-checkpoint experiment, scale-only swapping runs
$\sim125\times$ faster than a weight-space swap that re-quantizes the checkpoint. Against runtime-adapter
serving, and a server holding all variants resident, we claim only zero per-forward overhead and no
latency advantage. Rollback subtracts the on-grid scale delta applied at merge,
verified byte-for-byte and needing no second base checkpoint, and distribution is
patch-sized ($21.1\times$ less than a weight-space merge) with byte-level auditability. Those hold
likewise against merged artifacts only, since a retained base plus the small adapter
lets \emph{either} method roll back by re-merging and a runtime-adapter server emits no per-task
artifact to patch or audit at all. Training is
$3.9\times$ cheaper \emph{per step} on the dense 8B model measured, all
at model-dependent, not universally-better, accuracy. Why scales suffice is separate expressivity, two
findings of different standing. The trained weight update overlaps the per-block-scale subspace only at
chance (full-data, two cells). Trained independently, that subspace closes the QLoRA$-$PEQA gap to
within the pilot bridge protocol's noise floor where a gap exists, without touching the codes: a
null inside a floor wider than the gap, not a scale-space accuracy win. \method{} is thus the missing
middle
of the unrestricted families, $\text{per-row}\subsetneq\text{per-block frozen-code}\subsetneq\text{weight-space}$, and a low-rank
parameterization of that middle term. Our contribution is not
low-rank scale adaptation per se, but target-grid-aware, low-rank adaptation of the native
microscale bytes that preserves the E2M1 code plane. We add a lifecycle analysis and
measurement of the resulting artifact-level benefits. Future work: end-to-end E8M0 training at
scale, broader generative tasks, promoting the pilot expressivity studies to the full-data
protocol, and a direct empirical comparison with continuous low-rank scaling
(LoRDS; \citealp{tang2026lords}).

\section*{Acknowledgements}
We thank Hongliang Liu, Itay Lamprecht, Matan Halfon and Omri Berkovitch for reviewing early drafts of
this paper and for the feedback that shaped it.

\label{endmaintext}
{\footnotesize
\bibliographystyle{mlsys2024}
\bibliography{references}
}

\label{startappendix}
\appendix
\section{Protocol Details}
\label{app:protocol}

This appendix records the protocol minutiae behind Section~\ref{sec:setup}: the method and reference
definitions in full, the training recipe, the statistical floors and how each was measured, and the
serving stack.

\paragraph{Three methods (plus references).} \scaleqast{} (ours) trains a low-rank correction to the
per-block scales with an on-grid STE and freezes all E2M1 codes; \qat{}, the merge-aware weight-space
baseline, uses the STE of Section~\ref{sec:related} to fake-quantize the full weight onto the native
grid; and \ptq{} is the naive re-quantized merge (round-to-nearest, RTN, unless a quantizer is named),
the motivating failure mode. We instantiate the \ptq{} row with four post-hoc quantizers (RTN, MSE,
AWQ, GPTQ), all applied to the \emph{same} single naive adapter with no per-quantizer retraining, so
calibrated PTQ (GPTQ) is a strong baseline rather than a naive one.
Two \emph{unquantized} references bound the range: \texttt{fp16\_base} (the un-fine-tuned bf16 model,
the accuracy \emph{floor}) and \texttt{fp16\_lora} (standard fp16 LoRA at matched rank, the accuracy
\emph{ceiling}). \textbf{These fp16 references are not run per cell}: they exist for the dense model only, on
the reduced-$n$ ceiling sub-study of Table~\ref{tab:bf16}, and they carry no merge-loss concept, since
an fp16 LoRA merges into fp16 losslessly. We also name \texttt{fp16\_lora+PTQ} (fine-tune in fp16,
merge, then PTQ the merged weights to NVFP4) as the standard deploy workflow that
\scaleqast{}/\qat{} avoid. \textbf{That is a named workflow, not a row we measure separately}: its
merge step performs the identical dequantize--add--requantize operation as our \ptq{} rows and so
inherits their loss by construction, with the magnitude reported in Table~\ref{tab:ptqbaselines}.

\paragraph{Recipe (full-data protocol).} We use the \textbf{full training set} for every task
(Banking77 $\approx$9.6k / AGNews 19.6k / CLINC150 14.6k / Spider 6.85k / MBPP 374 examples).
Training runs up to \textbf{3 epochs} with \textbf{validation early-stopping} (patience 2, $\ge$1
epoch, base@step0 a candidate), so weak bases train fully and competent bases stop before
over-perturbation. A
\textbf{single uniform LR $5\times10^{-5}$} spans every model $\times$ dataset $\times$ method (no
per-task tuning); loss is \textbf{completion-masked} with \textbf{MAX\_LEN 768}. Scale ranks are
\textbf{param-matched} to \texttt{qlora} r32 (scale \textbf{r58 / r54 / r52 / r55} for llama / qwen /
gpt-oss / deepseek); the four \ptq{} rows all derive from the same naive (no-STE) r32
adapter. QAST uses the E4M3 $\min(\operatorname{eff},\,448\cdot\texttt{ws2})$ ceiling-clamp with the
matching clamp applied at merge (a ceiling-hitting scale otherwise casts to NaN and introduces an
artificial $\sim{-}92\pp$ merge loss), plus rollback-on-NaN. Merge loss $=$ merged (on the native grid) $-$
unmerged; \textbf{deployable accuracy $=$ accuracy of the merged, on-grid model.}
A few sub-studies retain numbers from a smaller \textbf{pilot protocol} (3k train / 500 eval /
1 epoch) used during development; the full-data protocol supersedes it, and we mark every retained
pilot number ``(pilot protocol)''.

\paragraph{Statistical rigor.} We evaluate the main matrix on the \textbf{entire} test set, giving
tight Wilson $95\%$ binomial confidence intervals (CIs) of roughly $\pm 0.5$--$1\pp$ (e.g.\ Banking77
$n{=}3080$, AGNews
$n{=}7600$), an order of magnitude below the collapse magnitudes of interest. We additionally run
\textbf{3 seeds} on the Llama Banking77 and CLINC150 cells and on the Qwen-30B classification cells.
Table~\ref{tab:main} reports single-seed values for every row except the two Llama
intent-classification rows, which are $3$-seed means; the multi-seed Qwen values are reported in
Section~\ref{sec:results-matrix}. Two
sub-studies use smaller evals and a correspondingly wider floor: the earlier seed battery and the
device\_map bridge evals of the EP-trained 120B / DeepSeek models use $n{\approx}500$--$1000$ (Wilson
half-width $\pm 2.6\pp$ at $\sim$90\% accuracy), but there merge exactness rests on $\max|\Delta|$ and
is independent of $n$. The generative Spider metric additionally carries \textbf{run-to-run} noise
well above its sampling floor: two runs of the same configuration \emph{and seed} differ by $1.84\pp$,
because execution-match scoring combined with validation-based checkpoint selection amplifies
nondeterminism. The classification cells carry a smaller but non-zero floor of the same kind: re-running a
$120$B bridge eval with \emph{identical} code and the \emph{same} adapter reproduces the same unmerged
accuracy only to within ${\sim}0.3\pp$ ($3$ samples at $n{=}1000$). We measured this three ways on
gpt-oss AGNews ($92.5$, $92.4$, and $92.7$), where the latter two used the \emph{same} four GPUs, so
the spread is ordinary run-to-run nondeterminism in the bf16 MoE forward and not an artifact of a changed
layer-to-device split. It matters because we report E8M0 merge losses of that same order, so we pair every
merged number with an unmerged reference from the same run and treat ${\sim}0.3\pp$ as the floor on those
cells. \textbf{Any accuracy gap smaller than the stated CI, or than $\sim2\pp$
on Spider, is indistinguishable from noise.} When we compare two \emph{quantizers} rather than a merged
number against its own unmerged reference, the relevant floor is wider by roughly $\sqrt2$, since two
independent proportions enter: at $n{=}500$ and ${\sim}88\%$ accuracy the $95\%$ half-width on a
\emph{difference} is $\pm4.1\pp$. We therefore do not claim orderings inside that band: on DeepSeek
AGNews, GPTQ's $-2.0$ and RTN's $-3.6$ are indistinguishable, whereas AWQ's $-10.8$ is separated from
both.

\paragraph{Real deployment stack.} Deployment experiments use vLLM 0.23.0 with modelopt-NVFP4 (Marlin)
on H200; merged checkpoints load as stock native checkpoints with \emph{no} custom kernel. To keep a
cross-backend floor from faking merge loss, we enforce \textbf{single-backend merge-loss discipline}:
we evaluate Llama with vLLM (native, $\sim$35$\times$ throughput) and Qwen with HF, and a cell's
merged and unmerged numbers always come from the \emph{same} backend. The 120B and DeepSeek models
train via DeepSpeed expert parallelism~\citep{rajbhandari2022deepspeedmoe}; because EP
\texttt{generate} is broken (KV-cached decode through EP-MoE degrades to chance), we evaluate those
models via a single-process bridge or an in-EP logprob-over-candidates path, never with in-EP
generation.

\section{Code-Invariance: Threat Model and the Five Payoffs in Full}
\label{sec:differentiators}
\label{app:codeinv}

This appendix gives the evidence summarized in Section~\ref{sec:codeinv-short}: the lifecycle threat
model, the deployment-quantizer sweep with its controls and replications, multi-adapter serving and
storage, rollback, deduplication, and the audit corollary. Recapping:
a \qat{} merge re-derives every E2M1 code, roughly $90\%$ of
the artifact's bytes, through a quantizer, while a \method{} merge writes only scale bytes and leaves
the code plane bit-identical to the base. The five payoffs below, one per subsection (1)--(5), are
\textbf{five consequences of one property, not five independent
ones}: four carry their own measurement (deploy-rule flatness, swap time and storage, patch bytes,
rollback restore), and auditability is a corollary of payoff (2)'s code-byte identity.

\paragraph{The threat model: an artifact state machine.} A merged checkpoint moves along the chain
\emph{training representation $\to$ exported bytes $\to$ conversion/load operation $\to$ served
bytes}, and each of the four event classes below is a transition on that chain that can re-derive
codes. \emph{(a) Export-tool mismatch (training representation $\to$ exported bytes):} HuggingFace,
modelopt, and llm-compressor implement round-to-nearest with different scale-selection and
tie-breaking rules, so exporting a merged model with a tool other than the one matching training
re-derives the codes under a different rule. Re-quantizing the merged
Llama-8B weights produced six NVFP4 quantization \emph{outputs} from five independent codebases (our
training STE, run both on the weights training optimized and on its own exported copy; our RTN harness;
NVIDIA modelopt; compressed-tensors, which is llm-compressor's NVFP4 weight path; and HuggingFace's
quantize-on-load backend),
and \textbf{no two of those outputs agree bit-exactly across all $224$ linears (0 of 15 pairs)}. One
pair is the input-divergence control rather than implementation divergence: the training
quantizer on its own bf16 export flips $1.5\times10^{-4}$ of codes. \textbf{Provenance bounds what
that census counts}: three of the six outputs are our own harnesses (the training STE on its two
inputs, and our RTN path) and three are independently shipping tools, so only three of the enumerated
pairs are tool-against-tool between implementations neither of which is ours. Agreement is not
uniformly absent: pairs implementing the \emph{same} rule are
bit-identical on up to $47/224$ layers, differing only where per-tensor-scale
arithmetic crosses an E4M3 boundary, at $\le0.2\pp$ on the full Banking77 test
($n{=}3080$), inside the $\pm0.9\pp$ band. What costs accuracy is a different
\emph{convention}: holding the per-tensor \emph{scalar} fixed at the base's calibration value while
retaining the per-block scales costs $-0.81\pp$, and HuggingFace's shipped
quantize-on-load convention $-1.04\pp$ ($91.59$ vs $92.63$), both significant under a paired
McNemar test ($p=0.002$) though inside the unpaired band. \textbf{Both costed divergences are one
task's measurement and only one of them ships}: both are measured on the banking-intent test set named
just above and on no other task, the first being a convention we construct to isolate
per-tensor-scalar handling, the second a shipped default.
\textbf{Every tool here implements
amax-derived per-block \emph{nearest} rounding, differing only in tie-breaking, per-tensor-scalar
handling and an adaptive scale rule, and every divergence this class produced costs about a point.}
The chain also has a verified
\textbf{safe transition}: a \qat{} checkpoint correctly materialized once with its byte-exact training
quantizer, then loaded by a code-reading stack, serves correctly, and our control measures exactly
$0$ differing bytes on that path.
\emph{(b) Format conversion (exported bytes $\to$ exported bytes):} converting
NVFP4$\leftrightarrow$MXFP4 or repacking block-16$\leftrightarrow$block-32 re-quantizes every
block, and it re-quantizes \emph{both} methods' artifacts, so neither guarantee survives it. The
measured cost is the cross-grid ablation (a QAST-e4m3 adapter merged to E8M0 loses $13.0/8.2\pp$;
Table~\ref{tab:qastablation}), so grid selection at training time must match the deployment
target. \emph{(c) Engine upgrades (load: exported bytes $\to$ served bytes):} serving stacks that
re-derive scales or repack weights at load apply their current quantizer to the stored artifact.
\emph{(d) A continuous merged tensor handed to a serving-side quantizer (a never-materialized training
representation $\to$ exported bytes):} merge in full precision (the dequantized base plus the
trained update, never written as native bytes), then quantize with a serving-side tool. The
transition defines the class; what varies inside it is which adapter produced the tensor.
\emph{(d.i) An fp16-trained adapter} is the \texttt{fp16\_lora+PTQ} workflow of
Section~\ref{sec:setup}, discharged by argument rather than by a row of our own: its merge performs the
identical dequantize--add--requantize operation as the naive merge, so it inherits that loss by
construction, at the magnitude Table~\ref{tab:ptqbaselines} reports.
\emph{(d.ii) A \emph{merge-aware}-trained adapter} is the object the sweep below hands to five rules,
a configuration we construct rather than one we observed a stack ship; those rules stand in for
whichever rule meets such a tensor at load, the event class (c) names.
What separates the sweep's magnitudes from the audit's is not the input (both apply differing
rules to a continuous merged tensor) but the \emph{rule family}, which we partition here once.
\emph{Inside} the amax-derived per-block nearest-rounding family: RTN, the
rule every audited export tool also implements, and the weight-MSE scale search, which on the
mantissa-bearing E4M3 grid the sweep runs on resolves to nearest rounding per block; it barely
moves codes and costs nothing. That same rule \emph{leaves} the
family on the coarser power-of-two grid, where the search re-selects the block exponent and
Section~\ref{sec:deletion} measures it as the deleting rule: membership is a property of the rule
\emph{and the grid it meets}, not of its name. \emph{Outside} the family: round-toward-zero,
stochastic rounding, and whole-tensor scaling (a granularity change rather than a rounding change).
\textbf{The family is the partition the two magnitude tiers fall either side of, and we report it as
that and not as the mechanism that sets the magnitude}: the audit's own costliest divergences are
\emph{scale} conventions, as is the costliest swept rule, so scale handling spans the observed range
and the family boundary is drawn across it. Where a deploy rule \emph{is} run over a
\scaleqast{}-merged artifact, at 120B, it is a verified identity ($\max|\Delta W|=0$);
Section~\ref{sec:deploysweep} scopes this once, in full, event class by event class and grid by grid.

\subsection{Payoff 1: the served code plane is invariant under the deploy-rule family}
\label{sec:deploysweep}

\textbf{Swept across five deployment rounding rules, the weight-space merge spans nearly the entire
accuracy range while the scale merge does not move at all.} We measure this, the sharpest event class,
on a real vLLM 0.23.0 NVFP4 stack: 5 deploy rounding
rules $\times$ 3 classification datasets on the \textbf{full test set}, re-run on the
correctly-trained full-data adapters (Figure~\ref{fig:deploysweep}). \textbf{RTZ and whole-tensor
scaling (the figure's \emph{per-tensor} column) collapse \qat{} to $\sim0\%$ on every dataset, and stochastic rounding collapses
it on two of three (AGNews retains $57\%$).} The per-dataset best-to-worst spread for \qat{} is
\textbf{$91.6\pp$} (Banking77, $n{=}3080$), \textbf{$93.7\pp$} (AGNews, $n{=}7600$), and
\textbf{$97.6\pp$} (CLINC150, $n{=}4500$); \textbf{\scaleqast{}'s spread is $0.0\pp$, flat, on every
dataset} (Banking77 $93.67$, AGNews $94.21$, CLINC150 $97.76$, identical across all five
quantizers). On Banking77, for instance, \qat{} moves from $91.6\%$ under round-to-nearest to
\textbf{$0.0\%$} under RTZ, while \scaleqast{} holds $93.67\%$ throughout.

\textbf{The two conditions are not the same measurement}: five actual perturbations of a \qat{} adapter
merged in full precision (a configuration we construct rather than one we observed a stack ship),
against \emph{one} \scaleqast{} artifact read across five columns with no rule applied to it
(Figure~\ref{fig:deploysweep} states the asymmetry in full).
\textbf{That leaves the sweep's scale condition without an on-grid positive control}: on this finer
grid the flat rows are a property of the artifact by construction; the positive control that turns them into evidence (the
rules run over a scale-merged artifact and returning a bit-level identity) exists only on the
power-of-two grid, a different grid, model and harness; and the one finer-grid control that does exist,
the zero-delta base control below, is the item we mark open. So on the grid the sweep is swept, the
evidence that a rule meeting such an artifact returns it unchanged is cross-grid. Two measurements would
close it: the five rules run over an NVFP4 scale-merged artifact, and the block-scale identity check that
settles the power-of-two case.
\textbf{This subsection therefore establishes a
property contrast on the rule axis, not a lifecycle price}: the lifecycle price is the audit's
in-family about-a-point plus the every-event requirement, both derived below.

\paragraph{The positive control at 120B: run the rules over a scale-merged artifact and flatness
becomes a bit-level identity.} The flat rows above are a claim about what a deploy rule finds when it
meets a scale-merged artifact, so we ran one against exactly that. A minimal RTN-vs-RTZ pair on
gpt-oss-120B (MXFP4/E8M0, CLINC150, $n{=}1000$ logprob bridge) closes the scale axis. The harness
gate reproduces the training-rule control ($95.6$ vs the Table~\ref{tab:main} reference $96.0$).
Re-quantizing the \qat{}-merged model with RTZ collapses it $95.6\to\textbf{9.1}$ ($-86.5\pp$) with
block scales \emph{bit-identical} between the two rules ($50.8\%$ of codes differ): pure code
rounding. Re-quantizing the \scaleqast{}-merged model with either rule is a \emph{verified functional
identity}: $\max|\Delta W| = 0.0$ across all $115.6$B expert weights, so its $96.9$ carries over
unchanged, a figure that sits a tenth of a point off this model's own entry in
Table~\ref{tab:main}, so \emph{both} sides of the control disclose a harness offset, the weight-space
side's being the larger of the two.
The confound control closes the argument: the un-adapted base reads $66.6$ under \emph{both} rules, as
does the \scaleqast{} merge, so RTZ harms only the model whose merge re-derived codes through a
continuous intermediate.

\paragraph{The collapse is a train-vs-deploy mismatch, not ``a bad quantizer.''} A zero-delta base
control (re-quantizing the \emph{un-adapted} base with each rule) is nearly harmless: RTZ shifts the
base by at most $-5.7\pp$ from its training-grid accuracy ($49.2$/$76.3$/$60.1$ on
Banking77/AGNews/CLINC150), stochastic rounding by at most $1.3\pp$, and only whole-tensor scaling
collapses the base too. The same RTZ quantizer is catastrophic on \qat{}, because on the merged model it flips
\textbf{$71\%$} of codes ($22.7\%$ mean weight change), versus \textbf{$0.40\%$} for the benign MSE
quantizer, which costs $0\pp$. The penalty is therefore genuine mismatch, and \textbf{its size spans the
full range of rules the format permits}: a near-optimal deploy quantizer costs $0\pp$ while a
coarse-but-valid one destroys the model. Which rule a given stack applies is a property of that stack,
which we did not survey and about which we assert nothing; consistent with the two-tier partition below,
the in-family cost is \emph{measured} (about a point) and the out-of-family collapse is priced as a
\emph{bound}, not a frequency. What the spread obliges is the standing bit-matching requirement: a
weight-space merge is correct only where the deploy rule matches the training rule exactly, at every
code-touching event. Because that whole-tensor rule collapses even the base, it reads as ``a coarse quantizer
breaks everything''; the clean isolators are RTZ and stochastic.

\paragraph{The control is our estimate for the scale arm here, and on this grid the base should not
have moved at all.} A \scaleqast{}-merged artifact differs from the base only in scale bytes, so what a
rule costs the base is what it would cost the merge, whereas the \qat{} collapses are measured. But an
exactly representable value is a fixed point of truncation and of stochastic rounding as much as of
nearest rounding, so a rule that only rounds codes at a fixed block scale must leave an on-grid base
untouched, and the native NVFP4 base is on-grid, to five decimals (Section~\ref{sec:deletion}). The
power-of-two grid behaves exactly so: in the pair above the un-adapted base reads identically under
both rules, block scales bit-identical between them, which is what licenses calling RTZ there pure code
rounding. The finer grid's control nonetheless records the base shifting under both. \textbf{We mark
the discrepancy open.} On that grid either the rule is doing something past code rounding, re-deriving
the block scale or the per-tensor scalar, or our harness is; which of the two is undetermined, because
the block-scale identity check that settles it on the power-of-two grid was not run on the finer one.
The cost is stated with the Claim in Section~\ref{sec:deletion}: premise (iii)'s NVFP4 half is already
joint evidence rather than a separate measurement, and this control now sits in tension with it too.
The scale arm's estimate here is bounded by what the control records, and bounded is not zero.

\paragraph{The collapse replicates on a 30B MoE.} We repeat the full sweep on the Qwen-30B NVFP4 MoE
(Banking77 $n{=}3080$, AGNews $n{=}7600$, CLINC150 $n{=}4500$, full test, same full-data adapters),
using a single-backend torch re-quantization harness (the vLLM harness is Llama-only;
Section~\ref{sec:limitations}). The pattern reproduces on all three datasets: RTZ and whole-tensor
scaling drive \qat{} to \textbf{$0.0\%$} everywhere, stochastic rounding degrades it to $17.1$--$38.8\%$
(at or below the $25\%$ four-class chance floor on AGNews, where the dense sweep retained $57\%$),
and best-to-worst spreads are \textbf{$93.2$--$97.2\pp$}, while a near-optimal deploy quantizer
again costs nothing ($93.2$--$97.2$ vs the training grid's $93.1$--$97.2$). \scaleqast{} is flat at
$93.9$/$93.4$/$96.3$ across all five quantizers ($0.0\pp$ spread on every dataset), identical to its
merged accuracy in Table~\ref{tab:main}.

\paragraph{What pre-materializing once does and does not cover (stated canonically here; every later
mention points back to this paragraph).}
\emph{Covered outright, and on which grid.} \scaleqast{} freezes codes, so nothing in its merge path
hands them to a quantizer: classes (a) and (d) are covered on both grids, the codes written to disk
being the vendor's own; class (b) re-quantizes both methods' artifacts and sits outside both
guarantees. Class (c) runs its rule on whatever artifact it is given, so it must be stated per grid: on
MXFP4 we measured it on the merged artifact and it is an identity (the gpt-oss pair above), while on
NVFP4 we did not run all five rules over the scale-merged artifact, so our estimate there is the
zero-delta base control. Either way \scaleqast{} hands the rule a stock native checkpoint differing
only in scale bytes and inherits the un-adapted base's exposure and no more (that exposure being
what the control measures rather than assumes away, non-zero on the finer grid for the reason just
marked open), whereas \qat{} hands it a code plane re-derived from a continuous, full-precision
intermediate.
\emph{The two tiers of exposure that creates} fall either side of the rule-family partition of
Section~\ref{sec:differentiators}, not either side of the bytes. \emph{Measured:}
export-tool divergence, in-family, costing about a point on the measured task. \emph{A sensitivity
bound, not an observed frequency:} driving the
weight-space artifact to ${\sim}0\%$ takes an out-of-family rule, and \textbf{of the five rules swept
only round-to-nearest is substantiated as shipping behaviour} (HuggingFace's $\lceil\log_2\rceil$
selection, the rule that serves gpt-oss, is bit-for-bit our RTN row), the other three spanning
deterministic conventions the format permits that we did not identify in any shipping stack applied to
4-bit weights at load. That is a \emph{different event} from the one the audit counts, so we read it
as a sensitivity bound rather than a frequency: the sweep establishes a property contrast on the rule
axis, not a lifecycle price, which is the audit's about-a-point plus the every-event requirement below.
\qat{} can be highly sensitive when its codes are re-derived under a mismatched rule; how
often that happens in deployment we do not measure.
\emph{Two parts are inferences, and one control we did not run.} \textbf{What a rule costs a
\emph{correctly} pre-materialized \qat{} artifact is an \emph{inference}, as is the scale arm's NVFP4
estimate above}: such an artifact is a stored native checkpoint, so a rule costs it what the zero-delta
control records for the base: not nothing, and not untouched. Its nearest support is the gpt-oss
pair above, and \textbf{a correctly pre-materialized \qat{} artifact swept under all five rules is a
control we did not run}.
\emph{What pre-materializing does not cover.} \textbf{Pre-materializing once with the exact
training rule is real but partial, and what it leaves open is not accuracy under a load-time
rule.} It covers the export event
(the verified safe transition above, $0$ differing bytes) and, at the base's own cost, the
load-time exposure with it. What remains is the requirement to re-run an exactly-matching
quantizer at \emph{every} later code-touching event, format conversion (class (b), which re-quantizes
both methods), and the code-plane payoffs (multi-adapter sharing, rollback, deduplication, audit),
which a re-derived code plane forfeits however carefully it was written. \qat{} is thus
\emph{accuracy-lossless when its training quantization convention is reproduced}, whereas \method{} is
\emph{bit-exact by construction} (in each case within a fixed native
format, scale grid, block layout and code plane), and the differentiator we rest on is that
every-event requirement plus the code-plane payoffs, not accuracy under one load-time rule.

\subsection{Payoff 2: multi-adapter serving on a shared code plane}
\label{sec:multiadapter}

Code-invariance turns $N$ task-merged models into one code plane plus $N$ scale planes, verified on
the four saved full-data \scaleqast{} adapters for Llama-8B NVFP4: across tasks, all
\textbf{224/224} target-linear code planes ($3.49$\,GB of uint8 codes) are \textbf{bitwise
identical} to the base and to each other ($0$ differing code bytes), and only $7.69\%$ of scale
bytes differ between tasks. \qat{}'s re-quantized merges change $7.4$--$8.0\%$ of \emph{code} bytes
per task, so its merged models share no code plane.

\paragraph{Storage, relative to $M$ independently merged weight-space checkpoints.} For $M{=}4$ tasks, scale-space stores one shared code plane plus four scale
deltas ($3328$\,MB $+\;4\times416$\,MB $=4992$\,MB) versus four full \qat{} checkpoints
($4\times3744$\,MB $=14{,}976$\,MB): \textbf{$3.0\times$ less}, with each per-task delta $416$\,MB
($\approx0.5$ bits/param). \textbf{The ratio is against weight-space \emph{merged} variants}: a base
plus $M$ runtime adapters stores less again, and we claim no storage advantage over it (below). Each
scale delta here is \emph{merged}, a stock native checkpoint with zero per-forward overhead.

\paragraph{Hot-swap, measured against a weight-space swap.} \emph{The first pair below times switching
the served task on an already-resident model; the second times producing a merged artifact once from
base plus adapter. Both are single runs, unlike the ten-repetition training-cost table.} Switching the
served task overwrites scale bytes with codes untouched:
\textbf{$0.059$\,s with a $3.5$\,MB transient} on one H200, versus $7.45$\,s and a $4552$\,MB
transient for a weight-space full re-quantization swap at equal resident memory,
\textbf{$\sim125\times$ faster with $\sim1300\times$ less transient memory}; against a runtime-adapter
server, or all variants held resident, we claim no swap advantage.
\textbf{The weight-space swap we time re-quantizes, and the nearest apples-to-apples swap is one we do
not time}: a practitioner who pre-materialized once holds one checkpoint resident with its siblings
already materialized and parsed, and switches task by copying a sibling's code plane into the resident
buffer, with no re-quantization at any point. We do not measure that configuration, so the ratio above
compares against a \emph{re-quantizing} swap and not that one; \textbf{we claim no latency advantage
over an all-resident pointer swap or a runtime-adapter system}. What survives that comparison is the
byte accounting rather than the timing: a scale swap writes only the scale plane, so the bytes moved and the
transient are properties of the format, not of our harness. Merge speed makes the scale side
viable: producing the artifact by writing
only the $\sim1/\text{block}$ scale bytes with no full-weight materialization is $18\times$
faster than the \qat{} merge on Llama-8B and $29\times$ on Qwen-30B (Table~\ref{tab:mergecost}), so a
merge-per-swap costs milliseconds instead of a dequantize--add--requantize--repack pass over every
code. The two ratios differ because the operations do: the in-place scale swap skips even the merge's
delta arithmetic, while the weight-space swap adds a repack-and-reload of the code plane on top of the
same re-quantization the weight-space merge performs. (In this research implementation both methods
keep bf16 code buffers, so \emph{resident}
memory is similar; the reliable signals are wall-clock time and the merge transient.)

\begin{table*}[t]
  \centering
  \footnotesize
  \caption{\textbf{Merge cost (scale vs \qat{}).} Scale merge overwrites only the
  $\sim1/\text{block}$ scale bytes ($0\%$ of codes change, no fp-weight peak); \qat{} dequantizes,
  adds, re-quantizes, and re-packs \emph{all} codes. Scale is $18\times$ faster on Llama-8B and
  $29\times$ on Qwen-30B.}
  \label{tab:mergecost}
  \begin{tabular}{@{}l cccc@{}}
    \toprule
    Model & Merge wall & Peak GPU & Transient & E2M1 codes changed \\
    \midrule
    llama-8B scale / \qat{} & \best{0.13} s / 2.33 s & 19 / 59 GB   & \best{1.7} / 30.5 GB & \best{0\%} / 100\% repacked \\
    qwen-30B scale / \qat{} & \best{0.20} s / 5.83 s & 66.7 / 67.6 GB & \best{0.56} / 8.1 GB & \best{0\%} / 100\% repacked \\
    \bottomrule
  \end{tabular}
\end{table*}

\paragraph{Serving correctness.} On the real vLLM native NVFP4 stack, each hot-swapped checkpoint
hits its task number on the full test set (Banking77 $93.77$, CLINC150 $97.71$), matching
Table~\ref{tab:main}: swapping scale bytes yields the correct task model. \emph{Provenance:} these
are single runs, as are the sweep's, whereas this model's Table~\ref{tab:main} entries are
$3$-seed means, the source of the second-decimal differences between the three figures, and why the
single-seed Qwen entries below instead reproduce their table numbers exactly.

\paragraph{The MoE replication.} On Qwen-30B NVFP4 ($192$ attention $+$ $192$ sampled expert
linears), the packed E2M1 code planes are byte-identical across all four tasks ($3.9$--$7.1\%$ of
scale bytes differ), while \qat{} flips $8.1$--$8.5\%$ of code bytes; storage for $M{=}4$ is
$21.4$\,GB vs $64.2$\,GB ($3.0\times$), and hot-swap is $6.8$\,s / $0.5$\,MB vs $98.4$\,s /
$647$\,MB ($\sim14\times$ faster, $\sim1300\times$ less transient). Each hot-swapped Qwen checkpoint
also reproduces its full-data table number exactly (Banking77 $93.93$, CLINC150 $96.33$;
$\max|\Delta W|{=}0$).

\paragraph{Alternatives we do not beat on every axis.} Three serving designs compete with merged
scale planes. \emph{Pre-materialized per-task native checkpoints} need no
merge machinery, but store $M$ full checkpoints ($14{,}976$\,MB at $M{=}4$ vs our $4992$\,MB) and face
a memory-for-latency trade of which we price one side. \textbf{A server holding all $M$ merged
checkpoints resident switches by pointer at essentially no latency, and we claim no swap-latency
advantage over that configuration}; its cost is $M\times$ weight memory, the storage axis this
subsection already prices. \emph{A third comparator, timed here for the first time}, is the one-resident
configuration that reloads a sibling checkpoint from disk rather than overwriting scale bytes in place:
a cold native NVFP4
checkpoint load measures $194$\,s ($178$\,s warm-cache) against our in-place scale-byte swap of
$0.059$\,s, a $\sim3000\times$ gap because the code plane is already resident and only the scale bytes
change. \textbf{Between that comparator and the pointer swap sits the configuration we do not time}:
one checkpoint resident, the sibling already materialized and parsed, its code plane overwritten in
place, paying neither re-quantization (as our timed weight-space denominator does) nor parse and
allocation (as the reload above does). \emph{Base-plus-runtime
adapters}, S-LoRA~\citep{sheng2023slora}, Punica~\citep{chen2023punica}, and
LoRA-Inlaid~\citep{xia2024lorainlaid}, which shares one quantized base across multiple runtime LoRA
task adapters as we share one code plane, avoid merging entirely but pay the adapter matmul at
runtime; our unmerged row prices that cost at $+5.6\%$ large-batch prefill latency and $+24\%$ low-batch
decode (Table~\ref{tab:deployable}), whereas a merged scale plane serves at zero per-task overhead.
\textbf{On storage and swap latency this design is the one to beat, and we do not}: a quantized base
plus $M$ small adapters is of the same order or smaller, and switching a resident pointer need not be
slower. Merged scale planes win on the axis we measure, zero per-forward overhead, plus the
artifact-level payoffs a runtime adapter never produces an artifact to have.
\emph{Copy-on-write and content-addressed stores} deduplicate whatever bytes coincide across
variants; our $64$\,KiB chunk-level dedup measurement (Section~\ref{sec:dedup}) \emph{is} that
scenario, and code-invariance is what makes it effective ($416$ vs $3740$\,MB per variant).

\subsection{Payoff 3: invertible merges and rollback}
\label{sec:rollback}

A scale merge is exactly reversible, and we verify the rollback empirically on Llama-8B NVFP4:
subtracting the on-grid scale delta the merge applied restores the base checkpoint byte-for-byte
($6.0$\,GB verified, zero residual bytes; the merge had touched only $4.7\%$ of scale bytes and no
code). One subtlety: the rollback delta must be the on-grid delta the merge applied, not the raw
training delta, since subtracting the latter leaves $0.61\%$ of scale bytes off. A \qat{} merge
destroys information: it re-derives $7.4\%$ of packed code bytes, re-quantization reassigns codes
many-to-one, so the base is constructively unrecoverable from the merged artifact. The honest
alternative works for either method (retain the base checkpoint plus the small adapter and re-merge
on rollback), so our contribution is that rollback needs no retained base: the merged artifact plus
the on-grid scale delta the merge applied (a $1.7$\,GB artifact if stored) restores the base
bit-exactly, where the alternative keeps a full second checkpoint and re-runs a merge.

\subsection{Payoff 4: storage deduplication and patch bandwidth}
\label{sec:dedup}

Every scale-merged variant shares its code pages with the base, and we measure what that buys over
the four Llama-8B task checkpoints against the same $5.7$\,GB base. The raw differing bytes are
$23.5$\,MB (scale plane only) vs $254$\,MB spread across the code plane for \qat{}. A zstd
\texttt{-{}-patch-from} compressed delta ships a scale-merged variant in $158$\,MB vs $3335$\,MB for
\qat{} ($21.1\times$): $2.7\%$ of a full checkpoint vs $58\%$. A $64$\,KiB chunk-level dedup patch,
the content-addressable-store approximation, is $416$\,MB vs $3740$\,MB ($9.0\times$), and edge or
over-the-air distribution inherits the same ratios. $N$ \qat{}-merged variants share nothing with the
base or each other: each re-derived code plane differs in $7.4$--$8.0\%$ of code bytes spread across
every block, so both deduplication and delta-compression degrade toward full-checkpoint cost.

\subsection{Payoff 5: verifiability and provenance}
\label{sec:audit}

\textbf{This payoff is a corollary of the code-plane identity already measured in payoff~(2), not an
independently measured one.} With codes frozen, a merged artifact is auditable against its base: the byte diff \emph{must} be
confined to the scale plane, and any code-byte difference is evidence of tampering or a corrupted
pipeline. Our multi-adapter verification is that audit ($224/224$ code planes byte-identical,
Section~\ref{sec:multiadapter}), and it runs as a byte comparison with no model execution. A
\qat{}-merged artifact diffs from its base across the code plane by construction ($7.4$--$8.0\%$ of
code bytes), so no such invariant exists and provenance reduces to trusting the merge pipeline.

\subsection{Scope and confounds of the sweep}
\label{app:sweepscope}

The deployment-quantizer sweep
(Section~\ref{sec:deploysweep}) runs on the real vLLM NVFP4 stack only for Llama-8B; the Qwen-30B
MoE replication uses a single-backend torch re-quantization harness on the same adapters and full test
sets, where stochastic rounding \emph{degrades} \qat{} rather than driving it to exactly $0$. At 120B we
measure a single RTN-vs-RTZ pair on one dataset (CLINC150, logprob bridge, never vLLM) with both
confound controls; stochastic and whole-tensor rules remain unmeasured at that scale. The whole-tensor
rule is confounded throughout (a granularity change rather than a rounding change, it collapses even
the un-adapted base), so RTZ and stochastic are the clean isolators; and the Spider sweep is incomplete
with a near-saturated base, so the
sweep's headline numbers come from the three classification datasets. Merge-loss discipline is
single-backend throughout (Section~\ref{sec:limitations}): a cell's merged and unmerged numbers come from
the same backend and losslessness is established in weight space, so the observed zeros are an identity
rather than a coincidence of two noisy accuracies.
\textbf{Its remaining three limits (that it is asymmetric by design, that it prices no lifecycle
event, and that two halves of it are inferences rather than measured rows) are stated in full in
Section~\ref{sec:deploysweep} and we do not re-argue them here.} For the \qat{}
generative eval we scored the merged model once and verified that
re-merging is idempotent, rather than re-running the unmerged eval.

\section{Why the Naive Merge Returns the Base: the Three Tests in Full}
\label{app:deletion}

This appendix gives the three tests summarized in Section~\ref{sec:deletion} at full length: the
intervention that reverses the effect, and the two pre-registered predictions. The Claim is stated where
it is used, in Section~\ref{sec:deletion}; three refinements of it belong here, the first of them a
correction to the form in which we previously stated Case~2. \emph{Case~1 is exact
element-wise}: given premise (iii), an on-grid value is a fixed point of nearest-rounding, so
re-quantizing $W+\Delta$ returns $W$ on \emph{every} sub-step element. \emph{And Case~2 assumes nothing
about scale re-selection}, which is why it covers the weight-MSE search on \emph{both} grids where
Case~1 covers only the finer, mantissa-bearing one: the grid on which the scale byte is stable under
a sub-step update, evidenced for premises (ii) and (iii) jointly by the percent-level displacement
below.

\paragraph{Case~2 at block level: the Voronoi condition, and why element-wise sub-step is not enough.}
Case~2 concerns a quantizer that minimizes block reconstruction error over \emph{both} the scale and the
codes, so its optimality question is a nearest-neighbour question in the block codebook
$\mathcal{Q} = \{\,s\,c : s \in \mathcal{S},\, c \in \mathcal{C}^{b}\,\}$, not a set of independent
single-coordinate tests. For an on-grid base $W \in \mathcal{Q}$, target $T = W + \Delta$ and any
competitor $Q' \in \mathcal{Q}$ with $D = Q' - W$, expanding
$\lVert\Delta\rVert^2 \le \lVert\Delta - D\rVert^2$ gives the exact condition for the base to remain the
minimizer:
\begin{equation*}
  \langle \Delta, D\rangle \;\le\; \tfrac12 \lVert D \rVert^{2}
  \qquad \text{for every } Q' \neq W ,
\end{equation*}
that is, $\Delta$ lies in the Voronoi cell of $W$ in the full microscaling block codebook. By
Cauchy--Schwarz a sufficient condition is $\lVert\Delta\rVert \le \tfrac12 d_{\min}(W)$ with
$d_{\min}(W) = \min_{Q' \neq W} \lVert Q' - W\rVert$, half the packing radius at $W$. The natural
per-block diagnostic is therefore
$m(W,\Delta) = \min_{Q' \neq W} \left[\tfrac12\lVert Q'-W\rVert^{2} - \langle\Delta, Q'-W\rangle\right]$,
positive exactly when the un-adapted block is still the reconstruction optimum.

\emph{Element-wise sub-step is strictly weaker than this, and we record the gap rather than gloss it.}
Take the two-element E2M1 block $W = (4, 1)$ at scale $s = 1$, i.e.\ codes $c = (4,1)$. The distances from
its coordinates to their nearest rounding midpoints are $(0.5, 0.25)$, so
$\Delta = (-0.475, -0.2375)$ is sub-step in \emph{every} coordinate. Yet returning the base costs
$\lVert T - W\rVert^{2} = 0.28203$, while the competitor at $s' = 0.5$ with codes $c' = (6, 1.5)$, i.e.\
$Q' = (3, 0.75)$, costs only $0.27578$, and an exhaustive search over power-of-two scales and E2M1 code
pairs confirms $Q'$ is the joint optimum. Both scales are representable on E8M0 and on E4M3, so the
counterexample is realizable on either grid. The reason is geometric: the element-wise box here admits
$\lVert D\rVert$ up to $0.559$ while $d_{\min}(W) = 0.5$, so the box \emph{protrudes} from the Voronoi
cell, and a joint scale-and-code move exploits the protrusion.

\emph{What this does and does not cost the Claim.} It costs the sufficiency of the element-wise premise
for Case~2, which we have restated at block level accordingly; it does not touch Case~1, where the scale
is held fixed by premise (iii) and the element-wise argument is exact, nor any measurement. The
counterexample needs $\Delta$ at $95\%$ of the element-wise threshold, whereas the regime we measure sits
two orders of magnitude inside it ($45$--$200\times$ below the midpoint distance, every element
sub-threshold in $>99.5\%$ of blocks). More to the point, the conclusion of Case~2 is \emph{measured} rather than deduced: the
weight-MSE search lands its scales $7.5\mathrm{e}{-}5$ from the base at an attained reconstruction error of
$\lVert\Delta\rVert$ itself, and the merged accuracy lands inside the un-adapted base's Wilson interval.
The criterion is computable exactly and we measure it per block below.

\paragraph{Computing the criterion exactly, and measuring it.} The minimisation over representable
competitors collapses to one dimension. At a \emph{fixed} scale $s'$ the reconstruction-minimising codes
are coordinatewise $\operatorname{round}(T/s')$, so each representable scale contributes exactly one
optimal competitor and a scan of the finite scale grid ($126$ values for E4M3, $61$ exponents for E8M0) is
\emph{exhaustive} over the codebook. Two cautions matter in practice. The base itself arises in that scan,
at its own scale, so it must be excluded or $\text{sse}_{\text{best}}\le\text{sse}_{\text{base}}$ holds by
construction and the margin can never be positive. And $d_{\min}$ is a \emph{different} quantity that the
scan cannot see: the nearest neighbour of $W$ is a one-code-step move at $W$'s own scale, never a rounding
of $W$, so any local enumeration bounds $d_{\min}$ from \emph{above} and the Cauchy--Schwarz radius
$\lVert\Delta\rVert\le\tfrac12 d_{\min}$ is safe only with the global value. At $0.98$ of a
locally-computed radius, $60$ of $200$ random updates on E4M3 were beaten by a competitor the local
enumeration never saw, against $0$ of $200$ on the coarser E8M0 grid.

Measured on Llama-8B NVFP4 with the full-data \qat{} r32 adapter, over $140{,}000$ blocks in $35$ layers
and from weights alone: the un-adapted block is the \textbf{global} reconstruction optimum on
$\mathbf{99.4\%}$ of blocks. On a $28{,}000$-block subset we also record the size of that margin: keeping
the base costs a median $0.117$ of the RMSE of the nearest \emph{distinct} representable block, so the base
wins by roughly $8\times$ rather than marginally. Premise (iii) is measurable directly, and is the part of the
Claim that is not definitional: a reconstruction \emph{minimiser} agrees with the Case-2 margin by
construction, since both are the arg\,min, whereas a fixed rule can disagree. The $\text{amax}/6$
nearest-rounding rule re-derives the \emph{same} block scale under the trained update on
$\mathbf{99.9\%}$ of blocks and returns the block exactly to base on $\mathbf{99.3\%}$, the two cases
disagreeing on $0.075\%$. \textbf{This analysis must be carried out in weight space.} The base is a fixed
point of the $\text{amax}/6$ rule on only $16.4\%$ of blocks, so the checkpoint's stored scales were
selected by a different rule, yet the reconstruction still returns the base \emph{weight vector} on
$99.3\%$: the same vector is reachable at several (scale, code) pairs and the degeneracy absorbs the
difference. Comparing stored scales instead of reconstructed weights would report premise (iii) as failing
on $83.6\%$ of blocks, inverting the conclusion.

\paragraph{The NVFP4 codebook is conditional; the MXFP4 one is not.} NVFP4 stores a block scale as an E4M3
byte times a per-tensor fp32, so the competitor set is fixed only when a deploying tool preserves that
fp32. Everything above holds it fixed. When it is recomputed, which our own MSE quantizer does from the
amax of the merged tensor, $21\%$ of blocks acquire a strictly better competitor and the geometry no longer
certifies deletion. MXFP4/E8M0 carries no per-tensor factor, so its codebook is genuinely fixed. The
packing radius differs in kind for the same reason: on E8M0 scale factors out of a power-of-two grid and
$d_{\min}=s\,f(c)$ depends on the code pattern alone, whereas on E4M3 it is $s\,f(c,m)$ and also depends on
where the scale sits within its binade ($0.2577$ at $s{=}1$ against $0.3436$ at $s{=}24$ for identical
codes), so \textbf{there is no single packing radius for NVFP4}. 

\paragraph{Both grids, and why one quantizer deletes while the other escapes.} We repeat the measurement
on a \emph{native} MXFP4 checkpoint (gpt-oss-120B, $120{,}000$ blocks over $40$ expert cells, reading the
packed \texttt{blocks}/\texttt{scales} tensors directly). Simulating this grid on an NVFP4 checkpoint does
not work, because rounding E4M3 scales to powers of two damages a base that was not built on that grid and
confounds that damage with the effect on the update; a native checkpoint has power-of-two base scales by
construction, which the measurement confirms (the stored scale equals the ceiling rule's choice on
$100\%$ of blocks, against $16.4\%$ on NVFP4, where a different selection rule was used).

\begin{center}\footnotesize
\setlength{\tabcolsep}{4pt}
\begin{tabular}{@{}lrr@{}}
\toprule
median over cells & E4M3 & E8M0 \\
\midrule
update element-wise sub-step & $>99.5$ & $100$ \\
premise (iii) holds & $99.9$ & $\mathbf{68.2}$ \\
fixed rule returns to base & $99.3$ & $\mathbf{84.5}$ \\
Case-2: base is \emph{global} optimum & $99.4$ & $\mathbf{100}$ \\
Case-2 optimal, rule moved it & $0.08$ & $\mathbf{15.5}$ \\
base's RMSE margin & $9\times$ & $\mathbf{1163\times}$ \\
\bottomrule
\end{tabular}\\[2pt]
{\footnotesize All columns are percentages of blocks except the last row.}
\end{center}

\noindent This separates the two cases quantitatively and on the grid the $100$B$+$ models actually use.
A reconstruction \emph{minimiser} finds the un-adapted block optimal on \emph{every} E8M0 block, and by a
margin three orders of magnitude wide, because the coarse grid puts the nearest competitor far away: such a
merge deletes the adapter with certainty, which is the weight-MSE behaviour of
Table~\ref{tab:ptqbaselines}. The \emph{fixed} $\lceil\cdot\rceil$ rule is not a minimiser, and premise
(iii) fails for it on $31.8\%$ of blocks even though the update is element-wise sub-step on all of them, so
it moves $15.5\%$ of blocks off the base and retains the adaptation. \textbf{The same geometry therefore
predicts that MSE deletes and RTN survives on E8M0, and that both delete on NVFP4}, matching the measured
retentions. The $31.8\%$ figure also reproduces, from weights alone and through an independent code path,
the $30.2\%$ exponent-bump census reported above.

\paragraph{The three tests, as originally reported.}
\emph{Independent test 1, an intervention that reverses the effect: the objective, not the grid, is the
cause.} Three observations rule out ``the
$4$-bit grid is just too coarse.'' On gpt-oss AGNews, AWQ attains a \emph{worse} reconstruction error
than RTN when aggregated over that cell's matrices
($7.85\mathrm{e}{-}2$ vs $4.45\mathrm{e}{-}2$, per-matrix max $0.79$) at statistically identical
accuracy; that aggregate is tail-driven, as the per-matrix max shows, and the \emph{median} on the same
cell is level between the two rules (quoted in the AWQ sub-investigation, Section~\ref{sec:awqmech}). The argument needs only
that AWQ is not a \emph{better} reconstructor than RTN, which both statistics support. GPTQ (which
does error-feedback on a fixed grid rather than minimizing per-block weight SSE over scales, so its
optimum is \emph{not} the base) is the one post-hoc quantizer that rescues, and it does so while
carrying the \emph{largest} per-matrix reconstruction error of the four on gpt-oss AGNews
($6.5\mathrm{e}{-}2$ median, against MSE's best-in-class $1.3\mathrm{e}{-}3$). And running GPTQ on top
of the MSE grid recovers that cell from $75.6$ to $92.1$, reclaiming $16.5\pp$ from the very scales
that had deleted the adapter. \textbf{Read for what it localizes, that one experiment puts the failure
in scale selection alone}: recovering that cell from $75.6$ to $92.1$ is a merge loss of only
$-0.3\pp$ and $+16.5\pp$ reclaimed, so error-feedback over \emph{codes} undoes the deletion; the cause is the
scale-selection objective and not the FP4 grid nor calibrated quantization in general, which is also
why plain GPTQ escapes. \textbf{Minimizing weight
reconstruction error is thus the wrong objective for a merge}: across these four quantizers the best
reconstructor is the worst model and the worst reconstructor the best one, so lower weight error does
not buy (and here actively costs) retained adaptation. (The relationship is directional, not
strictly monotone: RTN and AWQ swap order between the two rankings, their separations on both axes
being smaller than the corresponding floors.)
An independent replication on DeepSeek-V4 AGNews traces the same relation across an order of magnitude of
reconstruction quality, all five quantizers measured in the identical configuration against a single
$90.4$ unmerged reference: aggregate reconstruction error $4.48\mathrm{e}{-}3 \to -59.8\pp$,
$9.03\mathrm{e}{-}3 \to -14.0\pp$, $3.90\mathrm{e}{-}2 \to -3.6\pp$, $4.69\mathrm{e}{-}2 \to -2.0\pp$
(MSE, gptq\_mse, RTN, GPTQ respectively): monotone over those four and spanning the full range from
near-total deletion to none, with the fifth, AWQ on AGNews, breaking the ordering: it is the single
exception across every cell we measured, and we discuss it below rather than absorb it into the
ordering. The best reconstructor lands at $30.6$ against an un-adapted base of $33.6$: it does not
merely lose the adapter, it returns the base.

\paragraph{Reading Table~\ref{tab:retention}.} Each retention figure pairs a merged accuracy with its
own seed's unmerged accuracy and is then averaged over seeds (Llama Banking77/CLINC150 are 3-seed, the
rest single-seed), and every row draws base, unmerged and merged from one eval configuration. The
exception the table flags is AWQ on DeepSeek AGNews ($81\%$ against RTN's $94\%$), the one departure we
could not attribute to reconstruction quality. The italicised Spider rows are excluded by a
\emph{declared convention} rather than a threshold derived from the data: once headroom is small, a
numerator error of one noise floor becomes tens of points of retention, so we quote the ratio only where
headroom is large compared with this metric's run-to-run floor. At $8.0$, $2.6$ and $1.2\pp$ none of the
three clears that bar, and the instability shows in the readings themselves (Qwen Spider $170\%$, gpt-oss
Spider $125\%$); DeepSeek Spider at $31.2\pp$ does clear it and is included. All four DeepSeek rows use
the default capacity (token-dropping) in-EP setting, matching their Table~\ref{tab:main} entries; the
dropless variant of DeepSeek AGNews gives $97\%$ rather than $94\%$, so retention does not turn on that
choice, whereas the \emph{headroom} column does: capacity dropping costs the un-adapted AGNews base
$27\pp$, inflating the measured gain, which is why that column is an upper bound.

A fifth quantizer, \emph{gptq\_mse} (GPTQ run over the weight-MSE scale grid), has no column in Table~\ref{tab:retention} because it was never run on the NVFP4 models, so we report its eight E8M0 retentions here: gpt-oss $101/98/90\%$ and DeepSeek $75/98/100/105\%$ on AGNews/Banking77/CLINC150(/Spider), with gpt-oss Spider's $255\%$ sub-noise for the same reason as that row's other entries. It recovers what the scale search discarded on seven of the eight cells; DeepSeek AGNews is the partial one.

Banking77 is consistent with this but does not test it: there MSE at
$4.27\mathrm{e}{-}3$ loses $38.8\pp$ while the four quantizers above $1.2\mathrm{e}{-}2$ lose only
$0.6$--$2.2\pp$ and are \emph{mutually indistinguishable} at $n{=}500$ (the $1.4\pp$ AWQ-to-GPTQ gap
against a $\pm2.8\pp$ difference floor) despite a $12\times$ spread in reconstruction error, so that
cell separates MSE from everything else and orders nothing within the remainder.

Reconstruction error alone is not, however, a sufficient statistic for merge loss in raw percentage
points, and one pair rules that out cleanly: DeepSeek gptq\_mse reaches $9.03\mathrm{e}{-}3$ on AGNews and
$8.90\mathrm{e}{-}3$ on CLINC150 (the same reconstruction quality to two significant figures), yet
loses $14.0\pp$ on the first and $0.0\pp$ on the second. What differs is not the quantizer but how much
adaptation each task had to lose. Normalising by that (the retention statistic of
Table~\ref{tab:retention}) is what makes the picture stable, and it yields the sharper invariant:
\textbf{a weight-MSE scale search destroys $105$--$110\%$ of the adaptation gain on every DeepSeek dataset}
($105.3$, $109.6$, $110.3\%$), i.e.\ it lands at or just below the un-adapted base, exactly as the
argument above predicts for the SSE-optimal reconstruction of an on-grid base. The raw figures
($-59.8$, $-38.8$, $-36.4\pp$) mostly track each task's headroom: AGNews looks the most dramatic because
it has $56.8\pp$ of gain to destroy, while CLINC150 has $33.0\pp$ and sits at a $99\%$ ceiling. That
AGNews headroom is itself configuration-dependent, which is the second reason we normalize: it is
measured with expert-parallel capacity dropping on, the setting these cells were evaluated in, and
with dropping disabled the same cell reads $32.6\pp$ of headroom instead
(Table~\ref{tab:retention}), while its retention barely moves. The failure is not that $4$-bit cannot
\emph{hold} the update (it is sub-step by two orders of magnitude, and RTN on the identical grid keeps
almost all of it) but that a reconstruction objective cannot \emph{preserve} a sub-step update.
Correcting the block \emph{scale} never poses that objective: \scaleqast{} writes an already-on-grid
trained scale rather than reconstructing $W+\Delta$ post hoc, so it is not subject to this
reconstruction-induced deletion mechanism when merged onto the grid it trained for.

\textbf{Independent test 2, a pre-registered prediction with a stated falsifier on a second model: the
effect is not classification-specific, and it replicates.} On
DeepSeek-V4-Flash Spider (a \emph{generative} SQL task on a second, independently built native-FP4
checkpoint, evaluated on the authoritative device-map bridge), the un-adapted 4-bit base scores $40.4$
and naive fine-tuning reaches $71.6$ unmerged, i.e.\ $31.2\pp$ of headroom. RTN merge keeps essentially
all of it ($72.2$, $+0.6\pp$, $101.9\%$ retention). The weight-MSE scale search lands at $38.4$
(inside the base's Wilson interval, $-33.2\pp$ merge loss, $-6.4\%$ retention), a $33.8\pp$ separation
from RTN on the identical adapter and eval path. The search moved $11.5\%$ of blocks off the RTN exponent
here (vs $15.0\%$ on gpt-oss), so this is the mechanism operating, not a degenerate search that silently
reproduced RTN. This prediction and its falsifier (${>}60$ would refute it) were fixed in writing before
the cell ran.

\textbf{This is one mechanism with one exception, not two mechanisms; the NVFP4 collapse is
covered by Case~1 of the Claim, the fixed-rounding case, not by the scale-search case.} The
headroom-normalized signature that shows this across cells, the rule by which sub-headroom rows are
excluded from its ranges, and the ranges themselves are stated once in Section~\ref{sec:deletion} with
Table~\ref{tab:retention}, and we do not restate them here. The one reading this appendix adds is that
MSE's $3$--$9\%$ on the gpt-oss E8M0 cells with substantial headroom is the \emph{same} signature the
NVFP4 rounding rules show, on the other grid.

\textbf{Independent test 3, a pre-registered weight statistic whose content is the size of the
separation, not its sign.} \textbf{We state what the pre-registration does and does not buy before
quoting the number.} The finer grid's displacement is \emph{entailed} by premises (i) and (ii) as the
same probe measures them: an on-grid base met by a fixed nearest-rounding rule, with an update two
orders of magnitude below its own rounding step, can be displaced only negligibly, so the refuting
value could have been reached only if one of those two already-measured premises had failed. The
power-of-two grid's value, moreover, had been measured before the prediction was written down, so the
pre-registration binds one side of the comparison and the other side is a prior measurement, a
retrodiction of an unmeasured weight statistic rather than a forecast of an accuracy outcome. What the
test genuinely buys is the \emph{magnitude} of the separation and its co-location with retention on
both grids. Its force is that the refuting range was fixed in writing first (a separation under
${\sim}3\times$ would have refuted the unification outright rather than merely scoped it) and that
the probe then ran offline and weights-only on the native NVFP4 checkpoint, with no eval, no training
and no model instantiation. It came back inside the predicted range. The displacement a merge produces from the un-adapted base, in units of the adapter,
$\|Q(W{+}\Delta)-W\|/\|\Delta\|$, is $0.010$ for NVFP4 RTN and $0.009$ for NVFP4 MSE, versus
$12.0$--$12.3$ for E8M0 RTN (medians over $16$ sampled \texttt{down\_proj} target linears of the dense
NVFP4 checkpoint carrying its banking-intent naive adapter, and over one E8M0 layer's experts,
respectively): a ${\sim}1000\times$ separation that tracks retention across both grids.
\textbf{The two comparison groups are unequal in extent and the surviving one is far the smaller}: sixteen sampled
linears of one model against a single layer's experts of one model, so the separation must not be read
with the breadth of the sixteen-cell retention table whose pattern it tracks. The
secondary prediction, that the two rounding-only rules on the finer grid would cluster with each
other, holds as well.
The fourth cell of that $2{\times}2$, the deleting rule on the surviving grid, is the
power-of-two-grid weight-MSE search above, already in these
units: the fraction of $\Delta$ that search recovers \emph{is} $\|Q(W{+}\Delta)-W\|/\|\Delta\|$ there.
\textbf{The two quantities coincide only when the displacement lies along the update},
which holds by construction for a search returning a rescaled base and is, for the surviving rule,
the same inference we flagged above rather than a measurement, though the conclusion the test needs,
the thousandfold separation between the three deleting rules and the one surviving rule, does not
depend on it.
\textbf{The two instruments agree in kind but not in magnitude.} Section~\ref{sec:deletion} grades the
deletion claim by which of them supports it (literal, on both instruments, wherever the merged model
lands inside the un-adapted base's own interval, and \emph{open} on the NVFP4 rounding cells, where the
weight probe reads near-total deletion and accuracy keeps a minority of the gain that is nonetheless far
more than the weight residual predicts), and we do not restate that grading here. Three things belong
to the probe rather than to the grading. This mechanism does not
account for that NVFP4 gap; the sub-step fraction of the premise sets a floor under it but not its size. Only
the power-of-two-grid weight-MSE case above is quoted as a proportional correspondence, and the probe
is a subsample of the target linears on one arm and a single layer on the other, not a census. What this test needs,
and what both instruments support, is the thousand-fold separation between the deleting and the
surviving rules, not a map from displacement onto retention.
\emph{A second item stands open in the same register}: where the adaptation lives under the surviving
rule (necessarily in the bumped minority of blocks, per the Claim's own premises) is the
distribution flagged above and not measured.
Two exact signatures accompany it on NVFP4, as on E8M0: the native base is on-grid to $0.00000$, and the
achieved reconstruction error equals $\|\Delta\|$ to five decimals, i.e.\ the quantizer attains precisely
the ``return the base'' solution. The sub-step premise also holds symmetrically: on the same $16$ NVFP4
linears the median $|\Delta_i|$ is $0.018\times$ the distance to its E4M3 rounding midpoint (E8M0:
$0.005$--$0.022\times$), so the update survives nearest-rounding for a vanishing fraction of elements on
both grids; the census is not asserted on one grid and inferred on the other. Section~\ref{sec:deletion}
draws the consequence, that grid coarseness is not an independent cause of the NVFP4 collapse but acts
through the rounding rule's reconstruction quality.
The one exception to that reading is the AWQ cell flagged in test~1, whose mechanism
Section~\ref{sec:awqmech} takes up.

\paragraph{The grid axis, in full: what is confounded with what.}
Both MXFP4 models are the two largest MoEs, scored through a logprob bridge, while
the NVFP4 cells are a dense model on vLLM generation and a smaller MoE on an HF one; grid, model
scale and scoring path move together. Qwen's NVFP4 MoE controls MoE-ness within that grid, and we quote
headroom-normalized retention rather than raw points because the ratio is insensitive to harness level.
\textbf{We also broke the grid axis directly.} Holding the model, task, rank, learning rate,
early-stopping rule and eval path fixed on Llama-8B and varying \emph{only} the QAST scale grid, the
attainable accuracy is essentially unchanged: Banking77 reaches $93.75\pm0.09$ over three seeds on E4M3
against $92.88\pm0.81$ on the coarse E8M0 grid ($-0.88\pp$, Welch $p\approx0.20$), and CLINC150 $97.91$
against $97.22$. The control is what makes this readable: an NVFP4 base is not on the power-of-two grid,
so re-gridding it alone rewrites $99.73\%$ of stored scales and costs the \emph{base} $19.2$ and
$27.5\pp$ on the two tasks. Comparing raw accuracies would therefore have shown a large spurious grid
effect; comparing each grid's gain over its own base shows scale adaptation recovering essentially all of
that damage ($+61.3$ and $+64.1\pp$, against $+42.9$ and $+37.3$ on E4M3). \textbf{A coarser scale grid
costs about a point of reachable accuracy at 8B, so grid coarseness is not the mechanism behind the
larger-model Spider gaps}; that reading is not supported by this control, and the merge stayed exactly
lossless ($0.00\pp$) in all four arms, extending the exactness result to a second grid on a dense model.

\textbf{Two further instruments narrow the confound, in different ways.} The
offline displacement predictor of Section~\ref{sec:deletion} removes the \emph{eval}: it is weights-only,
with no scoring, no task and no model instantiation, so harness level and task drop out entirely, but
its two comparison groups are sixteen linears of one dense NVFP4 model against one layer's experts of one MXFP4 MoE,
so grid, model and dense-vs-MoE architecture still move together inside it. The comparison that holds
everything else fixed is instead the weight-MSE-versus-RTN pair on the same gpt-oss checkpoint
(Section~\ref{sec:deletion}): one model, one architecture, one grid, one adapter, one harness, and only
the rounding rule changes. The mildness on E8M0 is \emph{not} explained by a
saturated base or by an adapter that barely trained: gpt-oss-120B CLINC150 rises $67.5\to96.9$ unmerged
and naive merge still retains $+26.8\pp$ of that gain, as Banking77 rises $73.9\to91.7$ and retains
$+17.0\pp$, so a practitioner merging naively into these particular checkpoints would lose little. The one
E8M0 cell that \emph{is} saturated is gpt-oss Spider (base $71.3$, unmerged $72.4$, $1.16\pp$ of
headroom), where naive merge lands $0.3\pp$ \emph{below} not fine-tuning at all, a
merge loss small in pp only because there was nothing left to lose.

\paragraph{Where the claim is literal, and the instrument disagreement, in full.} \emph{Deletion is
literal, and both instruments say so}, wherever the merged model lands inside the un-adapted base's own
interval: the weight-MSE cells on the power-of-two grid, and the Qwen banking-intent cell, which merges
to the base it started from. \emph{On the NVFP4 rounding cells the two instruments disagree by roughly an
order of magnitude, and we record that as open rather than resolve it}: the weight probe reads near-total
deletion (a percent-level fraction of the adapter surviving) yet accuracy still leaves several points of
gain above base on the Llama cells of Table~\ref{tab:retention}, far more than the weight residual
predicts, and this mechanism does not account for that gap. Two exact signatures nonetheless accompany
the probe on both grids (base on-grid to $0.00000$; achieved reconstruction error equal to $\|\Delta\|$ to
five decimals), and the sub-step premise holds symmetrically on both, so the census is not asserted on
one grid and inferred on the other. Grid coarseness is therefore \emph{not} an independent cause of the
NVFP4 collapse: it acts \emph{through} the rounding rule's reconstruction quality: E8M0 RTN preserves
adaptation only because it is a \emph{poor} reconstructor. The same signature runs across cells
(Table~\ref{tab:retention}; single-seed except the two $3$-seed Llama cells, so a regularity rather than a
fitted law): on the NVFP4 cells whose headroom is large against the
protocol's noise floor, so that the retention ratio's own propagated uncertainty stays small (the
declared convention by which Table~\ref{tab:retention} italicises and excludes its Spider rows, here the
six above $13\pp$), \emph{every} rounding-only quantizer
retains only $-1$ to $24\%$ of the fine-tuning gain, RTN, MSE and AWQ agreeing with each other inside a
cell far more closely than they differ across the two grids, and the outlier is E8M0's
$\lceil\cdot\rceil$ rule at $86$--$102\%$.

\begin{table*}[t]
  \centering
  \footnotesize
  \caption{\textbf{Headroom retention} $(\text{merged}-\text{base})/(\text{unmerged}-\text{base})$
  \textbf{after a naive merge, per post-hoc quantizer.} Every rounding-only quantizer on NVFP4 discards
  the fine-tuning gain ($-1$ to $24\%$), as does MSE on E8M0 ($-10$ to $9\%$). The exception is E8M0's
  $\lceil\cdot\rceil$ rule ($86$--$102\%$), which preserves adaptation because it is a \emph{poor}
  reconstructor; GPTQ escapes on both grids. \emph{Italicised} Spider rows are excluded from these ranges:
  their headroom is too small for the ratio to be stable. The DeepSeek \emph{headroom} column is an upper
  bound. Provenance, seeding, the exclusion criterion, the AWQ exception and the \texttt{gptq\_mse}
  retentions are in Appendix~\ref{app:deletion}.}
  \label{tab:retention}
  \begin{tabular}{@{}ll r rrrr@{}}
    \toprule
    Grid & Cell & Headroom & RTN & MSE & AWQ & GPTQ \\
    \midrule
    \multirow{8}{*}{\shortstack[l]{NVFP4\\(E4M3)}}
      & Llama Banking77  & 43.8 & 15\% & 13\% & 13\% & \best{94\%} \\
      & Llama AGNews     & 17.4 & 24\% & 24\% & 24\% & \best{91\%} \\
      & Llama CLINC150   & 37.5 & 22\% & 20\% & 19\% & \best{98\%} \\
      & Qwen Banking77   & 23.1 & \bad{$-1$\%} & 0\% & 2\% & \best{96\%} \\
      & Qwen AGNews      & 13.1 & 15\% & 14\% & 17\% & \best{98\%} \\
      & Qwen CLINC150    & 16.5 & 15\% & 15\% & 20\% & \best{98\%} \\
      & \textit{Llama Spider} & \textit{8.0} & \textit{$-20$\%} & \textit{$-17$\%} & \textit{$-18$\%} & \textit{93\%} \\
      & \textit{Qwen Spider}  & \textit{2.6} & \textit{$-26$\%} & \textit{$-26$\%} & \textit{11\%} & \textit{170\%} \\
    \midrule
    \multirow{8}{*}{\shortstack[l]{MXFP4\\(E8M0)}}
      & gpt-oss AGNews    & 18.5 & \best{86\%} & \bad{9\%} & \best{86\%} & \best{97\%} \\
      & gpt-oss Banking77 & 17.8 & \best{96\%} & \bad{3\%} & \best{101\%} & \best{100\%} \\
      & gpt-oss CLINC150  & 29.4 & \best{91\%} & \bad{6\%} & \best{90\%} & \best{98\%} \\
      & \textit{gpt-oss Spider} & \textit{1.2} & \textit{$-25$\%} & \textit{0\%} & \textit{125\%} & \textit{100\%} \\
      & DeepSeek AGNews   & 56.8 & \best{94\%} & \bad{$-5$\%}  & 81\%           & \best{96\%} \\
      & DeepSeek Banking77 & 35.4 & \best{94\%} & \bad{$-10$\%} & \best{94\%}  & \best{98\%} \\
      & DeepSeek CLINC150 & 33.0 & \best{98\%} & \bad{$-10$\%} & \best{100\%} & \best{100\%} \\
      & DeepSeek Spider   & 31.2 & \best{102\%} & \bad{$-6$\%}  & \best{96\%}    & \best{102\%} \\
    \midrule
    \multicolumn{3}{@{}l}{\emph{Displacement from base,} $\|Q(W{+}\Delta)-W\|/\|\Delta\|$}
      & \multicolumn{2}{c}{$0.010$ / $0.009$ (NVFP4)} & \multicolumn{2}{r}{$12.0$--$12.3$ (E8M0 RTN)} \\
    \bottomrule
  \end{tabular}
\end{table*}

Scope, case by case: Case~2 concerns quantizers selecting a per-block scale for a fixed code grid over
the scale range such methods actually search (we claim no global optimum over all conceivable scales,
and a finer grid could in principle attain $\varepsilon<\|\Delta\|$), while Case~1 assumes no search at
all, which is why it and not the search argument is what the rounding-rule cells rest on. Both cases
conclude ``returns the base up to the sub-step fraction'' rather than exactly: elements that do survive
nearest-rounding, and blocks in which some element is above threshold, are not returned to the base.

\section{Matched Parameters, Training Cost, and Deployment Options}
\label{app:cost}

\paragraph{Llama at matched params: a tie, not a win.} Against the merge-aware baseline on Llama-8B
(scale r58 $\approx$ qlora r32), the two merge-aware methods are tied (mean $88.7$ vs
$89.0$; Table~\ref{tab:llamamatch}); \scaleqast{} reads higher on the two coarse classification tasks by
$<1\pp$, and the single-run Spider pair shows \qat{} ahead by $2.1\pp$, but seven repetitions per method
reverse that to \scaleqast{} $+0.50\pp$ (CI $[-0.55,+1.55]$), and an earlier $+4.95\pp$ \scaleqast{} win
from a pilot run likewise does not reproduce.

\begin{table}[t]
  \centering
  \footnotesize
  \caption{\textbf{Llama-8B NVFP4, merged accuracy at matched parameters (scale r58 $\approx$ qlora
  r32), full-data protocol.} Against the merge-aware baseline the two lossless methods are
  \emph{tied} (mean $88.7$ vs $89.0$): \scaleqast{} reads marginally
  higher on the two coarse-grained classification tasks, \qat{} on Spider/SQL and CLINC150. Each gap
  sits inside its own cell's Wilson band, and a between-method difference needs the wider
  two-proportion band of Section~\ref{sec:setup}, so we read these as ties rather than orderings.
  \textbf{The Spider row in particular is not a method measurement}: seven runs per method of this
  configuration reverse its sign (Section~\ref{sec:results-matrix}). The
  two intent-classification rows are $3$-seed means and the other two are single-seed. We make no
  \method{} accuracy claim on Llama; both merge losslessly (merge loss in parentheses).}
  \label{tab:llamamatch}
  \setlength{\tabcolsep}{4pt}
  \begin{tabular}{@{}l ccc@{}}
    \toprule
    Dataset & \scaleqast{} & QLoRA & $\Delta$ \\
            & r58          & r32           & scale $-$ qlora \\
    \midrule
    banking77 & 93.72 (0.0)     & 92.99 ($-0.01$) & $+0.73$ \\
    agnews    & 94.26 (0.0)     & 94.08 (0.0)     & $+0.18$ \\
    clinc150  & 97.59 ($+0.01$) & 97.66 ($-0.01$) & $-0.07$ \\
    spider    & 69.15 (0.0)     & 71.28 ($+0.10$) & $-2.13$ \\
    \midrule
    \textbf{mean} & 88.68 & 89.00 & $-0.32$ \\
    \bottomrule
  \end{tabular}
\end{table}

\paragraph{The 100B+ MoEs on native E8M0.} Both gpt-oss-120B and DeepSeek-V4-Flash are native
\textbf{MXFP4/E8M0} MoEs (128 / 256 experts), trained via DeepSpeed expert parallelism and evaluated
through a single-process logprob bridge. On both, \scaleqast{}'s merge is \textbf{bit-exact} ($\max|\Delta W|=0$, codes
copied verbatim) while \qat{} is \textbf{accuracy-lossless} (merge loss $0.0$) but re-derives the code
plane, so the exact-merge result holds
beyond NVFP4-E4M3 on the MXFP4-E8M0 grid real serving stacks use, while the \ptq{} path is again
lossy ($-0.8$ to $-3.6\pp$; the E8M0-MoE base is strong, so the collapse is mild here).
\textbf{These are single-seed cells and we draw no ordering from any of them}: \scaleqast{} reads
nominally higher on the DeepSeek and gpt-oss classification cells, whose gaps sit inside their own
cells' Wilson bands, and \qat{} higher on both Spider cells (DeepSeek $76.0$
against \scaleqast{}'s $70.2$), \textbf{neither of which is a method measurement either}: gpt-oss's gap
exceeds that cell's entire headroom, and the
DeepSeek pair is addressed by Section~\ref{sec:limitations}, which states this paper's Spider
position in full. \scaleqast{} merges bit-exactly on DeepSeek Spider
($\max|\Delta W|=0$) and \qat{} is accuracy-lossless there, where the naive merge is also near-lossless ($+0.6\pp$) because the strong E8M0
base leaves little movement to lose. Earlier DeepSeek runs on other checkpoints gave different
absolute numbers, including a naive-merge \emph{sign flip} on Banking77; Table~\ref{tab:main}
reports the full-data MXFP4-E8M0 run as authoritative, and all runs agree qualitatively: naive-PTQ
merge is lossy and sign-unpredictable on DeepSeek, while both merge-aware methods stay at $\approx0$.

\paragraph{Isolating QAST.} On native NVFP4 Llama-8B Banking77 under the full-data protocol, the
QAST-vs-no-QAST-vs-weight-space triangle isolates each effect (Table~\ref{tab:money}): QAST ships
$+38.7\pp$ over merged naive QLoRA, while adapting the scales \emph{without} training on the
native grid and rounding at merge still loses $27.5\pp$. That second row is the operating point
grid-unaware continuous scale adaptation must pass through (e.g.\ LoRDS~\citep{tang2026lords}, Section~\ref{sec:related}),
and QAST removes its post-hoc rounding by construction.

\begin{table}[t]
  \centering
  \footnotesize
  \caption{\textbf{QAST isolation triangle (Llama-8B NVFP4, Banking77, native E4M3 merge; full-data
  protocol, $n{=}3080$).} QAST ships $+38.7\pp$ over merged naive QLoRA at the same pre-merge accuracy
  and the same single fine-tuning run; the no-QAST scale row (grid-unaware continuous scaling, the
  LoRDS operating point reduced to a native artifact) still loses $27.5\pp$ at merge in this run; a
  separately trained run of the same ablation, on this same protocol and harness, loses $18.2\pp$
  (Section~\ref{sec:related}), and the pilot-protocol QAST-off row of Table~\ref{tab:qastablation} is a
  third such run, so the penalty is run-dependent across $18$--$28\pp$ and QAST is essential. The merged column is the accuracy of the
  standalone native checkpoint, so it is also each row's deployable accuracy.}
  \label{tab:money}
  \setlength{\tabcolsep}{3pt}
  \begin{tabular}{@{}l ccc@{}}
    \toprule
    Method & Unmerged & Merged & Merge loss \\
    \midrule
    QLoRA, naive merge         & 93.6 & 54.5        & \bad{$-39.2$} \\
    \method{} without QAST     & 93.2 & 65.6        & \bad{$-27.5$} \\
    \textbf{\method{}}         & 93.1 & \best{93.1} & \best{0.0} \\
    \bottomrule
  \end{tabular}
\end{table}

\subsection{Training cost per step ($3.9\times$ dense, $2.2\times$ MoE)}
\label{sec:traincost}

QAST rounds only the block scales; \qat{}'s STE fake-quants the \emph{full weight} every forward (on
MoE, every active expert). Over 10 repetitions this is a $3.9\times$ (dense, $356.6$ vs $1400.8$\,ms) /
$2.2\times$ (MoE, $3958$ vs $8882$\,ms) per-step speed advantage at matched parameters on the
models measured (Table~\ref{tab:traincost}). QAST operates on
$\text{out}\times(\text{in}/\text{block})$ scale elements rather than $\text{out}\times\text{in}$
weight elements, a block-size reduction in the number of quantized values processed. This explains the
substantially lower STE overhead, although hardware efficiency and method-common computation prevent
the element-count ratio from translating directly into wall-clock speedup.

\begin{table*}[t]
  \centering
  \footnotesize
  \caption{\textbf{Training cost} (controlled micro-benchmark, Banking77, bs4/seq512, matched
  parameters, \textbf{median of 10 repetitions}, reported as the upper of the two central reps of a 100-step (dense) / 40-step (MoE) block; MoE uses
  gradient checkpointing, dense does not). \scaleqast{} trains $3.9\times$ faster than \qat{} on the
  dense model and $2.2\times$ faster on the MoE. The mechanism is the STE tax: the weight-space STE
  fake-quants the full weight every forward, QAST only the block scales, so it touches
  \emph{block-size} fewer elements ($16$ for NVFP4). The last column is the per-step cost over the
  no-STE \ptq{} baseline on the same model.}
  \label{tab:traincost}
  \begin{tabular}{@{}l cccc@{}}
    \toprule
    Method & ms/step & tok/s & Peak alloc GB & vs.\ no-STE \\
    \midrule
    \multicolumn{5}{@{}l}{\emph{Dense: Llama-8B NVFP4 (100 steps/rep, no grad-checkpointing)}} \\
    \midrule
    \textbf{scale r58 (QAST e4m3)}  & \best{356.6} & \best{5742} & \best{49.3} & $+49$ ms \\
    \qat{} r32 (merge-aware STE)    & 1400.8       & 1462        & 57.7        & \bad{$+1093$ ms} \\
    \ptq{} r32 (no STE)             & 308.0        & 6648        & 56.6        & --- \\
    \midrule
    \multicolumn{5}{@{}l}{\emph{MoE: Qwen3-30B-A3B NVFP4 (40 steps/rep, grad-checkpointing on)}} \\
    \midrule
    \textbf{scale r54 (QAST e4m3)}  & \best{3958}  & \best{517}  & 73.6        & \best{$-96$ ms} \\
    \qat{} r32 (merge-aware STE)    & 8882         & 231         & \best{66.4} & \bad{$+4828$ ms} \\
    \ptq{} r32 (no STE)             & 4054         & 505         & \best{66.4} & --- \\
    \bottomrule
  \end{tabular}
\end{table*}

At \textbf{$3.9\times$ dense} and \textbf{$2.2\times$ on MoE}, \scaleqast{} is also marginally the
fastest method overall there, $2.4\%$ ahead of the
no-STE \ptq{} baseline. The STE tax is the mechanism: relative to that no-STE baseline the weight-space
STE costs $+1093$ ms (dense) and $+4828$ ms (MoE) per step, while QAST costs $+49$ ms on dense and is
within noise of it on MoE ($-96$ ms), i.e.\ at least $20\times$ cheaper. The speedup ratio shrinks on MoE
not because the STE tax is smaller (it is $4.4\times$ \emph{larger} in absolute ms) but because the MoE
step carries more method-common fixed compute (expert GEMMs, routing, grad-checkpoint recompute) that
dilutes the STE share.

\paragraph{Memory is model-dependent, so the consistent win is speed.} On dense, scale is lighter
($49.3$ vs $57.7$ GB, because the weight-space path keeps a resident fp32 \texttt{baseW}); on MoE,
scale is $\sim11\%$ heavier ($73.6$ vs $66.4$ GB, because it retains bf16 codes and builds $W$).
Optimizer state is roughly equal at matched parameters.

\paragraph{Per step is not per run.} The benchmark holds the step count fixed, so it prices the
fake-quant and nothing else. Our protocol early-stops on validation (patience two), so the methods may
reach their accuracies in different numbers of steps; we did not log step-to-convergence on the matrix
cells, so no per-run cost claim follows from this table and we make none. A second cost sits outside the
timer: the scale path needs the ceiling clamp, rollback-on-NaN and LR-halving of
Section~\ref{sec:method} to train stably on MoE, which the weight-space baseline does not.

\subsection{Deployment options: zero overhead among the merge-based options measured}
\label{sec:deployopts}

\textbf{Among the four deployment options we measure on this stack, \scaleqast{} is the only one
simultaneously native-4-bit-compact ($2.7\times$ smaller than fp16), zero-per-forward-overhead (vs
unmerged's $+24\%$ decode-latency penalty), bit-exact to merge, and
merge-quantizer-independent within the target native format; each of the other three gives up at least
one of those axes, as that table's
caption itemizes.} \emph{Merge-quantizer-independent} means the merged artifact does not depend on which
quantizer a deploying tool would apply, because none runs at merge; it does not mean no quantization
occurs anywhere. QAST quantizes the scale onto the target grid in every forward pass during training
(Section~\ref{sec:method}), and the merge then writes that already-rounded byte. With both
merge-aware methods accuracy-lossless, merging into 4-bit must justify itself against keeping the
adapter unmerged and against dequantizing to merge at fp16. We measure all four options on the real
vLLM stack (Llama-8B NVFP4, Banking77; Table~\ref{tab:deployable}).

\begin{table*}[t]
  \centering
  \footnotesize
  \caption{\textbf{The four deployment options measured} (real vLLM stack, Llama-8B NVFP4,
  Banking77). Among these options \scaleqast{} is the only one simultaneously native-4-bit-compact,
  zero-per-forward-overhead, bit-exact to merge, and \emph{merge-quantizer-independent within the
  target native format}: it is the only row that
  passes all four axes, and each of the others gives up at least one. Option~1 pays a per-forward
  cost, option~2 gives up $4$-bit compactness and is merge-quantizer-independent only while it stays
  unquantized, and
  option~3 is conditional on its export convention being honoured at every code-touching event, for both
  accuracy-losslessness and merge-quantizer independence. ``cond.''$=$conditional;
  $^{\dagger}$on reproducing the training quantization rule.
  Section~\ref{sec:multiadapter} covers the unmeasured alternatives
  (pre-materialized checkpoints, runtime multi-adapter batching). Rows are different adapters and
  separate runs, so the accuracy column is not a controlled comparison
  (Section~\ref{sec:deployopts}). $^{*}$Option~1 is the only row measured as decode \emph{latency} rather than throughput: $144.9$ vs $116.8$\,ms/token, i.e.\ $+24\%$ latency ($-19.4\%$ throughput); the other three rows are vLLM decode tok/s.}
  \label{tab:deployable}
  \begin{tabular}{@{}l r l r r c l c@{}}
    \toprule
    Option & Store & Decode & Wt mem & Acc & Accuracy- & Bit-exact & Merge quant.\ \\
           & GiB   & tok/s$^{\!*}$ & GiB &     & lossless? & merge? & indep.? \\
    \midrule
    1. Unmerged (4-bit base $+$ fp16 LoRA)
      & 5.77 & $+24\%$ latency$^{\!*}$ & 5.81 & 92.4 & n/a & n/a & yes \\
    2. QLoRA dequant$\to$fp16 merge
      & 14.96 & 4057 & 14.99 & 95.0 & yes & yes, dequant.\ & cond. \\
    3. QLoRA re-quant 4-bit (\qat{})
      & \best{5.61} & 3769 & 5.65 & 93.6 & yes$^{\dagger}$ & no & cond. \\
    4. \textbf{\method{} merge (ours)}
      & \best{5.61} & \best{4081} & \best{5.65} & 94.8 & \best{yes} & \best{yes} & \best{yes} \\
    \bottomrule
  \end{tabular}
\end{table*}

The zero-overhead point is
exact: because the scale-QAST merge produces a \emph{byte-format-identical} native 4-bit checkpoint
(codes untouched, one E4M3 scale byte per block overwritten), serving is indistinguishable from the
base. Measured VRAM is bit-identical ($112.76$ GiB for both base and merged) and prefill/throughput
are within noise. Unmerged overhead is small at prefill and large batch ($+5.6\%$) and
only material at low-batch decode, so ``keep it unmerged'' is a legitimate choice; \scaleqast{}
removes even that residual tax at no storage, quality, or robustness cost. Two caveats on reading
the table. First, the rows are different adapters and separate runs, so the $92.4$-vs-$94.8$
accuracy difference between options 1 and 4 is run-to-run variation, not a merge gain: the
\scaleqast{} cell reads $94.8$ here (single run, inside its own $\pm2.6\pp$ Wilson floor) and $93.7$
in the full-test deploy sweep ($n{=}3080$, $\pm0.9\pp$ CI). Second, two rows are marked
\emph{conditional} in the merge-quantizer-independent column, for different reasons. The fp16 merge of
option 2 is safe only for as long as the checkpoint stays fp16: a serving stack that quantizes on load turns it
back into the \texttt{fp16\_lora+PTQ} workflow, which inherits the naive merge's loss by construction
(argument in Section~\ref{sec:results-matrix}, magnitude in Table~\ref{tab:ptqbaselines}). \qat{}'s
condition is that its export convention be honoured at \emph{every} code-touching event, not merely the
first: an in-family tool costing about a point and an out-of-family rule collapsing it, with format
conversion, code sharing, rollback, dedup and audit exposed either way
(Section~\ref{sec:deploysweep} scopes both halves, measured and inferred). Only the \method{} row
is unconditional. (Implementation note: the
lossless fp16 merge in option 2 must store the \emph{on-grid} $\operatorname{quantize}(W)$ in bf16,
not the raw continuous $\texttt{baseW} + B@A$, which is off-grid for a QAT-trained adapter and scores
$0\%$.)

\section{Capability Retention and Generative Behaviour}
\label{sec:forgetting}
\label{app:retention}

Beyond classification and SQL, we test (i) general-capability retention after task fine-tuning
$+$ lossless merge, and (ii) the generative, execution-scored MBPP task, our additional generative evaluation.

\paragraph{No consistent general-capability degradation observed.} On a broad lm-eval-harness suite
(MMLU/ARC-C/HellaSwag/TruthfulQA-mc2 0-shot, GSM8K 5-shot), neither merge-aware method shows a
consistent degradation relative to the base after task fine-tuning $+$ lossless merge: the worst single
delta from base across all merged states $\times$ benchmarks is \textbf{$-2.28\pp$}, so all negative
deltas are within about
$2.3\pp$ and benchmarks move in both directions with no consistent sign (Table~\ref{tab:general}); the
Qwen-30B MoE repeat is tighter still (worst delta $-0.59\pp$). \textbf{These are single runs with no
repeated seeds}, so this is an absence of consistent degradation, not a demonstration that no small
forgetting occurred.

\begin{table*}[t]
  \centering
  \footnotesize
  \caption{\textbf{General-capability retention (Llama-8B NVFP4, lm-eval-harness 0.4.12).} Each row is
  a merged, task-fine-tuned state (merged bit-exactly for \scaleqast{}, accuracy-losslessly for \qat{}) vs the un-adapted base; $\Delta$-from-base
  in parentheses. No consistent general-capability degradation is observed for \emph{either}
  method after task fine-tuning plus an accuracy-lossless merge: the worst single delta across all
  states $\times$ benchmarks is $-2.28\pp$, and retention
  deltas sit within $\sim2.3\pp$ of base for both methods with no consistent direction; single run per
  cell, so deltas of either sign are not separated from run-to-run variation.
  MMLU/ARC-C/HellaSwag/TruthfulQA-mc2 0-shot, GSM8K 5-shot.}
  \label{tab:general}
  \begin{tabular}{@{}l ccccc@{}}
    \toprule
    State & MMLU & ARC-C & HellaSwag & TruthfulQA-mc2 & GSM8K \\
    \midrule
    \textbf{base} & 65.91 & 53.07 & 78.79 & 52.29 & 73.39 \\
    \scaleqast{}~$\cdot$~banking77 & 64.87 ($-1.0$) & 54.35 ($+1.3$) & 78.14 ($-0.7$) & 51.98 ($-0.3$) & 71.72 ($-1.7$) \\
    \scaleqast{}~$\cdot$~clinc150  & 65.82 ($-0.1$) & 57.42 ($+4.4$) & 78.99 ($+0.2$) & 50.01 ($-2.3$) & 72.55 ($-0.8$) \\
    \qat{}~$\cdot$~banking77       & 65.40 ($-0.5$) & 53.84 ($+0.8$) & 78.38 ($-0.4$) & 52.67 ($+0.4$) & 72.10 ($-1.3$) \\
    \qat{}~$\cdot$~clinc150        & 65.13 ($-0.8$) & 54.95 ($+1.9$) & 78.83 ($0.0$)  & 52.83 ($+0.5$) & 71.87 ($-1.5$) \\
    \bottomrule
  \end{tabular}
\end{table*}

\textbf{The null is controlled, not merely flat.} A table of small deltas is ambiguous between ``no
forgetting'' and an instrument that would not have seen it, so we locate the positive control: on the
dense NVFP4 Llama model fine-tuned on the banking-intent cell under the pilot protocol, an
over-perturbed weight-space
adapter \emph{did} register degradation on this same suite: MMLU and the grade-school-math
benchmark both moved down, on the same harness that reads flat here. \textbf{That control shows the
harness can detect a large deterioration, not that small forgetting is absent}, so the flat deltas
above are a null at that instrument's resolution. We report it as a pilot-protocol observation rather than a table
precisely because that contrast does not reproduce on correctly-trained
adapters, so reduced forgetting is \emph{not} one of our differentiators (both methods preserve MMLU
near base, $-0.1$ to $-1.0$).

\paragraph{Lossless merge extends to generative code-gen.} On MBPP (pass@1 by executing the generated
function against its asserts), \scaleqast{} merges bit-exactly ($57.6/57.6$, $288/288$ exact).
Its $+1.4\pp$ over the bf16 base of $56.2$ is within the $n{=}500$ CI, though the validation trend
($56.7\to60.0$) supports genuine learning; \qat{} is $+0.2\pp$ (re-quant lossless), and every
naive-PTQ merge is negative ($-0.6$ to $-1.6\pp$, GPTQ worst). The MoE repeat (Qwen-30B) is also
exact: \scaleqast{} $73.2/73.2$ ($\max|\Delta W|{=}0$), \qat{} $73.6/73.6$. On Qwen MBPP, however, no
method learned above the base (the cell is base-saturated), so the exact-merge repeat there
demonstrates exactness only, not preserved learning. (On MBPP the naive collapse is
\emph{mild} because the strong base leaves little fine-tuning headroom for an 8B model; the dramatic collapse
contrast stays carried by the many-class classification tasks.)

\section{Why the Activation-Aware Quantizer Behaves Like the Naive One}
\label{sec:awqmech}
\label{app:awq}

\paragraph{AWQ's scaling axis \emph{is} the block axis.} AWQ's purpose is to protect salient channels,
so its buying nothing on E8M0 deserves a mechanism. On gpt-oss-120B it is
indistinguishable from naive RTN (Table~\ref{tab:ptqbaselines}: $-2.6$ vs $-2.6$ on AGNews, $+0.1$ vs
$-0.8$ on Banking77, $-2.8$ vs $-2.6$ on CLINC150); on the DeepSeek-V4 AGNews cell taken up below it is materially \emph{worse}
($-10.8$ vs RTN's $-3.6$, same $90.4$ unmerged reference and eval batch size). AWQ divides $W$ by a per-\emph{input-channel} scale
$s_j=\mathbb{E}[x_j^2]^{\alpha/2}$ and folds $1/s$ into the preceding op; on a microscaling format that
axis is not free, since MXFP4 shares one E8M0 exponent across exactly the $32$ contiguous input channels
$s$ varies along, so a within-block-varying $s$ only redistributes range inside a group that must round
to one shared power of two, pushing that exponent up and costing a mantissa bit for all $32$ weights at
once. AWQ's usual per-tensor or per-group deployment has no such collision.

Measured directly, $s$ has almost no room. On DeepSeek Banking77 the per-expert \texttt{down} scale is
\emph{byte-identically} $s\equiv1$ (exactly RTN) on $1374$ of $1376$ experts ($99.85\%$), while the
layer-shared \texttt{gate\_up} scale varies by $1.042\times$ within a $32$-channel block ($1.084\times$
across the tensor) and is $s\equiv1$ on $20$ of $43$ layers: a ${\sim}4\%$ perturbation of RTN on one
projection. The $\alpha$ search agrees: with $\alpha=0$ always in the candidate set, DeepSeek
\texttt{down} selects it in $1371/1376$ experts and gpt-oss's \texttt{gate\_up} in $22/36$ modules,
while DeepSeek's \texttt{gate\_up} splits evenly ($20$ layers at $0.0$, $23$ at $0.1$). This is a
near-flat landscape, not a decisive activation signal. That the penalty is a \emph{block} effect and not
activation-awareness in general is confirmed on $9$ real DeepSeek expert matrices: a block-constant $s$
of the same per-block geometric mean (equally activation-aware, unable to widen a $32$-channel group)
removes it entirely ($0.999\times$ RTN's Frobenius error, against $1.79\times$ for per-channel $s$ at
$\alpha{=}0.1$). \textbf{We quote that control for the sign and locus of the effect, not its size}: its
activation statistics are synthetic and the measured within-block spread ($1.042\times$) is flatter than
any regime we simulated. Either way an activation-aware objective has almost no room here, so it
inherits the naive quantizer's reconstruction quality, and its merge behaviour with it; on gpt-oss
AGNews AWQ's per-matrix error matches RTN's at the median ($5.11\mathrm{e}{-}2$ vs
$5.22\mathrm{e}{-}2$).

\textbf{One cell does not fit, and we mark it rather than absorb it.} Of the \textbf{eight} E8M0 AWQ
cells measured, \textbf{seven} place AWQ within noise of RTN: gpt-oss ($-2.6$ vs $-2.6$, $+0.1$ vs
$-0.8$, $-2.8$ vs $-2.6$, $+0.3$ vs $-1.5$) and DeepSeek Banking77 ($-2.2$ vs $-2.0$), CLINC150 ($0.0$ vs
$-0.8$) and Spider ($-1.2$ vs $+0.6$), every gap at or under $1.8\pp$. DeepSeek AGNews is the lone
exception at $-10.8$ against RTN's $-3.6$, and reconstruction quality does not explain it: in the same
configuration RTN is $3.9041\mathrm{e}{-}2$ against AWQ's $3.9340\mathrm{e}{-}2$, so AWQ reconstructs
$0.77\%$ \emph{worse} while losing $7.2\pp$ more accuracy; this is the opposite of what this mechanism
requires, since a better reconstructor should delete more. Because the deficit does not reproduce on
Banking77 at the same batch size, model and adapter, \textbf{we treat it as a single unexplained cell
rather than a property of AWQ on microscaled grids}. One structural candidate lies outside what a
weight-space metric sees: AWQ's merged model needs the reciprocal scale at runtime, so the deployed
forward computes $(x/s)Q^{\!\top}$ with the prescale on \textbf{bf16} activations while the
reconstruction metric evaluates the ideal $Q/s$ in fp32. That activation-side rounding step is
invisible to weight-space measurement and untested here (the test is to fold $1/s$ into fp32 weights
instead), so we label it a hypothesis, and a reminder that ``mergeable'' and ``reconstructs
well'' differ, an AWQ merge not being a stock checkpoint at all. Note too
that $\alpha=0$'s presence in the candidate set makes AWQ no worse than RTN on \emph{its own}
activation-weighted objective by construction, so the tail penalty above is about weight
error, precisely the objective AWQ is entitled to trade away.

Finally, an unanticipated measurement indicates \emph{why} $\alpha$ stays small: the selection is
largely a property of the \emph{model}, not the calibration data. The \texttt{down} projections carry
$s\equiv1$, i.e.\ byte-identically RTN, on $99.85\%$, $99.85\%$ and $99.91\%$ of experts on
Banking77, CLINC150 and Spider, and for \texttt{gate\_up} the Banking77 and CLINC150 statistics are
\emph{identical} to four decimals ($\alpha$ histogram $20$ layers at $0.0$ and $23$ at $0.1$ in both;
$s\equiv1$ fraction $0.4651$; within-block spread $1.0427$ vs
$1.0415$), which two tasks with different label sets and prompt lengths would not share if the
statistic tracked the token distribution.
\textbf{Spider is the partial exception and we record it rather than generalising past it}: there
$s\equiv1$ holds on only $20.9\%$ of \texttt{gate\_up} layers, so that projection's $\alpha$ is
\emph{mostly} structural with a genuine task-sensitive component. Either way the scale barely moves,
which further marks AGNews AWQ as anomalous rather than characteristic.

\section{Geometric Subspace Analysis: Why Adapting Only Scales Suffices}
\label{sec:expressivity}

The headline, \textbf{separate expressivity}, is two findings of different standing: the weight update's
task-useful direction does not live in the scale subspace (full-data, two cells), and an independently
trained scale correction is not distinguishable from weight-space LoRA at the protocols we ran (a null
inside a floor wider than the gap, plus a single-seed full-precision control read against an imported
floor). This section establishes both and places \method{} between PEQA and QLoRA.

\paragraph{Containment (analytical).} Order the families by the reachable weight update, codes
holding the frozen sign/relative pattern inside each block:
Set the low-rank parameterization aside for a moment and compare the \emph{unrestricted} update
families, which is where the containment is exact:
\begin{itemize}\itemsep1pt
  \item \textbf{PEQA family:} $\Delta W = \operatorname{diag}(\delta)\,W_q$, one multiplicative
        degree of freedom per output \emph{row}.
  \item \textbf{Per-block frozen-code family:} $\Delta W[o,i]=\code[o,i]\cdot\Delta S[o,g(i)]$ with
        $\Delta S$ unrestricted: one degree of freedom per (row, block), i.e.\ any per-block
        magnitude reweighting of the frozen code pattern, with no within-block sign or pattern flip.
  \item \textbf{Unrestricted weight-space:} $\Delta W$ arbitrary.
\end{itemize}
Per-row rescaling is contained in per-block frozen-code scaling, which is in turn contained in
unrestricted weight-space adaptation; the two scale families are strict subsets of the last because
they preserve the sign and relative code pattern inside every block. PEQA is recovered at granularity
$G{=}1$.

\method{} does not optimize the unrestricted per-block family directly. It parameterizes the correction
as $\Delta S=(BA)(\alpha/r)$ of rank at most $r$ (Section~\ref{sec:method}), reducing the trainable count
from $\text{out}\times G$ to about $r(\text{out}+G)$, exactly as ordinary LoRA parameterizes a
weight-space update with about $r(\text{out}+\text{in})$ parameters rather than $\text{out}\times
\text{in}$. \textbf{The low-rank methods are therefore regularized parameterizations of their respective
families, and we claim no strict set inclusion between fixed-rank \method{} and fixed-rank weight-space
LoRA}: a per-row rescaling can require a scale-delta matrix whose rank exceeds $r$, and a frozen-code
update generated by a low-rank $\Delta S$ need not be rank-$r$ in ordinary weight space once multiplied
by the block code pattern. Two knobs turn PEQA into \method{} (scale granularity $G$ and the low-rank
cap), both staying inside the \textbf{code-preserving} subspace, so every variant inherits PEQA's exact
merge and merge-quantizer independence.

The finding itself (separate expressivity, isolated by a full-precision control) is established in
Section~\ref{sec:bridge}; we first examine the one cell whose ordering appears to say otherwise, and a
harder generative target that tests the same worry.

\paragraph{On the Spider cell: the code-preserving family contains a \qat{}-quality solution when fitted,
and the shared-LR ordering does not survive a change of learning rate.} Spider is the one cell
where Table~\ref{tab:main} shows \qat{} ahead of \scaleqast{} ($71.3$ vs $69.2$ on Llama), so we
interrogated it with two experiments. \textbf{Both are single-run, and neither is this paper's Spider
position}, which is the null-plus-instability statement of Section~\ref{sec:limitations}; by the standard
that section applies (seven repetitions reversed the sign of a single-run Spider gap), neither
experiment here can attribute a cause, and what they add is only that the code-preserving family is not
obviously the binding constraint on this cell.
First, fitting rather than training: we take \qat{}'s trained Spider update and solve
per block for the least-squares best code-preserving scale delta (no training, no tuning). Kept in
continuous scale space that fitted correction reaches $69.73\%$, inside \qat{}'s $95\%$ interval
$[68.4, 74.0]$, so \textbf{the code-preserving family does contain a \qat{}-quality Spider solution}.
Rounded onto the native E4M3 grid the same correction collapses to $61.90\%$, below the $62.38\%$
zero-shot base, and the reason is quantitative: the fitted relative scale deltas average $0.0037$ while
E4M3's relative half-step is $0.031$, roughly $8\times$ larger, so the grid erases the correction. A
scale solution obtained \emph{post hoc} is therefore destroyed by the grid it was not trained on, which
is the sharpest statement of why QAST exists and the mechanism behind the no-QAST row of
Table~\ref{tab:money}.

Second, tuning the learning rate per method rather than sharing one. Our protocol fixes
LR $5\times10^{-5}$ for every method and cell, which is uniform but not neutral: sweeping
$\{5\times10^{-5}, 10^{-4}, 2\times10^{-4}\}$ on Llama Spider, both methods peak at $10^{-4}$ and the
ordering \textbf{reverses} (\scaleqast{} $71.86$ vs \qat{} $71.28$, merge loss still exactly $0.00$),
the shared LR having been suboptimal for both and costlier for \scaleqast{}. With this metric's
run-to-run spread ($1.84\pp$ between two runs of the same configuration and seed), the uniform-LR
Spider ordering reflects protocol and generative-eval noise rather than a capacity difference; the full
scope is stated in Section~\ref{sec:limitations}. We did not repeat the
sweep on the MoE Spider cells, so their orderings carry the same caveat and are untested.

\paragraph{A harder generative target, and on an MoE: Lean-4 proof fitting shows no capacity deficit
for the code-preserving family that this protocol can detect.} The concern behind the Spider cell is
that scale space might lack the capacity complex generation needs. Lean-4 theorem proving tests that
more severely than SQL ($4096$-token proof targets with plan-then-formal-proof structure), and it
answers the MoE gap the previous paragraph leaves open. We fine-tune on proof targets and score
cross-entropy over $350{,}765$ held-out assistant tokens
(Table~\ref{tab:lean}).

\begin{table}[t]
\centering\small
\caption{\textbf{Lean\,4 proof fitting}: cross-entropy on held-out assistant tokens (lower is better)
and the merge delta. \textbf{One run per method}, on a protocol lighter than the main matrix (one epoch,
truncated targets; scope in the text). No cross-run dispersion is available for this metric, so we
read the comparison as a null (no fitting-capacity deficit this protocol can detect), and claim no
margin here as an effect size, in either direction.}
\label{tab:lean}
\begin{tabular}{@{}llcc@{}}
\toprule
Model & Method & CE & merge $\Delta$ \\
\midrule
\multirow{3}{*}{Llama-8B} & \scaleqast{} r58 & \best{$0.4190$} & \best{$+0.0000$} \\
 & \qat{} r32 & $0.4336$ & \best{$+0.0000$} \\
 & \ptq{} r32 & $0.4204$ & \bad{$+0.1659$} \\
\midrule
\multirow{3}{*}{Qwen3-30B MoE} & \scaleqast{} r54 & \best{$0.4504$} & \best{$+0.0000$} \\
 & \qat{} r32 & $0.6210$ & \best{$+0.0000$} \\
 & \ptq{} r32 & $0.4415$ & \bad{$+0.1965$} \\
\bottomrule
\end{tabular}
\end{table}

\noindent \textbf{We claim no direction here, only a null.} \scaleqast{} posts a lower cross-entropy
than merge-aware \qat{} on both models, on the MoE by $0.17$ nats. That margin is the measurement, not
an effect size in either direction: one run per method, no cross-run dispersion for cross-entropy
anywhere in the paper, and the standard set by the seven-run Spider comparison, where repetition
reversed the sign of a single-run gap, applies to this comparison exactly as it does there. Two
\emph{hypotheses} for the outlying MoE baseline cell, neither tested here: \qat{}'s full-weight STE may
be the harder optimisation, and, since every measured baseline in this paper is our own instantiation
(Section~\ref{sec:limitations}), an unusually poor cell in that arm is as consistent with a
configuration effect as with a harder optimisation. The CE column is the unmerged model; its merge is exact in loss as well as in weights ($0.4190 \to 0.4190$ on Llama, $0.4504 \to 0.4504$ on Qwen), while the naive
path degrades by $+0.17$ to $+0.20$ nats at merge: the same collapse the accuracy tables show, visible in
fitting terms. \textbf{The conclusion is a null, symmetrically applied}: against
\emph{merge-aware} weight-space training (the comparison the merge guarantee is about), this
protocol detects no fitting-capacity deficit for the code-preserving family. We mark the third row
rather than let the column be read against that sentence: it is
the unconstrained no-STE weight-space run, present as the naive-merge reference, and on the MoE it fits
marginally better than \scaleqast{} by a margin an order of magnitude smaller again, while being
unable to merge losslessly at all, so it buys its fit with the property this paper is about.
\textbf{We also name the control this null lacks}, since elsewhere we insist on one
(Appendix~\ref{app:retention}): the naive row's merge degradation shows the metric responds to a weight
change, not that it would respond to a \emph{capacity} restriction, which is what ``this protocol can
detect'' claims. A configuration restricted enough to register a fitting deficit here (a rank or
granularity floor low enough to bind) is the positive control we did not run.

We are explicit about the scope. It is a \emph{fitting-capacity} result, not a task-success one: Lean is scored by loss here because proof success requires a Lean\,4 $+$ Mathlib
toolchain, and in a $50$-theorem spot check pass@1 is ${\approx}0$ for \emph{every} configuration
including the un-adapted base, so no method-level success claim is available. The runs are also a lighter
protocol than the main matrix: $1$ epoch, $3000$ training records on Llama and $400$ on Qwen, and $73\%$ of
records truncated at $\text{max\_len}=4096$, identically for all methods.

\subsection{How much capacity the task actually asks for}
\label{sec:rankneed}

The projection statistics below say what the scale family does \emph{not} do: it does not capture a
rescaling component of $\Delta W$, and its overlap with a freely trained update sits at the chance value
for a subspace of its dimension. That leaves an obvious question, which is answerable directly rather
than by asking whether the scale subspace is somehow privileged. \textbf{How many directions does the
task require?} Truncate a trained weight-space update to rank $k$ per layer by SVD, install it, and
measure the fraction of the fine-tuning gain that survives,
$\text{need}(k) = (\text{acc}_k - \text{acc}_{\text{base}})/(\text{acc}_{\text{full}} -
\text{acc}_{\text{base}})$.

\begin{center}\footnotesize
\setlength{\tabcolsep}{5pt}
\begin{tabular}{@{}lrrrr@{}}
\toprule
cell & headroom & need(1) & need(2) & $\sigma_1$ energy \\
\midrule
Banking77 & $37.0\pp$ & $0.966$ & $\mathbf{1.000}$ & $0.827$ \\
CLINC150  & $26.0\pp$ & $\mathbf{1.000}$ & $0.990$ & $0.858$ \\
AGNews    & $17.8\pp$ & $0.972$ & $\mathbf{1.000}$ & $0.790$ \\
\bottomrule
\end{tabular}\\[2pt]
{\footnotesize Llama-8B, \qat{} r32; need(4) equals need(2) on all three cells.}
\end{center}

\noindent \textbf{On these cells the task asks for one or two directions per layer out of the thirty-two
trained}, and the trained adapter has already concentrated $79$--$86\%$ of its spectral energy into a
single direction without being asked to. A completion-masked cross-entropy read-out agrees on Banking77
($\text{need}$ rising monotonically $0.983 \to 0.992 \to 0.997 \to 1.000$ over ranks $1,2,4,32$), which is
the control that licenses using that higher-resolution metric where accuracy is under-powered.

This reframes the expressivity question and explains the results below rather than competing with them.
\textbf{A family at $1/16$ of full dimension is not a binding restriction on a task that uses two
directions}, so weight-space and frozen-code scale adaptation succeed for the same reason and neither
needs to be privileged. It also accounts for three findings that otherwise look unrelated: the
chance-level $6.25\%$ overlap predicts nothing because most of $\Delta W$ is surplus to the task; a
rank- and norm-matched \emph{random} update is reproduced by the scale family almost as well as the
trained one, because reproducing any low-rank target is easy while only two of its directions matter; and
the granularity ladder is flat because every rung already affords two directions.

\emph{Scope, and what the generative cell does and does not add.} These are classification cells on one
dense model at a single seed, with $\text{need}(1)$ carrying roughly $\pm0.08$ at $n_{\text{eval}}{=}400$;
the resolution supports ``rank one nearly suffices, rank two suffices'' and nothing finer. On Llama
Spider the accuracy measurement is under-powered by its own arithmetic ($5.0\pp$ of headroom against a
$\pm6.9\pp$ half-width), and the cross-entropy read-out on the full $1034$-example test set shows
$\text{need}$ \emph{above} one at every truncation and falling monotonically toward it
($1.047, 1.020, 1.011, 1.003$ for ranks $1,2,4,8$), i.e.\ rank-$1$ truncation attains a \emph{better}
completion likelihood than the full adapter. We read that as the likelihood-relevant content of the
update being essentially rank one there, with the additional directions mildly harmful, which is
independent support for the over-perturbation behaviour reported in Section~\ref{sec:limitations}.
\textbf{We do not read it as a statement about generation quality}: cross-entropy and execution match are
demonstrably decoupled on this cell, where a run reaching a lower validation loss produced worse SQL, so
whether rank-$2$ truncation preserves \emph{exec-match} on a generative task remains open.

\subsection{The PEQA$\leftrightarrow$QLoRA bridge: separate expressivity, not a projection}
\label{sec:bridge}

\begin{figure*}[t]
  \centering
  \begin{minipage}{0.32\textwidth}
    \centering
    \includegraphics[width=\textwidth]{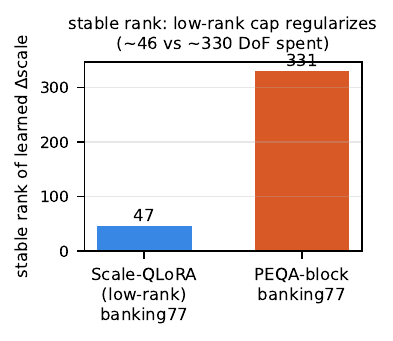}
    \caption{\textbf{Stable rank of the learned $\Delta$-scale.} Direct PEQA-block spends nearly all
    DoF ($\approx330$); low-rank scale-QAST concentrates ($\approx46$): the low-rank cap regularizes.}
    \label{fig:peqasvd}
  \end{minipage}\hfill
  \begin{minipage}{0.32\textwidth}
    \centering
    \includegraphics[width=\textwidth]{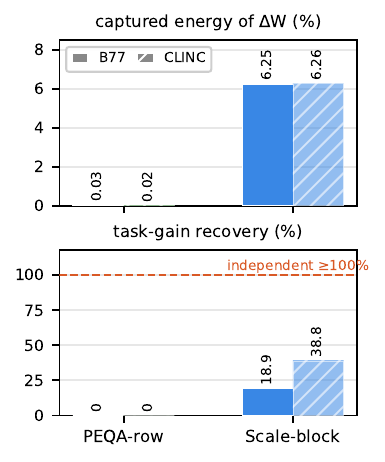}
    \caption{\textbf{Projection recovery.} Captured energy / task-gain of $\Delta W$
    projected onto each scale subspace (full-data protocol) vs.\ the $\ge100\%$ from independent
    training, a point estimate under the \emph{pilot bridge} protocol, i.e.\ a null on the gap within
    that protocol's own noise floor. The two are not commensurable measurements: separate
    expressivity, not a $\Delta W$ projection.}
    \label{fig:peqaproj}
  \end{minipage}\hfill
  \begin{minipage}{0.32\textwidth}
    \centering
    \includegraphics[width=\textwidth]{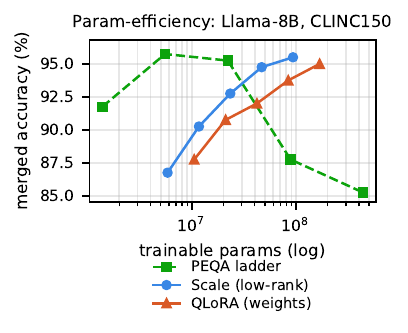}
    \caption{\textbf{Param-efficiency Pareto (CLINC150).} The exact-merge family (PEQA ladder,
    low-rank Scale) reaches the accuracy band at lower parameter cost; on this task, under the pilot
    bridge protocol we observe no QLoRA point on the frontier; at $n_\text{eval}{=}400$ the accuracy
    differences among frontier points are within noise, so this concerns parameter cost, not accuracy.
    Because the scale-QAST merge produces a byte-format-identical native 4-bit checkpoint, serving is
    indistinguishable from the base: on a real vLLM NVFP4 stack, base and merged-\scaleqast{} share the
    same $112.76$ GiB VRAM, with prefill and decode throughput within noise (zero inference overhead).}
    \label{fig:peqapareto}
  \end{minipage}
\end{figure*}

\paragraph{Separate expressivity, not a projection of $\Delta W$.} It is tempting to claim
scale-space ``captures the rescaling component'' of the weight update. Two full-data Llama-8B probes
are inconsistent with that claim, and the statistic carrying the argument is task-gain recovery rather
than captured energy. \textbf{We expect a matched-norm random direction to recover no task gain, but
that is the control we did not run}: we ran precisely that control for the geometric statistics below,
and used it to set them aside, so we mark this as an untested expectation and rest the argument on the
independent-training result, which does not depend on it. Projecting a
freely-trained weight-space $\Delta W$ (Llama Banking77, 224 linears) onto each family's rescaling
subspace (Table~\ref{tab:peqaproj}, Figure~\ref{fig:peqaproj}) shows QLoRA's beneficial update lives in
\emph{neither}. The captured-energy column, by contrast, is the chance row and we read it as one: the
per-block subspace captures exactly the random-chance overlap of a 16-element
block with one fixed direction,
\[
\frac{\lVert\mathcal{P}_{\text{scale}}(\Delta W)\rVert_F^2}{\lVert\Delta W\rVert_F^2}
  \;=\; \frac{1}{\text{block}} \;=\; 6.25\%,
\]
so the trained $\Delta W$ has \emph{no special alignment} with the scale subspace: it sits at the value
any direction would give at this dimension ratio, a sanity check rather than a measurement. What the
update does say is in the recovery numbers. Under the full-data protocol the block projection recovers
only $18.9\%$ of the task gain
(Banking77) and the PEQA per-row projection $\approx0\%$ (that per-row subspace captures $0.03\%$ of
the energy); a
second cell (Llama CLINC150) reproduces the energy fraction to within $0.01\pp$ of $1/16$ with block
recovery $38.8\%$ and per-row $\approx0\%$. So on both cells the projection of the weight update
recovers only a minority of the gain, yet an
\emph{independently}-trained \method{} closes the QLoRA$-$PEQA gap to within the pilot bridge
protocol's own noise floor, with $\ge100\%$ of QLoRA's gain as the point estimate (gap decomposition
below). \textbf{Those two figures come from different protocols and are not directly commensurable}:
the projection recoveries are full-data measurements, whereas the independent-training figure is a
point estimate from the lighter pilot bridge protocol, inside a floor wider than the gap it closes.
\textbf{A different trained update reads the opposite way, and we say why it is the wrong probe rather
than merely preferring ours.} A weight update trained \emph{on} the native grid, unlike this grid-unaware
$\Delta W$, projects almost perfectly onto the block-scale subspace. But such an update was constrained
toward the representable set by its own training, so projecting it measures that constraint; it cannot
tell us whether the task-useful direction is aligned with scale space, which is the question. The
grid-unaware update is the one that answers it, and it is the one the overlap statistic is computed on.
Scale-space finds a different, task-equivalent solution inside
the code-preserving subspace, which has $\text{out}\times G \approx 10^6$ DoF per layer, far more
than a rank-32 LoRA. The PEQA-row subspace captures $\sim\!200$--$300\times$ less of $\Delta W$'s energy
than the block subspace ($0.02$--$0.03\%$ vs $6.25\%$), but that ratio is a count of dimensions
(both projections sit at their own chance values), which is why it fails to predict the accuracy gap
the full-precision control below measures.

\begin{table}[t]
  \centering
  \footnotesize
  \caption{\textbf{Mechanism: scale-space is separate expressivity, not a $\Delta W$ projection}
  (Llama-8B, 224 linears, Banking77 / CLINC150, full-data protocol). Energy and task-gain of a
  freely-trained weight-space $\Delta W$ captured when projected onto each family's rescaling subspace.
  The captured energy is $1/16$, the random-chance overlap, on both cells. The last row is a pilot
  observation, not a matched comparison: independently-trained low-rank \method{} closes the whole
  QLoRA$-$PEQA gap on the cells where a gap exists (CLINC150, Spider), but it comes from the pilot
  bridge protocol and different cells than the projection rows above. It is a point estimate inside that
  protocol's own noise floor, which is wider than the gap it closes, and therefore not commensurable
  with the full-data projection rows.}
  \label{tab:peqaproj}
  \begin{tabular}{@{}l cc@{}}
    \toprule
    $\Delta W$ projected onto & Energy \% & Gain rec.\ \% \\
    \midrule
    PEQA (per-row)        & 0.03/0.02 & ${\approx}0$/${\approx}0$ \\
    \method{} (per-block) & 6.25/6.26 & 18.9/38.8 \\
    \midrule
    \emph{independent training} (pilot) & --- & \best{$\ge100$} \\
    \bottomrule
  \end{tabular}
\end{table}

\paragraph{The capacity is not a quantization artifact \emph{on the cells controlled}: no
full-precision deficit detected on two controlled cells.} If the code-preserving subspace only works because $4$-bit
coarsening has already discarded
fine-grained directions, the effect should vanish without quantization. It does not: on one dense
model and two intent-classification tasks, single-seed, which is the scope of the claim. On
\emph{unquantized}
bf16 Llama-8B we freeze the within-block pattern of the original weights and learn one low-rank
multiplicative coefficient per (row, block-of-16), $W' = W\,(1+\Delta S[o,g(i)])$, with no codes, no grid
and no STE anywhere, at effective parameters matched to LoRA ($1.005\times$). It recovers $99.3\%$ and
$100.7\%$ of weight-space LoRA's task gain: Banking77 $93.99$ vs $94.25$ ($-0.26\pp$, from a $57.50$ base)
and CLINC150 $97.91$ vs $97.69$ ($+0.22\pp$, base $67.98$), both within the $\le0.7\pp$ cross-seed spread
we measure on the corresponding quantized cells. \textbf{These runs are single-seed on one dense model
and two tasks, and that floor is imported}: no cross-seed dispersion was measured for these
full-precision arms, so this is \textbf{no detectable deficit rather than evidence of equivalence}. Confining the representable update to the
per-block multiplicative family, whose \emph{unrestricted} dimension is $1/16$ that of full weight space
for block-$16$ NVFP4, therefore costs nothing measurable in these two full-precision experiments,
which also reconciles the projection result above: the scale-only solution is a different solution of
equal quality, not an approximation of the weight-space one. The $4$-bit grid is what makes this useful
(it is what the merge writes), not what makes it work.

Two controls sharpen the reading. Raising the learning rate to $2\times10^{-4}$ makes the same arm
\emph{worse} ($90.68$/$95.47$), so the primary arm is not under-trained and the match is not an artifact of
stopping early. And the granularity knob degrades gently: at $G{=}1$, the
full-precision PEQA analogue, one coefficient per row with $61\times$ fewer parameters still reaches
$92.24$/$96.56$, i.e.\ $2.0$/$1.1\pp$ below LoRA, which is milder than the projection geometry suggests, so
we read the ladder as a smooth capacity dial rather than a threshold and do not claim that
PEQA's subspace is impoverished \emph{in proportion} to its captured energy. These controls are
single-seed on one model and two intent-classification tasks.

\paragraph{Two structural limits of the code-preserving family.} The family has two
structural limits: it cannot change the sign/relative pattern \emph{inside} a block (that needs code
reassignment), and it cannot move an exactly-zero weight, since $0\cdot s=0$ for any scale. The second
is the harder ceiling: $7.1\%$ of NVFP4 weights carry code $0$, so that fraction of any target update is
unreachable outright, permanently, for every frozen-code family (PEQA, \method{}, and continuous-scale
variants alike). It is a property of the format's code alphabet, not of the adapter. Neither
prohibition, however, is newly triggered by fine-tuning. The geometric statistics one would reach
for first to quantify how binding they are turn out to be degenerate (they sit at their
random-direction values), so we report them only to set them aside,
in Section~\ref{sec:ablations}, and we do not use them to explain where scale-space trails.

\paragraph{Granularity alone over-perturbs.} We sweep a direct (non-low-rank) additive scale
delta at granularity $gd$ ($gd{=}1$ is PEQA, full per-block is ``PEQA-block'') with codes frozen and QAST
rounding, so \textbf{every point merges exactly} ($\max|\Delta|=0$). Table~\ref{tab:peqaladder}
(pilot bridge protocol) shows PEQA is a strong floor, within the pilot noise floor ($\sim\pm4\pp$ at $n_\text{eval}{=}400$) of the best ladder point on every cell, i.e.\ not distinguishable from it, and that extra raw granularity
helps briefly then \emph{over-perturbs} on competent bases (Llama CLINC150
$95.75@gd4\to87.75@gd64\to85.25$ block). The family must regularize the extra DoF, not spend them raw;
the pattern holds on the Qwen-30B MoE, where all 20 points
fold bit-exactly over attention plus all 18{,}432 expert linears.

\begin{table*}[t]
  \centering
  \footnotesize
  \caption{\textbf{Granularity ladder, merged accuracy.} Direct (non-low-rank) scale-FT at
  granularity $gd$: $gd{=}1$ is PEQA, PEQA-block is full per-block. \textbf{Every Qwen point merges
  exactly in weight space} ($\max|\Delta W|=0$ recorded on all $20$ points, covering all $18{,}432$
  expert linears). We record accuracy-level merge loss only for Llama, where $18$ of $20$ points are
  exactly $0.0$; the two exceptions are Spider $gd{=}16$ ($-0.25$) and PEQA-block ($-0.5$), i.e.\ one
  and two samples at $n{=}400$. Qwen's merged-vs-unmerged deltas span $[-0.75,+0.25]$, likewise
  $\le3$ samples. All of these are eval noise, not merge error, but the weight-space identity is
  verified on Qwen only, so we do not assert it for Llama. PEQA is a strong
  floor; raw granularity over-perturbs competent bases ($\downarrow$). Pilot bridge protocol
  ($\max_\text{train}{=}2000$, 2 epochs, $n_\text{eval}{=}400$).}
  \label{tab:peqaladder}
  \begin{tabular}{@{}ll ccccc@{}}
    \toprule
    Model & Dataset & PEQA ($gd1$) & $gd4$ & $gd16$ & $gd64$ & PEQA-block \\
    \midrule
    \multirow{4}{*}{Llama-8B}
      & Banking77 & \best{87.25} & 78.25 & 81.5 & 84.5 & 69.75 \\
      & CLINC150  & 91.75 & \best{95.75} & 95.25 & 87.75$\downarrow$ & 85.25$\downarrow$ \\
      & AGNews    & 89.75 & 89.5 & \best{91.25} & 89.75 & 88.5 \\
      & Spider    & 64.0 & \best{66.0} & 64.5 & 65.0 & 64.5 \\
    \midrule
    \multirow{4}{*}{Qwen3-30B}
      & Banking77 & \best{89.75} & 86.75 & 89.75 & 81.5 & 78.0$\downarrow$ \\
      & CLINC150  & \best{92.75} & 90.75 & 89.5 & 89.5 & 89.5 \\
      & AGNews    & \best{91.0} & 90.0 & 89.25 & 87.75 & 90.25 \\
      & Spider    & \best{73.0} & 71.25 & 71.75 & 71.25 & 71.25 \\
    \bottomrule
  \end{tabular}
\end{table*}

\paragraph{The low-rank cap is the regularizer.} Low-rank scale adaptation is the clean alternative to
raw granularity: on CLINC150 \scaleqast{} climbs smoothly $r4{=}86.8\to r64{=}95.5$, reaching the best
direct-ladder point without collapse. An SVD of the stacked learned $\Delta$-scale matrix
$\Delta S\in\mathbb{R}^{\text{out}\times G}$ (Llama, 224 linears; Figure~\ref{fig:peqasvd}) explains why:
direct PEQA-block spends nearly every DoF (\textbf{stable rank $\approx330$}, over-perturbing), whereas
low-rank scale-QAST concentrates (\textbf{stable rank $\approx46$}) yet stays task-competitive.
Decomposing the gap $\text{acc}(\text{QLoRA}_\text{best})-\text{acc}(\text{PEQA})$ into a scale-space
term and a weight-space residual (Table~\ref{tab:peqagap}; pilot bridge protocol), low-rank
\method{} closes $\ge100\%$ of any real gap with a $\sim0$/negative residual, on both dense Llama
and the Qwen-30B MoE. \textbf{Only the null reading of that ratio is claimed, because its magnitude is
uninterpretable}: the denominator is the QLoRA$-$PEQA gap, which sits inside the pilot bridge
protocol's own noise floor and on some cells is at or below zero, so the statistic divides two
sub-noise quantities; this is the same reason Table~\ref{tab:retention} declines to quote retention where its
denominator is not large against that metric's floor. The
``PEQA-block\,$-$\,PEQA'' column is $\le0$ on all cells except Spider ($+0.5$, within noise): low-rank
scale-space closes the gap, not raw
granularity. We read this as evidence the exact-merge family reaches the same accuracy band, not
a general scale-space accuracy win, since on some cells PEQA already meets QLoRA and the remaining gap is
negative.

\begin{table*}[t]
  \centering
  \footnotesize
  \caption{\textbf{Gap decomposition (Llama-8B NVFP4; pilot bridge protocol).} $\text{Gap}=\text{acc}(\text{QLoRA}_\text{best})
  -\text{acc}(\text{PEQA})$, split into a scale-space gain ($\text{\method{}}_\text{best}-\text{PEQA}$) and a
  weight-space residual. Low-rank \method{} closes $\ge100\%$ of any real gap; the PEQA-block\,$-$\,PEQA
  column is $\le0$ on all cells except Spider ($+0.5$, within noise), so low-rank scale-space closes the
  gap, not raw granularity.}
  \label{tab:peqagap}
  \begin{tabular}{@{}l ccc c cc c@{}}
    \toprule
    Dataset & PEQA & Scale-best & QLoRA-best & Gap & Scale gain & Frac of gap & PEQA-block $-$ PEQA \\
    \midrule
    CLINC150  & 91.75 & \best{95.5}  & 95.0 & $+3.25$ & \best{$+3.75$} & \best{1.15} & $-6.5$ \\
    Spider    & 64.0  & \best{73.75} & 72.0 & $+8.0$  & \best{$+9.75$} & \best{1.22} & $+0.5$ \\
    Banking77 & 87.25 & 87.0 & 84.0 & $-3.25$ & $-0.25$ & PEQA $\ge$ QLoRA & $-17.5$ \\
    AGNews    & 89.75 & 88.75 & 89.75 & $\sim0$ & $-1.0$ & saturated & $-1.25$ \\
    \bottomrule
  \end{tabular}
\end{table*}

\paragraph{Param-efficiency Pareto.} On Llama-8B NVFP4, \method{} costs
$1.454$M params/rank vs QLoRA's $2.621$M ($\sim1.8\times$ cheaper). Plotting trainable params vs merged
accuracy on CLINC150 (Figure~\ref{fig:peqapareto}), the exact-merge family reaches the same accuracy
band at lower parameter cost: PEQA $gd4$
reaches $95.75$ at $5.5$M params, \textbf{matching QLoRA's best accuracy ($95.0$ at $168$M,
$\sim30\times$ more params) within noise}, and Scale-$r64$ matches QLoRA at $\sim1.8\times$ fewer
params. At $n_\text{eval}{=}400$ the accuracy differences among frontier points are within noise, so
this is a statement about parameter cost, not a dominance ordering on accuracy.

\subsection{Capacity, rank curves, and ablations}
\label{sec:ablations}

\paragraph{Constrained capacity acts as a regularizer.} Scale-space corrects one scalar per block, a
subset of what weight-space expresses at the same rank, and at matched parameters the two lossless
methods tie on deployable accuracy (Table~\ref{tab:llamamatch}). A pilot-protocol sub-study suggests the
constrained capacity buys generalization: on the competent base CLINC150, QLoRA reaches $+18.4\pp$ higher val but
only $+9.2\pp$ higher test, so its val$\to$test gap ($24.0$) is $\sim1.6\times$ scale-space's ($14.8$);
on the weak base Banking77 both fit easily and the gaps match ($6.1$ vs $6.5$). This is not
unconditional: a single scale error is amplified block-fold, so within-round val can crater (CLINC150
scale val hit $12.5\%$ before early-stop restored $75.8$). The defensible claim is narrow: at matched
parameters, constrained capacity yields a tighter generalization gap when trained stably.

\begin{figure*}[t]
  \centering
  \includegraphics[width=0.19\textwidth]{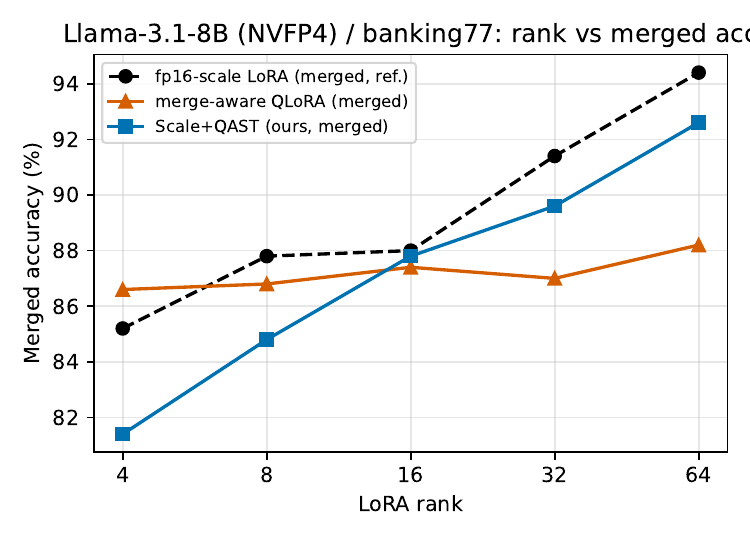}\hfill
  \includegraphics[width=0.19\textwidth]{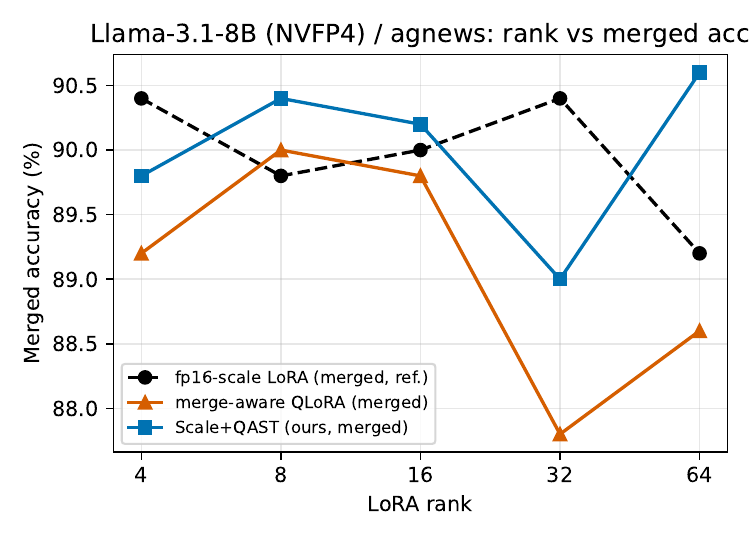}\hfill
  \includegraphics[width=0.19\textwidth]{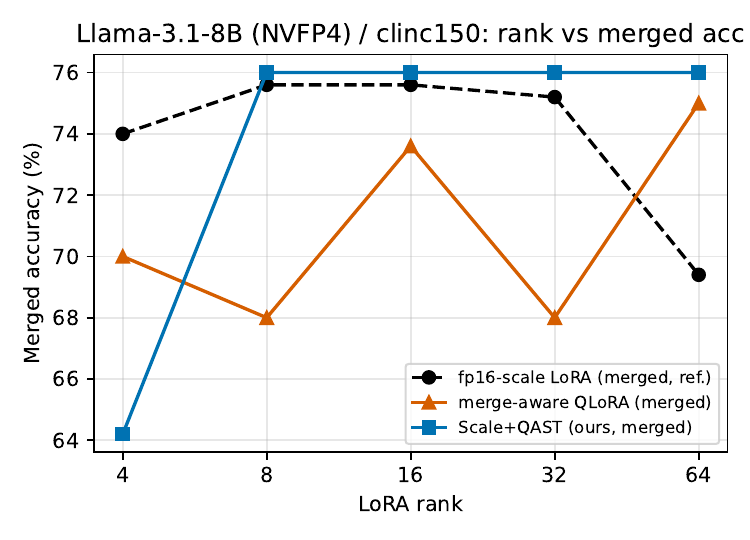}\hfill
  \includegraphics[width=0.19\textwidth]{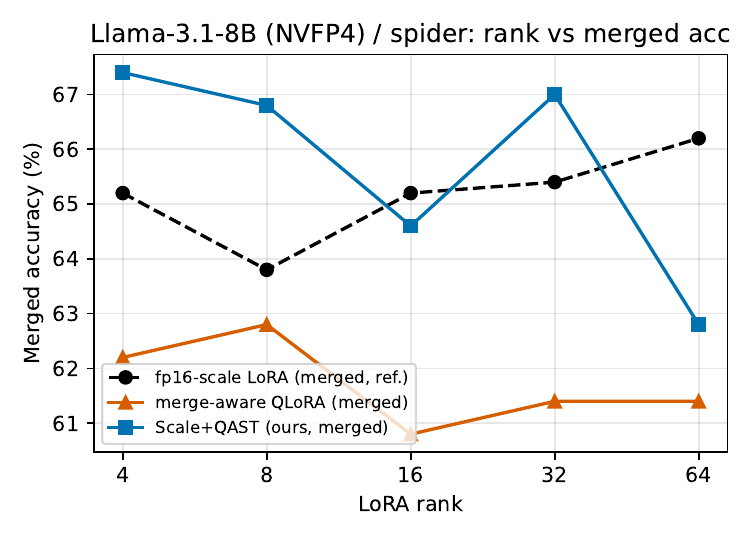}\\[4pt]
  \includegraphics[width=0.19\textwidth]{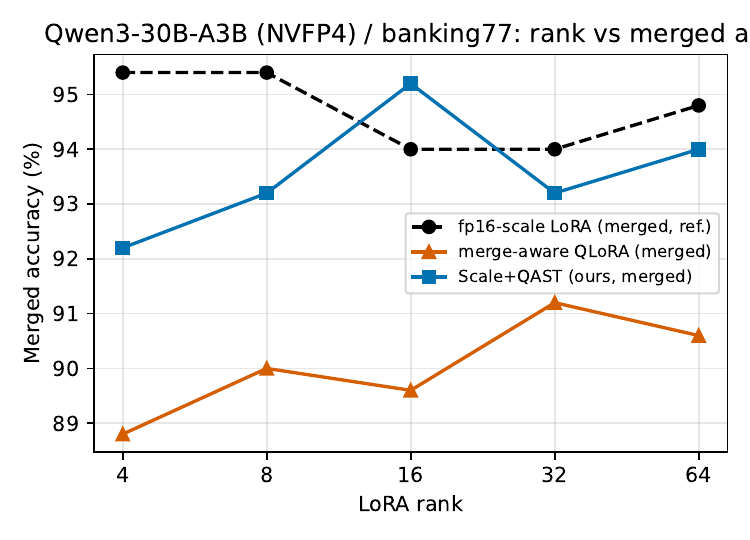}\hfill
  \includegraphics[width=0.19\textwidth]{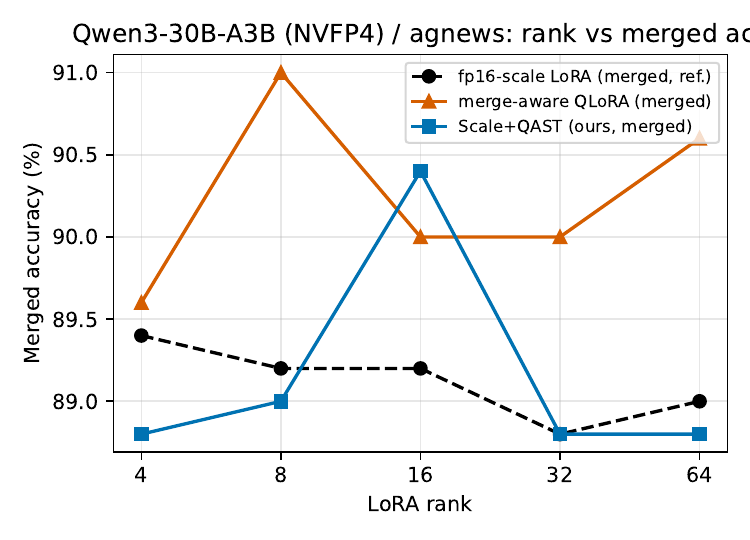}\hfill
  \includegraphics[width=0.19\textwidth]{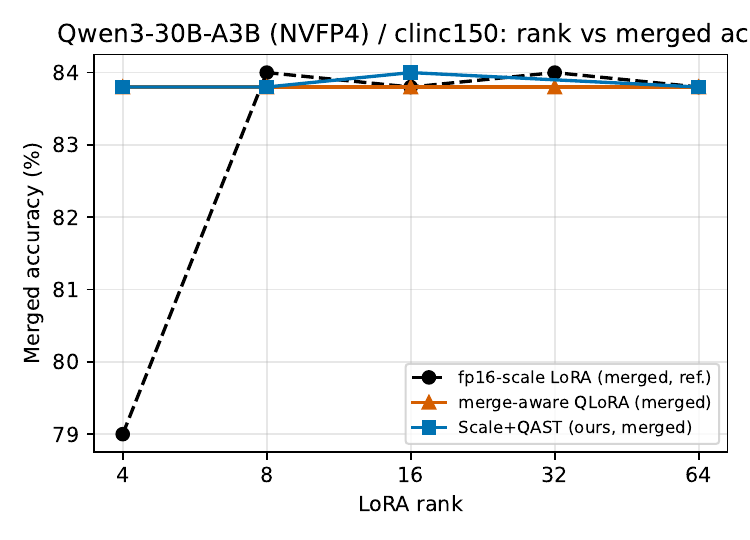}\hfill
  \includegraphics[width=0.19\textwidth]{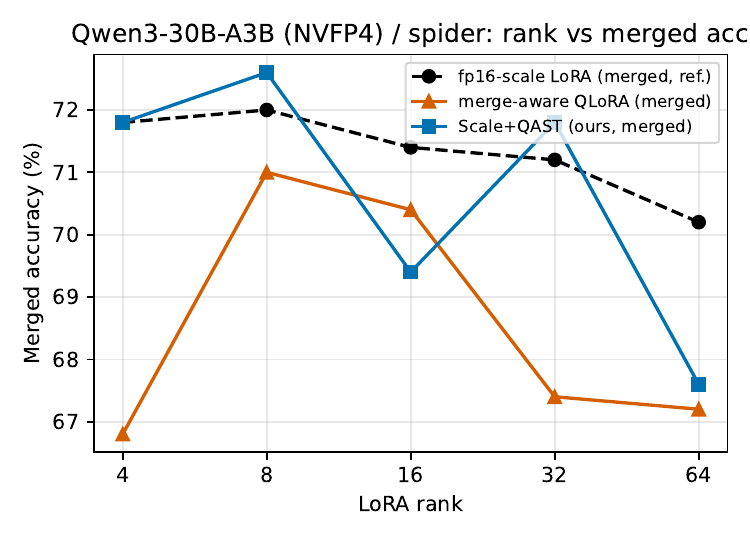}
  \caption{\textbf{Accuracy vs.\ LoRA rank}, all three methods, Llama-8B (top) and Qwen-30B (bottom),
  across Banking77 / AGNews / CLINC150 / Spider (pilot protocol). Merge is lossless for all three
  lines on every cell.}
  \label{fig:rankcurve}
\end{figure*}

\paragraph{Rank curves: per-rank gains are base-dependent, merge is always lossless.}
Figure~\ref{fig:rankcurve} plots all three methods (pilot protocol); merge is lossless for all
three lines on every cell. Whether scale-space uses added rank more effectively is base-dependent: on
Banking77 it climbs cleanly (scale r4$=81\to$r64$=92.6$, overtaking flat \qat{} by $\sim$r16); on Spider
it wins the level by $\sim4$--$5\pp$ at every rank but stops climbing; AGNews is a saturated tie; on
competent-base CLINC150 it under-fits at r4 then early-stops at the base ceiling.

\paragraph{QAST grid ablation.} On Llama-8B Banking77 (Table~\ref{tab:qastablation}) all four
predicted behaviors hold: QAST-off is lossy at merge ($-18.8\pp$ in this run, a different no-QAST run than Table~\ref{tab:money}'s $-27.5\pp$; both are lossy and the magnitude is run-dependent); QAST-on is exactly $0.0$ on both the
E4M3 and E8M0 grids it trains for; and the cross-grid cell (a QAST-e4m3 adapter merged to E8M0) is
\emph{not} lossless ($-13.0\pp$; $-8.2\pp$ AGNews), so ``exactly lossless'' carries the ``for the grid
QAST targets'' clause. E8M0-QAST's unmerged accuracy is lower ($86.4$ vs $92.8$), since the coarser
power-of-two grid caps attainable accuracy, but merge is still exactly lossless on that grid.

\begin{table*}[t]
  \centering
  \footnotesize
  \caption{\textbf{QAST-grid ablation (merge loss, pp), Banking77, Llama-8B (seed 0, scale r58,
  early-stop).} QAST-off is lossy at merge; QAST-on is exactly $0.0$ on the grid it trained for; the
  cross-grid cell (QAST-e4m3 merged to E8M0) is \emph{not} lossless, so ``exactly lossless'' carries the
  ``for the format QAST targets'' clause.}
  \label{tab:qastablation}
  \begin{tabular}{@{}ll ccc l@{}}
    \toprule
    Training grid & Merge grid & Unmerged & Merged & Merge loss & Verdict \\
    \midrule
    no-QAST (continuous) & fp16 (ref)      & 93.4 & 93.2 & $-0.2$          & $\sim$lossless ref \\
    no-QAST (continuous) & e4m3            & 93.4 & 74.6 & \bad{$-18.8$}   & lossy (PTQ round) \\
    QAST-e4m3            & e4m3            & 92.8 & 92.8 & \best{0.0}      & lossless (matched) \\
    QAST-e8m0            & e8m0            & 86.4 & 86.4 & \best{0.0}      & lossless (matched) \\
    QAST-e4m3            & e8m0 (mismatch) & 92.2 & 79.2 & \bad{$-13.0$}   & NOT lossless (cross-grid) \\
    \bottomrule
  \end{tabular}
\end{table*}

\paragraph{Stability and param-matching.} The E4M3-ceiling clamp is essential for numerical
stability on MoE (Qwen-30B: clamp
off $\to$ first non-finite at step 18 $\to$ merged $0.0$; clamp on $\to62.6$, loss $0.0$); it does not
trigger on dense Llama. No fixed LR wins both regimes and validation early stopping provides the robust
single-setting strategy in the regimes we test
(Table~\ref{tab:lrknob}). In an equal-rank ablation (pilot protocol), scale r32 $\ge$ \qat{} r32 on every Llama
cell (several gaps within the $\pm2.6\pp$ floor), suggesting the equal-rank parameter deficit does not
disadvantage scale-space; the full-data matched-param comparison is a tie
(Table~\ref{tab:llamamatch}). Scaling r32$\to$r58 helps where there is head-room
(Banking77 $89.6\to92.2$), does nothing on saturated CLINC150, and \emph{hurts} Spider ($67\to63$,
over-perturbation). Across every knob \scaleqast{} merges losslessly: the knobs move unmerged and merged
accuracy together, never the merge gap.

\begin{table}[t]
  \centering
  \footnotesize
  \caption{\textbf{Which LR setting is decisive (merged accuracy, Llama-8B, \scaleqast{} r32).} No
  single fixed LR wins both regimes; validation early-stopping at $2\times10^{-4}$ is near-best on both
  with no collapse.}
  \label{tab:lrknob}
  \setlength{\tabcolsep}{4pt}
  \begin{tabular}{@{}l cc@{}}
    \toprule
    Tuning & banking77 & clinc150 \\
           & (weak base) & (base 76) \\
    \midrule
    fixed LR $5\times10^{-5}$    & 82.0        & 70.6 \\
    fixed LR $2\times10^{-4}$    & \best{92.4} & \bad{18.6 (collapse)} \\
    \textbf{early-stop $+$ val}  & 89.6        & \best{76.0} \\
    tanh-clamp $\gamma{=}0.1$    & 52.8        & 72.2 \\
    \bottomrule
  \end{tabular}
\end{table}

\begin{table*}[t]
  \centering
  \footnotesize
  \caption{\textbf{bf16 LoRA ceiling vs quantized Scale+QAST (deployable), Llama-8B, $n{=}500$.}
  Quantized \scaleqast{} is indistinguishable from the full-precision LoRA ceiling at $n{=}500$
  ($\pm2.6\pp$) on Banking77/AGNews/CLINC150, and trails it by up to $\sim4\pp$ on Spider. On CLINC150
  early stopping kept the base weights, because fine-tuning only hurt this already-competent base, so
  the bf16 ceiling \emph{is} the $76.0$ zero-shot. \textbf{The last column is not one protocol}, and we
  label each entry rather than presenting the ranges as seed spread: Banking77/AGNews are 3-seed means
  \emph{within this ceiling sub-study}, a separate protocol from the full-test main matrix, whose own
  seed inventory is stated in Section~\ref{sec:setup};
  CLINC150 spans the learning-rate recipes of Table~\ref{tab:lrknob} ($70.6$ at a fixed
  $5\times10^{-5}$ up to $76.0$ with validation early-stop); Spider spans the two rank settings of the
  param-match ablation ($63.0$ at r58, $67.0$ at r32). The CLINC150 and Spider entries are therefore
  \emph{sensitivity} ranges over a hyper-parameter, not confidence intervals.}
  \label{tab:bf16}
  \begin{tabular}{@{}l ccl@{}}
    \toprule
    Task & bf16 zero-shot & bf16 LoRA FT (upper bound) & Quantized Scale+QAST (deployable) \\
    \midrule
    banking77   & 60.0 & \best{90.6}          & 91.3 \hfill{\footnotesize 3-seed mean} \\
    agnews      & 79.0 & \best{90.8}          & 90.4 \hfill{\footnotesize 3-seed mean} \\
    clinc150    & 76.0 & \best{76.0}          & 70.6--76.0 \hfill{\footnotesize over LR recipes} \\
    spider (SQL)& 66.0 & \best{67.2}          & 63.0--67.0 \hfill{\footnotesize over rank r58/r32} \\
    \bottomrule
  \end{tabular}
\end{table*}

\paragraph{Iterated merging is inconclusive.} Sequentially merging $N$ adapters did not compound
\qat{}'s code-reassignment error over 5 rounds, and \scaleqast{} collapsed once on CLINC150 under a
noisy $n{=}120$ early-stop selector, so we draw no multi-adapter claim from this probe.

\paragraph{Set aside: the geometric statistics one would reach for first.} These bear on how binding the
two structural limits of Section~\ref{sec:bridge} are; all three sit at their random-direction values.
We measured how often reaching a freely-trained weight-space optimum $W+\Delta W$ would additionally
require a within-block reorientation, over all $224$ linears ($7.0$B weights) on four Llama tasks.
Measured against the full target, neither prohibition is newly invoked: sign flips are needed on
$10^{-7}$--$10^{-6}$ of nonzero weights and $10^{-6}$--$10^{-5}$ of blocks need reorientation beyond a
positive rescale (mean within-block cosine $>0.9998$; the positivity constraint $s>0$ binds on no block
we found). \textbf{These statistics are provably uninformative about reachability, and we report them only to
set them aside.} A matched-norm control makes the point empirically: a \emph{random} $\Delta W$ with the same
per-layer Frobenius norm agrees with the trained one to six digits on mean within-block cosine
($0.999963$ both) and to four significant figures on the target-normalized residual. Two of the
quantities are degenerate by construction: the projection residual measured against $\Delta W$ is exactly
scale-invariant and equals the random-direction value $1-1/16$ for the trained update, the random
control, and every rescaling of it alike, so the trained $\Delta W$ sits \emph{at} the
null; and the target-normalized residual is a deterministic function of the update norm, its closed form
$(1-1/16)\,(\lVert\Delta W\rVert/\lVert W\rVert)^2$ predicting $5.4483\times10^{-5}$ against a measured
$5.4475\times10^{-5}$, so it reports $\lVert\Delta W\rVert/\lVert W\rVert$ rather than anything about the
cone. Scaling the trained update locates where the prohibitions do bind: $5\times$ its norm barely moves
the distribution, $20\times$ ($\lVert\Delta W\rVert/\lVert W\rVert = 0.15$) puts $48\%$ of blocks below
cosine $0.99$ and $100\times$ reaches $18\%$ sign flips, while real fine-tuning sits at $0.008$, roughly
$20\times$ below that regime.

We therefore do not use these statistics to explain where scale-space trails. For completeness,
reorientation need does not predict the accuracy gap and if anything runs backwards (AGNews needs the
most and \scaleqast{} wins there by $0.18\pp$; Spider needs nearly the least and \scaleqast{} loses by
$2.03\pp$), but with a
degenerate predictor at $n{=}4$ tasks (Spearman $-0.6$, exact permutation
$p=0.42$) this is uninformative in either direction.

\paragraph{Full-precision upper bound.}
\label{sec:upperbound}
Fine-tuning the un-quantized bf16 base with ordinary LoRA r32 (same LR/data/epochs, $n{=}500$;
Table~\ref{tab:bf16}), quantized \scaleqast{} is \textbf{indistinguishable from the full-precision LoRA ceiling} at $n{=}500$ on
Banking77 ($91.3$ vs $90.6$), AGNews ($90.4$ vs $90.8$), and CLINC150 ($\le76$ both), and approaches
it on Spider ($63$--$67$ vs $67.2$). On the classification tasks NVFP4 plus scale-space adaptation
is within the $\pm2.6\pp$ noise floor of a full-precision LoRA; on Spider the gap is up to $\sim4\pp$, against a bf16
LoRA that itself reaches only $67.2$.

\section{Extended related work}
\label{app:related}

This appendix states in full the prior families summarised in Section~\ref{sec:related}, and what each
one pays at merge time.

\paragraph{PEQA: scale-only, a strong floor whose deficit is a direction, not a range.}
PEQA~\citep{kim2023peqa} fine-tunes one quantization scale
per output channel with codes frozen; AlphaTuning~\citep{kwon2022alphatuning} is the earliest
frozen-code precedent, fine-tuning only the scaling factors of a post-training-quantized LM with
its binary codes frozen. Merge is free and exact because only the scale moves. Independently trained,
PEQA is a strong floor and even matches weight-space QLoRA on some cells; where a gap does exist it
trails higher-capacity families by $3$--$8\pp$ (Appendix~\ref{sec:bridge}) on the two tasks with
headroom, both from the lighter pilot bridge protocol, whose between-method difference floor separates
them: the Spider end of that range clears the floor, the classification end sits inside it, and Spider is
our noisiest metric. \textbf{The deficit is therefore a direction supported on one cell, not a measured
range.} We do \emph{not} explain it by projection geometry either: PEQA's per-row subspace
recovers $\approx0\%$ of a trained weight update under projection, but that is the chance value for a
subspace of its dimension, as \method{}'s own overlap is for its dimension, so the projection predicts
nothing about the accuracy gap. The better-controlled statement of the same gap is the full-precision
granularity ladder of Appendix~\ref{sec:bridge}, which compares that granularity against LoRA on an
unquantized model at matched seeds.

\paragraph{QLoRA: weight-space, off-grid, serving-fragile.} QLoRA~\citep{dettmers2023qlora} trains a
low-rank \emph{weight} delta at high capacity, but the delta is off the native grid, so the
re-quantized merge is lossy (the collapse above) and, when made lossless by retraining
(next), stays coupled to the serving stack's rounding rule. GPTQ~\citep{frantar2023gptq} and
AWQ~\citep{lin2024awq} are the error-compensating and activation-aware re-quantizers we benchmark as
the calibrated-PTQ baseline; LoftQ~\citep{li2024loftq} co-initializes base and adapter but still merges
in weight space. The 2024 quantized-PEFT wave, IR-QLoRA~\citep{qin2024irqlora},
LQ-LoRA~\citep{guo2024lqlora}, ApiQ~\citep{liao2024apiq}, and EfficientQAT~\citep{chen2024efficientqat},
improves the quantized base or its initialization, and all of them still merge through weight-space
re-quantization.

\paragraph{Merge-aware weight-space quantization-aware training (QAT): accuracy-lossless but quantizer-dependent and costly.} Ordinary weight-space LoRA trained with a native-grid straight-through estimator (STE),
$W \leftarrow W + (\operatorname{quantize}(W) - W)\texttt{.detach()}$, trains on-grid so its
re-quantized merge reproduces training. We call this baseline \qat{}; L4Q~\citep{jeon2024l4q} is
the published QAT$+$LoRA prior in this family, and our \qat{} baseline is a faithful instantiation of
that published merge-aware family on the native microscaling grid. It removes the
collapse and is accuracy-lossless when the training quantization rule is reproduced. Its guarantee is
conditional: it re-derives
every E2M1 code at merge, so it holds only when whatever tool writes those codes uses the exact
scale-selection and rounding rule of the training STE. A tool that merely differs still rounds to
nearest per block and costs about a point; a rule from outside that family (round-toward-zero,
stochastic rounding, whole-tensor scaling) collapses
the model silently (Appendix~\ref{sec:deploysweep}), while re-quantizing every weight each forward step
makes it costly to train (Appendix~\ref{sec:traincost}).
LoTA-QAF~\citep{chen2025lotaqaf} also merges losslessly, via grid-aligned \emph{ternary} adjustments
to the quantized weights themselves. It sharpens a distinction this paper relies on: lossless merge
and code-invariance are different properties. LoTA-QAF's merge rewrites the codes, so it forfeits
the shared code plane, byte-exact rollback, and audit payoffs of
Section~\ref{sec:codeinv-short}.
LR-QAT~\citep{bondarenko2024lrqat} is the closest weight-space relative of our \qat{} baseline: it trains
low-rank auxiliary matrices under quantization and absorbs them into the quantized tensor at the end, with
no inference overhead. The distinction is the same one we draw throughout: absorbing an update
\emph{into} the quantized weights re-derives them, so the artifact is again a function of the quantizer,
whereas we write only scale bytes and leave the code plane identical. Its INT/fixed-point setting also
differs from the microscaling grids studied here, so we do not run it as a baseline and the comparison
stays analytical.

\paragraph{Format-aware PTQ for FP4.} A recent line of work quantizes \emph{specifically} for the
microscaling grids rather than treating them as generic 4-bit. MR-GPTQ~\citep{mrgptq2026} adapts GPTQ to
FP4 with block-wise Hadamard transforms and format-specific handling, and ARCQuant~\citep{arcquant2026}
augments the activation matrix with quantized residual channels to keep a strictly unified NVFP4 format.
Two implications for this paper. First, they are stronger post-hoc baselines than the four quantizers we
sweep, so our naive-merge collapse should be read as a statement about what a \emph{merge} does to an
adapter, not as a claim that FP4 PTQ is weak in general. Second, MR-GPTQ independently reports that
MXFP4's power-of-two scale quantization is the harder of the two grids, reached from the PTQ side rather
than the adaptation side. Our own same-model grid ablation (Section~\ref{sec:results-matrix}) finds the
coarser grid costs little \emph{attainable} accuracy once the base damage it causes is controlled for,
so the two results bound the effect from different directions. Neither method targets merge-time code preservation, so neither is a
substitute for the property we study; a direct empirical comparison on an FP4-native stack is future work.

\paragraph{Adjacent scale/STE techniques.} QA-LoRA~\citep{xu2023qalora} is the nearest INT4
neighbor: it absorbs the LoRA delta into INT4 \emph{zero-points} for exact merge. Ours is the zero-point-free
microscaling analogue, where the absorbing parameter is the per-block scale and the coarse E8M0/E4M3
grid \emph{forces} the quantization-aware scale-training (QAST) STE of Section~\ref{sec:method},
which QA-LoRA's asymmetric INT4 setting does not need.
LSQ/LSQ+~\citep{esser2020lsq,bhalgat2020lsqplus}, PACT~\citep{choi2018pact}, and
PoT/APoT~\citep{li2020apot} are the learned-step-size and power-of-two STE lineage QAST is adjacent to
(E8M0 is power-of-two). Recent FP4 pre-training and fine-tuning recipes train on the native grid
from the start~\citep{wang2025fp4training}; we address the orthogonal adapter-merge artifact
problem on already-quantized checkpoints. The STE technique is not new; the application to lossless
adapter merge on microscaling hardware is.

\paragraph{LoRDS: low-rank scale adaptation without a native artifact.} LoRDS~\citep{tang2026lords}
is the nearest algorithmic neighbor: it also learns a low-rank correction in scale space,
parameterizing the quantization scaling manifold as continuous low-rank matrices $S=BA$ and
unifying PTQ initialization, QAT, and multiplicative PEFT over NormalFloat-style grids. Three
differences separate the two. First, LoRDS's scaling is continuous, element-wise-capable, and
served through custom Triton kernels, so it produces no standard-format checkpoint; \method{}
adapts the \emph{existing} per-block scale bytes of a native checkpoint and QAST-quantizes the effective
scale onto the hardware grid during training, so the merged artifact is byte-compatible with stock
serving. Second, materializing a continuous-scale solution as native bytes requires exactly the post-hoc
scale rounding our no-QAST configuration performs, so that configuration instantiates the operating point
LoRDS's continuous scaling must pass through to become a native artifact. \textbf{What we ran is our
own grid-STE-off ablation, not LoRDS's implementation}: LoRDS is a full method with its own PTQ
initialization, parameterization, training recipe and kernels, and the equivalence we argue for is
about that single component (a direct comparison against the released implementation is future work).
That ablation loses $18.2\pp$ when rounded to the E4M3 grid at export and a separately trained run of it
loses more, so the penalty is large and run-dependent ($18$--$28\pp$), set by how far each run's
continuous scales drift off-grid; QAST removes it entirely ($0.0$) by training on the grid in the first
place. Third, the lifecycle and
code-invariance payoffs of Section~\ref{sec:codeinv-short} require a native artifact and are
ours alone. LoRDS therefore sharpens rather than pre-empts the contribution the introduction
states: grid-aware adaptation of native microscale bytes, plus the artifact-level payoffs.

\section{Extended scope and discussion}
\label{app:scope}

This appendix states in full the scope limits summarised in Section~\ref{sec:limitations}.

\paragraph{Scope of the contribution, and target-grid specificity.}
The apparent accuracy orderings between the two merge-aware methods do not survive repetition where
repetition exists (Qwen classification, Llama Spider) and are unsupported where it does not (the
single-seed DeepSeek and gpt-oss-120B cells), so we assert no direction
(Section~\ref{sec:results-matrix}). The contribution is code-invariance and its payoffs
(Section~\ref{sec:codeinv-short}, Appendix~\ref{app:codeinv}) plus the expressivity and bridge
characterization (Appendices~\ref{sec:expressivity},~\ref{sec:bridge}); the naive-PTQ collapse is
motivation. Among the payoffs, the deploy sweep, multi-adapter serving, rollback and patch bandwidth are
directly measured (Llama-8B, plus the Qwen-30B MoE for the first two, with the sweep's Qwen arm on our own
re-quantization harness, not on vLLM; Appendix~\ref{app:sweepscope}), as is the cross-tool
implementation-divergence audit (six NVFP4 quantization outputs from five codebases, but three of the six
are our own harnesses, so only three of the enumerated pairs are tool-against-tool between
implementations neither of which is ours; Appendix~\ref{app:baselines}). QAST also targets a
specific hardware scale grid, so train-time grid selection must match the deployment target: a
QAST-e4m3 adapter merged to E8M0 loses $13.0/8.2\pp$ (Appendix~\ref{sec:ablations}), format conversion
re-quantizes \emph{both} methods' artifacts, and the precise guarantee is: within a fixed native format,
scale grid, block layout, and code plane, no quantizer runs at merge and the merge is bit-exact. This is a cross-grid
caveat, not a failure of the method: on the grid QAST trains for, merge is exact on both E4M3 and E8M0.

\paragraph{Evaluation scope.} The study spans a dense 8B model and three MoEs (30B, 120B, and the
158B DeepSeek-V4-Flash) across four main-matrix tasks (intent and topic classification and
text-to-SQL), with MBPP as an additional generative, execution-scored evaluation and a
general-capability retention study (Appendix~\ref{app:retention}). The main matrix
uses the full test set with $\pm0.5$--$1\pp$ Wilson CIs; three seeds were run on the
Llama Banking77/CLINC150 and Qwen-30B classification cells, and every other row is single-seed, while
the deploy sweep and EP-bridge sub-studies use $n{\approx}500$--$1000$ with a $\pm2.6\pp$ floor
(Section~\ref{sec:setup}). Scale-adaptation is adjacent to QA-LoRA (zero-points), PEQA and
AlphaTuning (direct scale fine-tuning), LoRDS (continuous low-rank scaling), and LSQ/PoT (STE); our
novelty is empirical/systems, not a new quantization primitive (Section~\ref{sec:related}).
The bridge study (Appendix~\ref{sec:bridge}) evaluates a PEQA operating point ported to
microscaling$+$QAST rather than the original INT-scale codebase, on the lighter pilot bridge protocol,
so its absolute numbers show the shape and the merge control, not a SOTA comparison. DeepSeek-V4 Spider
is evaluated through the same device-map generation bridge as gpt-oss, since in-expert-parallel
generation is confounded by capacity token-dropping; \method{} merges bit-exactly there
($\max|\Delta W|=0$) and \qat{} is accuracy-lossless there, and we read the in-EP DeepSeek headroom
figures as an upper bound rather than as the cell's true headroom. The iterated-merge probe is inconclusive and we draw no claim from it
(Appendix~\ref{sec:ablations}).

\paragraph{Single-backend protocol and sweep confounds.} A cell's merged and unmerged numbers always
come from the same backend, and \method{}'s exactness is established in weight space
($\max|\Delta W|$), so the observed zeros are a mathematical identity rather than a coincidence of two
noisy accuracies. The sweep is narrower than it looks, and \textbf{three of its limits are
load-bearing}, all stated in full in Appendix~\ref{app:sweepscope}. \emph{It is confounded in one column}: the whole-tensor rule changes
granularity rather than rounding and collapses even the un-adapted base, so RTZ and
stochastic are the clean isolators, and the whole-tensor column changes granularity \emph{and} damages
the base rather than isolating a rounding rule. \emph{It is asymmetric by design}: the weight-space
condition is five re-quantizations against one \scaleqast{} evaluation read across five columns,
licensed by the power-of-two-grid pair, where the rules \emph{are} run over a scale-merged artifact and
return an identity, so on the finer grid the scale condition has no on-grid positive control.
\emph{And it prices no lifecycle event}: those ${\sim}0\%$ rows
come from an out-of-family rule applied to a configuration we construct rather than one we observed
shipping, only round-to-nearest of the five being substantiated as shipping behaviour. This is a property
contrast on the rule axis, not a lifecycle price, whose two tiers Appendix~\ref{sec:deploysweep} states
in full. \qat{} can be highly sensitive when its codes are re-derived under a mismatched quantization
rule; how often that occurs we do not measure. \textbf{Two parts of that argument are inferences, not
measured rows}: what
a load-time rule costs a \emph{correctly} pre-materialized \qat{} artifact (a control we did not run) and
what it costs a scale-merged NVFP4 artifact (estimated by the zero-delta base control, whose non-zero
reading on that grid we record as open).

\paragraph{Baselines are our own instantiations.} Every measured baseline here is implemented by us:
the merge-aware weight-space STE (\qat{}), four post-hoc quantizers, a PEQA operating point ported to
microscaling$+$QAST, and LoRDS reduced to a native artifact. We do not run the original
implementations of PEQA, QA-LoRA, L4Q, LoTA-QAF, or LoRA-Inlaid, so comparisons to those works are
analytical. LoTA-QAF is the sharpest untested one: it also merges losslessly, and because its ternary
adjustment rewrites the quantized codes it should forfeit the code-plane payoffs we measure, but testing
it faithfully needs its own trainer. Appendix~\ref{app:baselines} states this in full, together with three
further limits: the single-checkpoint divergence census, the four-task reorientation null, and the
pilot protocol behind the bridge sub-studies.

\paragraph{We do not interpret any observed Spider ordering as a method-level accuracy difference.} Spider is our noisiest metric ($1.84\pp$ run-to-run
at fixed seed, from execution-match scoring combined with validation-based checkpoint selection), and it is
the one place where the two lossless methods appear to separate. We therefore state explicitly that none of
those separations is a method result. On Llama we have seven runs per method and the single-run sign
\emph{reverses} (Section~\ref{sec:results-matrix}). On gpt-oss the gap exceeds the cell's entire
fine-tuning headroom. On DeepSeek-V4, the only Spider cell with both ample headroom ($31.2\pp$) and a
gap above the noise floor ($-5.8\pp$), we repeated both arms, and the repetition settles the question in
a way a single pair could not. \textbf{\qat{} reproduces and \scaleqast{} does not.} Re-evaluating the
\qat{} adapter returns $76.4$ against its published $76.0$; retraining \scaleqast{} at the identical
configuration returns $51.8$ against its published $70.2$, an $18.4\pp$ swing between two clean runs,
ten times this metric's $1.84\pp$ floor. The second run was not a failed one: it completed all $2538$ steps
with zero non-finite events and reached a \emph{lower} validation loss ($0.0676$ vs $0.0772$), so the run
that fit the objective better generated worse SQL. Per-example scoring locates the difference precisely:
the two methods are indistinguishable on semantically wrong-but-executable output ($18.0\%$ of items for
\scaleqast{} against $17.2\%$ for \qat{}), and differ almost entirely on output that fails to execute at
all ($30.2\%$ against $6.2\%$). \textbf{The observed performance gap on this run is concentrated in
non-executable outputs rather than executable-but-semantically-wrong outputs.} This points to
decoding/syntactic instability rather than clear evidence of a representational capacity ceiling. We therefore claim no ordering here (an arm with an $18\pp$ run-to-run spread cannot be
ordered against anything), and we flag the instability itself, not a capacity gap, as the open problem.
The earlier evidence of hyperparameter fragility on this cell points the same way (\qat{}'s reference is a
retrain after a non-finite-value bug; \scaleqast{}'s published run is the unclamped variant, where a
$\gamma{=}0.1$ delta clamp instead scores $41.0$, a $29\pp$ swing from one hyperparameter). Our second
generative task does not fill the gap either, and we say so rather than quoting it: MBPP has only $374$
training examples, so its fine-tuning headroom on Llama is $1.4\pp$ ($57.6$ against a $56.2$ base) against a
$\pm4.3\pp$ Wilson half-width at $n{=}500$; \scaleqast{} is nominally ahead there ($57.6$ vs $55.8$) and
level on Qwen ($73.2$ vs $73.6$), but a cell with less headroom than its own confidence interval cannot
support an ordering in either direction. \textbf{The generative evidence in this paper is therefore one
adequately powered task (Spider on DeepSeek), and strengthening it needs a generative benchmark with both a
large test set and enough training data to open real headroom}, not more cells like MBPP.

\paragraph{Calibrated PTQ on a sparse MoE is Hessian-starved.} GPTQ's error feedback needs a per-expert
Hessian, but an expert only sees the tokens routed to it, so on a sparse MoE under a calibration-token
budget a third to a half of the \texttt{gate\_up} Hessians are rank-deficient and some experts receive no
tokens at all, leaving the inverse carried by damping rather than by data. This is a property of
calibrating a sparse MoE, not of our implementation, and we report the per-cell token and
rank-deficiency statistics with every GPTQ row so it is auditable rather than implicit. Against our own
interest in a clean caveat, the conditioning does \emph{not} visibly degrade any GPTQ cell in this
matrix, and the worst-conditioned cell is also the one with real headroom and no accuracy ceiling
(Appendix~\ref{app:gptq} gives the statistics). We therefore state the caveat as a limit on the
\emph{guarantees} of calibrated PTQ on a sparse MoE (a Hessian below full rank means the
error-feedback objective is not the one the method assumes), and not as a doubt about these numbers.

\section{Baselines, and Three Further Limits of the Evidence}
\label{app:baselines}

\paragraph{Baselines are our own instantiations.} Every measured baseline here is implemented by us: the
merge-aware weight-space STE (\qat{}), four post-hoc quantizers, a PEQA operating point ported to
microscaling$+$QAST, and LoRDS reduced to a native artifact (its continuous low-rank scale correction with
grid rounding deferred to export; Section~\ref{sec:related}). We do not run the original implementations of
PEQA, QA-LoRA, L4Q, LoTA-QAF, or LoRA-Inlaid, so comparisons to those works are analytical. LoTA-QAF is the
sharpest untested comparison: it also merges losslessly, and because its ternary adjustment rewrites the
quantized codes it should forfeit the code-plane sharing, rollback, and dedup payoffs we measure. Testing
this faithfully requires its own trainer: a single ternary code step applied to a weight-space update
merely re-encounters the sub-step deletion of Section~\ref{sec:deletion} (it changed $<0.01\%$ of code
bytes and returned the base), not a fair stand-in for LoTA-QAF's trained adjustments. We therefore leave
that same-task code-byte measurement to future work.
Three further limits of the present evidence: the cross-tool divergence census is per-layer over every
target linear of a \emph{single} checkpoint (one model family, one grid, one merged artifact, and,
of its six outputs, three produced by our own harnesses against three by independently shipping tools,
so only three of the enumerated pairs are tool-against-tool between implementations neither of which is
ours), while the two divergences carrying the about-a-point price are measured on one task, one a
convention we construct rather than a shipped default, so the census establishes that divergence is the
norm among these tools on that artifact, not a rate over models, tasks or formats; the reorientation
null rests on four tasks with a near-degenerate statistic (Appendix~\ref{sec:expressivity}); and the
bridge sub-studies use the lighter pilot protocol.

\section{Calibrated PTQ on a Sparse MoE Is Hessian-Starved}
\label{app:gptq}

GPTQ's error feedback needs $H = \mathbb{E}[xx^{\!\top}]$ per expert, but an expert only sees the tokens routed
to it, so rank$(H)$ is bounded by that expert's token count. On DeepSeek-V4 with $1.57$M calibration tokens,
$33.7\%$ of \texttt{gate\_up} Hessians on Banking77 have fewer routed tokens than the $4096$ input channels
(agnews: $18.6\%$), and $40$ of $11008$ experts receive \emph{no} tokens at all, falling back to a data-free
grid on $80$ of $22016$ matrices. For the rank-deficient experts the inverse is carried by the $1\%$ damping
rather than by data. This is a property of calibrating a sparse MoE under a token budget, not of our
implementation (Section~\ref{sec:limitations}). We note, against our own interest in a clean caveat, that the
conditioning does \emph{not} visibly degrade any GPTQ cell in this matrix, and the sharpest case is also
the one with real headroom. On DeepSeek \textbf{Spider}, $54.3\%$ of \texttt{gate\_up} Hessians are
rank-deficient and the \emph{median} expert saw only $3493$ routed tokens against $4096$ input channels,
yet the merge lands at $+0.6\pp$, matching the naive RTN merge exactly, on a cell with $31.2\pp$ of fine-tuning headroom
and no accuracy ceiling. CLINC150 ($35.5\%$ rank-deficient) merges at $0.00\pp$ and Banking77 ($33.7\%$) at
$-0.8\pp$. The bridge cells are the worst-conditioned because they calibrate from fewer tokens per expert
than the in-EP cells (512 windows, dropless routing, no capacity ceiling), which is exactly why they are the
informative test. We therefore state the caveat as a limit on the \emph{guarantees} of calibrated PTQ on a
sparse MoE (a Hessian below full rank means the error-feedback objective is not the one the method
assumes) and not as a doubt about these numbers, which a $54\%$-rank-deficient cell with ample headroom
declines to support. Token counts are additionally clipped from
above by the expert-parallel capacity ceiling, so the upper tail of the distribution is an artifact of the
serving configuration rather than of routing.

\paragraph{Rounding-only quantizers, calibration, and Hessian starvation, in full.} We benchmark four post-hoc quantizers on the same
single adapter (Table~\ref{tab:ptqbaselines}). On the NVFP4 models RTN, MSE,
and AWQ differ from each other by a small fraction of the collapse they all produce, and AWQ is sometimes
\emph{worse} (Llama CLINC150 $-30.4$ vs RTN's $-29.3$); on the E8M0 MoEs a weight-MSE search is far worse still, for a
reason we isolate in Section~\ref{sec:deletion}. A calibrated quantizer is the natural alternative, and our
baseline is not a weak one: HuggingFace's MXFP4 path, the one
that actually serves gpt-oss, selects block scales by $\lceil\log_2(\mathrm{absmax}/6)\rceil$ with the
mantissa masked off, bit-for-bit the RTN rule we report, so our headline baseline \emph{is} the shipping
default. AWQ, default-on in its own toolchain, is statistically indistinguishable from it at $120$B
scale (gpt-oss AGNews, identical GPU set and unmerged reference: $-2.5$ against RTN's $-2.6\pp$, a
separation below the $0.3\pp$ run-to-run floor of Section~\ref{sec:setup}, so not a difference we can
claim in either direction); why an activation-aware objective has so little room on a microscaled grid is
Appendix~\ref{app:awq}.
\emph{Every gpt-oss AGNews merge loss quoted in prose (here, GPTQ below, weight-MSE in
Section~\ref{sec:deletion}) pairs its merged accuracy with its own run's unmerged reference, whereas
that cell's Table~\ref{tab:ptqbaselines} column uses the authoritative one; the two therefore differ by
less than that same floor: they come from separate runs and are not to be combined
(Appendix~\ref{app:protocol}).}
\textbf{Only GPTQ's Hessian error-feedback partially
rescues} (Llama Banking77 $-37.4\to-2.5\pp$; Qwen Banking77 $-23.3\to-0.9\pp$), and on gpt-oss-120B
AGNews it rescues to within noise ($92.0$ merged vs $92.7$ unmerged measured in the same run,
$-0.7\pp$, $95\%$ CI $[90.2, 93.5]$ containing zero, $n{=}1000$). Our exactness claim is precise about
what it is and is not: GPTQ can be \emph{statistically} lossless on a cell but never \emph{bit}-exact,
since it re-derives every E2M1 code, forfeiting every code-invariance property of
Section~\ref{sec:codeinv-short} regardless of accuracy; and it is cell- and calibration-dependent,
needing a calibration set and an $O(\text{in}^2)$ per-layer Hessian ($\sim39$GB here) re-run at every
merge, re-emitting calibration-specific codes that re-couple correctness to the serving quantizer, and on
a sparse MoE it is starved of routed tokens, which limits the \emph{guarantees} calibrated PTQ can offer
there (Section~\ref{sec:limitations}; quantified in Appendix~\ref{app:gptq}). The exact-$0$,
calibration-free, deploy-quantizer-independent merge is \scaleqast{}'s. The naive merge can even be
\emph{positive} when the adapter is under-fit (observed in pilot runs), so its sign is unpredictable per
deployment. The one positive entry, Qwen Spider GPTQ at $+1.8\pp$, is within the noise of a
base-saturated cell ($2.6\pp$ of headroom). Earlier DeepSeek runs on different checkpoints gave
different absolute numbers, including a naive-merge \emph{sign flip} on the banking-intent cell
(Appendix~\ref{app:cost}).

\begin{table*}[t]
  \centering
  \footnotesize
  \caption{\textbf{Naive-merge collapse across four post-hoc quantizers} (merge loss, pp; \emph{Base} is
  the un-adapted 4-bit checkpoint's accuracy on the same harness). All four quantizers apply to the
  \emph{same} single naive adapter, no per-quantizer retraining, and each merge loss is paired with an
  unmerged reference measured on its own harness. \textbf{The two grids behave oppositely, which is the
  point.} On NVFP4 the three rounding-only quantizers collapse together: inside a cell they differ from
  each other by a small fraction of the collapse they all produce. Reading the AWQ column, the merged model ends up only
  $0.3$--$7.1\pp$ above simply \emph{not fine-tuning} across the six classification cells and $1.5\pp$
  \emph{below} it on Llama Spider, while the RTN column runs wider at both ends and further below base
  there, which is why the merged-minus-base readings for Spider differ between this table and the main
  matrix: one measurement through two quantizers, not two disagreeing ones. On MXFP4/E8M0 the same three separate sharply: RTN stays mild across both
  MoEs ($-3.6$ to $+0.6\pp$) and AWQ tracks it on seven of eight cells ($-2.8$ to $+0.3$; the exception is
  DeepSeek AGNews at $-10.8$, discussed in Section~\ref{sec:deletion}), while a weight-MSE scale
  search is catastrophic ($-1.2$ to $-59.8\pp$), landing at or just below the un-adapted base. It
  destroys $105$--$110\%$ of the adaptation gain on every DeepSeek dataset, which is the
  headroom-normalised form of that result. Section~\ref{sec:deletion} shows this is one
  mechanism rather than two, and scopes how far it goes: reconstruction quality tracks the collapse
  \emph{directionally}, not as a predictor of its size. E8M0's $\lceil\cdot\rceil$ rule survives by
  reconstructing badly, while RTN and AWQ swap order between the two rankings and the one AWQ cell above
  is not accounted for by reconstruction at all. GPTQ's Hessian
  error-feedback rescues on both grids because its objective is not per-block weight error.
  \scaleqast{} is exactly $0.0$ by construction on every cell.}
  \label{tab:ptqbaselines}
  \begin{tabular}{@{}ll r cccc@{}}
    \toprule
    Model & Dataset & Base & RTN & MSE & AWQ & GPTQ \\
    \midrule
    \multirow{4}{*}{\shortstack[l]{Llama-8B\\NVFP4}}
      & banking77 & 49.2 & \bad{$-37.4$} & \bad{$-39.1$} & \bad{$-38.9$} & $-2.5$ \\
      & agnews    & 76.3 & \bad{$-13.3$} & \bad{$-13.3$} & \bad{$-13.3$} & $-1.5$ \\
      & clinc150  & 60.1 & \bad{$-29.3$} & \bad{$-30.1$} & \bad{$-30.4$} & $-0.6$ \\
      & spider    & 62.4 & $-9.7$        & $-9.4$        & $-9.5$        & $-0.6$ \\
    \midrule
    \multirow{4}{*}{\shortstack[l]{Qwen3-30B\\NVFP4 (MoE)}}
      & banking77 & 69.6 & \bad{$-23.3$} & \bad{$-23.1$} & \bad{$-22.7$} & $-0.9$ \\
      & agnews    & 80.8 & \bad{$-11.1$} & \bad{$-11.2$} & \bad{$-10.8$} & $-0.3$ \\
      & clinc150  & 81.0 & \bad{$-14.0$} & \bad{$-13.9$} & \bad{$-13.2$} & $-0.3$ \\
      & spider    & 71.3 & $-3.3$        & $-3.3$        & $-2.3$        & $+1.8$ \\
    \midrule
    \multirow{4}{*}{\shortstack[l]{gpt-oss-120B\\MXFP4/E8M0 (MoE)}}
      & banking77 & 73.9 & $-0.8$ & \bad{$-17.3$} & $+0.1$ & $-0.1$ \\
      & agnews    & 74.0 & $-2.6$ & \bad{$-16.9$} & $-2.6$ & $-0.5$ \\
      & clinc150  & 67.5 & $-2.6$ & \bad{$-27.7$} & $-2.8$ & $-0.4$ \\
      & spider    & 71.3 & $-1.5$ & $-1.2$        & $+0.3$ & $+0.4$ \\
    \midrule
    \multirow{4}{*}{\shortstack[l]{DeepSeek-V4-Flash\\MXFP4/E8M0 (MoE)}}
      & banking77 & 60.6 & $-2.0$ & \bad{$-38.8$} & $-2.2$          & $-0.8$ \\
      & agnews    & 33.6 & $-3.6$ & \bad{$-59.8$} & \bad{$-10.8$}   & $-2.0$ \\
      & clinc150  & 66.0 & $-0.8$ & \bad{$-36.4$} & $0.0$           & $0.0$ \\
      & spider    & 40.4 & $+0.6$ & \bad{$-33.2$} & $-1.2$          & $+0.6$ \\
    \bottomrule
  \end{tabular}
\end{table*}

\begin{figure*}[t]
  \centering
  \includegraphics[width=0.62\textwidth]{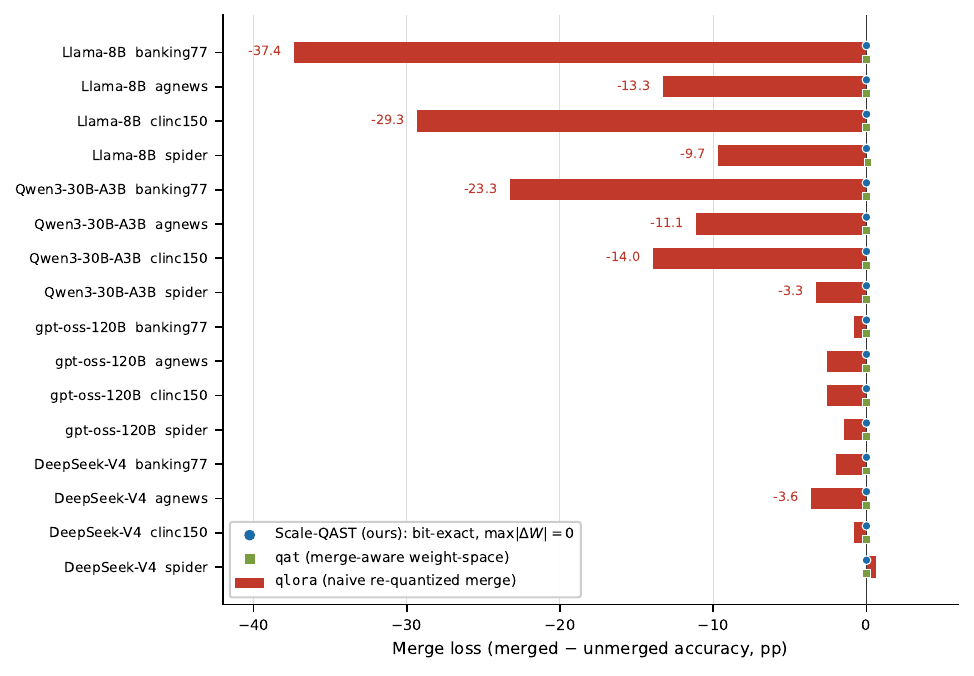}
  \caption{\textbf{Merge loss across the full matrix} (generated directly from
  Table~\ref{tab:main}). Both merge-aware methods sit on the zero line in every cell; the naive
  re-quantized merge is lossy in all but one, and its magnitude is not predictable from the model or the
  task. \scaleqast{}'s zeros are an identity verified in weight space ($\max|\Delta W|=0$), not an
  accuracy coincidence. \textbf{The two merge-aware markers are not to be read against each other}
  (Section~\ref{sec:limitations}).}
  \label{fig:mergeloss}
\end{figure*}

\end{document}